\documentclass{article}

\usepackage[final,nonatbib]{neurips_data_2021}

\usepackage[utf8]{inputenc} % allow utf-8 input
\usepackage[T1]{fontenc}    % use 8-bit T1 fonts
\usepackage{url}            % simple URL typesetting
\usepackage{booktabs}       % professional-quality tables
\usepackage{amsfonts}       % blackboard math symbols
\usepackage{nicefrac}       % compact symbols for 1/2, etc.
\usepackage{microtype}      % microtypography
\usepackage{xcolor}         % colors

\usepackage{times}
\usepackage{epsfig}
\usepackage{graphicx}
\usepackage{amsmath}
\usepackage{amssymb}

\usepackage{url}
\usepackage{adjustbox}
\usepackage{booktabs}

\usepackage{multirow}
\usepackage{color, colortbl}
\usepackage{placeins}
\usepackage{subfig}
\usepackage{bbm}

\usepackage{authblk}
\makeatletter
\renewcommand\AB@affilsepx{ \hspace{1em}   \protect\Affilfont}
\makeatother

\usepackage[pagebackref=true,breaklinks=true,letterpaper=true,colorlinks,bookmarks=false]{hyperref}

\usepackage{times}
\usepackage{epsfig}
\usepackage{graphicx}
\usepackage{float}
\usepackage{wrapfig}
\usepackage{amsmath,amssymb,amsthm}
\usepackage{algorithm,algorithmicx,algpseudocode}
\usepackage{bm,xspace}
\usepackage{comment}
\usepackage{verbatim}
\usepackage{multirow}
\usepackage{balance}
\usepackage{url}
\usepackage{booktabs}
\usepackage{etoolbox,siunitx}
\usepackage{calc}
\usepackage{pifont,hologo}
\usepackage{nicefrac}
\definecolor{TableauOrange}{RGB}{242,142,44}
\definecolor{TableauGreen}{RGB}{89,161,79}

\usepackage[super]{nth}
\usepackage{nicefrac}
\robustify\bfseries

\usepackage{dsfont}

\DeclareMathSymbol{@}{\mathord}{letters}{"3B}

\def\latex/{\LaTeX}
\def\bibtex/{\hologo{BibTeX}}

\definecolor{forestgreen}{RGB}{34, 139, 34}
\definecolor{LightGrayForTableRule}{gray}{0.92}
\definecolor{Black}{rgb}{0.0, 0.0, 0.0}

\newcommand*\samethanks[1][\value{footnote}]{\footnotemark[#1]}

\begin{document}

\title{Trust, but Verify: Cross-Modality Fusion \\for HD Map Change Detection} %\vspace{-0.18 mm}}

% The \author macro works with any number of authors. There are two commands
% used to separate the names and addresses of multiple authors: \And and \AND.
%
% Using \And between authors leaves it to LaTeX to determine where to break the
% lines. Using \AND forces a line break at that point. So, if LaTeX puts 3 of 4
% authors names on the first line, and the last on the second line, try using
% \AND instead of \And before the third author name.

\author[ ]{\textbf{John Lambert}\thanks{Work completed while at Argo AI.}\hspace{1mm} }
\author[1]{\textbf{James Hays}\samethanks \hspace{1mm} }
%\affil[1]{Argo AI}
\affil[1]{Georgia Institute of Technology}

%\maketitle

% \onecolumn[{
% \renewcommand\onecolumn[1][]{#1}
\vspace{-3em}
\maketitle
%\centering
\vspace{-2.5em}
\begin{tabular}{@{}c@{}}
	\includegraphics[width=\linewidth]{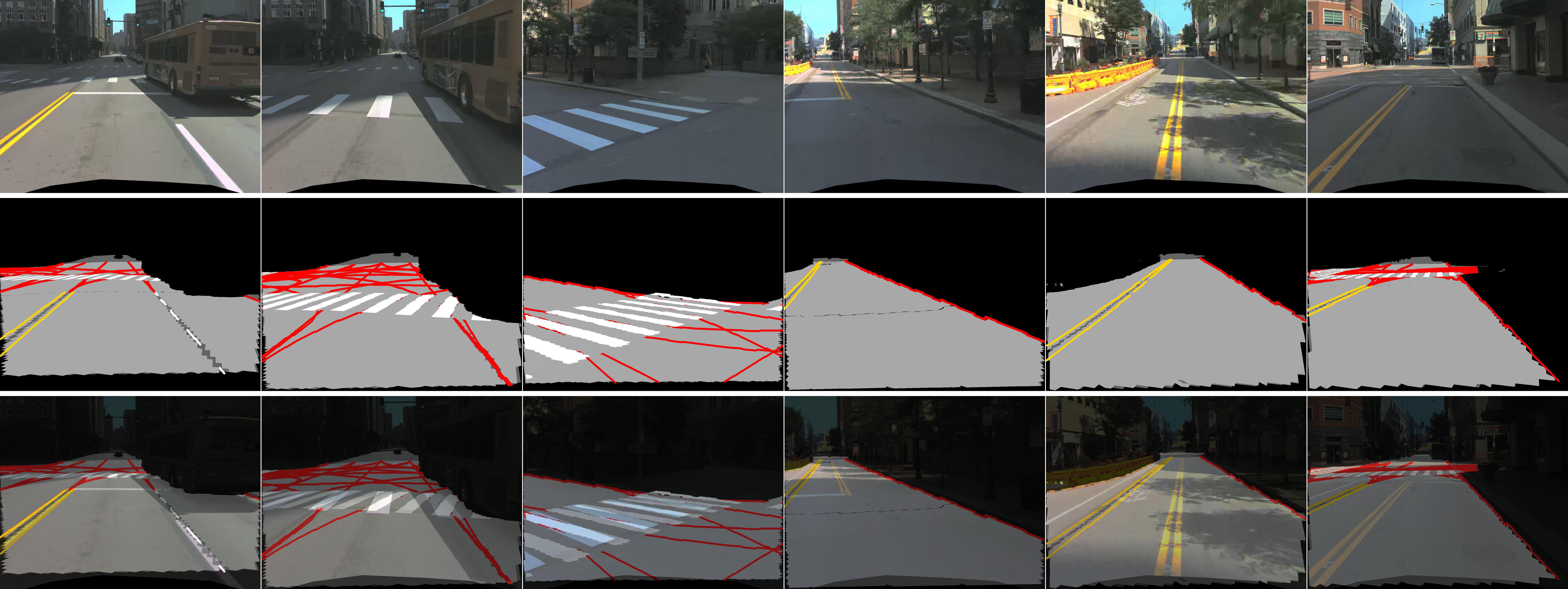}
\end{tabular}
\begin{tabular}
	{@{}p{0.165\linewidth}@{\hspace{0.1mm}}p{0.165\linewidth}@{\hspace{0.1mm}}p{0.165\linewidth}@{\hspace{0.1mm}}p{0.165\linewidth}@{\hspace{0.1mm}}p{0.165\linewidth}@{\hspace{0.1mm}}p{0.165\linewidth}@{}}
\hspace{8mm} \textcolor{forestgreen}{Match} & \hspace{8mm} \textcolor{forestgreen}{Match} & \hspace{8mm} \textcolor{forestgreen}{Match} & \hspace{7mm} \textcolor{red}{Mismatch} & \hspace{7mm} \textcolor{red}{Mismatch} & \hspace{7mm} \textcolor{red}{Mismatch} \\
\end{tabular}
\captionof{figure}{Scenes from an example log from our TbV dataset where a real-world change causes an HD map to become stale. Corresponding sensor data (top), map data (middle), and a blended combination of the two (bottom) are shown at 6 timestamps in chronological order from left to right. Within each column, all images are captured or rendered from identical viewpoints. We use red to denote implicit lane boundaries. Faint dark road strips in row 1, columns 4-6 show where the previous paint was stripped away.}
\label{fig:teaser}
\vspace{1em}
% }]

\begin{abstract}
%\vspace{-4mm}
High-definition (HD) map change detection is the task of determining when sensor data and map data are no longer in agreement with one another due to real-world changes. We collect the first dataset for the task, which we entitle the \emph{Trust, but Verify} %\footnote{From the Russian proverb, `\foreignlanguage{russian}{Доверяй, но проверяй}' (`\emph{Trust, but verify}').} 
(TbV) dataset, by mining thousands of hours of data from over 9 months of autonomous vehicle fleet operations. We present learning-based formulations for solving the problem in the bird's eye view and ego-view. Because real map changes are infrequent and vector maps are easy to synthetically manipulate, we lean on simulated data to train our model. Perhaps surprisingly, we show that such models can generalize to real world distributions. The dataset, consisting of maps and logs collected in six North American cities, is one of the largest AV datasets to date with more than 7.8 million images. We make the data\footnote{Data is available at \href{https://www.argoverse.org}{Argoverse.org}. Sample videos can be found at \href{https://johnwlambert.github.io/tbv-dataset/}{tbv-dataset.github.io}.} available to the public, along with code and models\footnote{Code and models are available at \href{https://github.com/johnwlambert/tbv}{github.com/johnwlambert/tbv}. \vspace{-1.6em} } under the the CC BY-NC-SA 4.0 license.
\end{abstract}

%%%%%%%%%% BODY TEXT
\section{Introduction}

We live in a highly dynamic world, so much so that significant portions of our environment that we assume to be static are, in fact, in flux. Of particular interest to self-driving vehicle development is changing road infrastructure. Road infrastructure is often represented in an onboard map within a geo-fenced area. Geo-fenced areas have served as an operational design domain for self-driving vehicles since the earliest days of the DARPA Urban Challenge \cite{Urmson07tr_TartanRacing,Montemerlo08jfr_JuniorUrbanChallenge,Bacha08jfr_VictorTangoDARPA}.\\

%The Society of Automotive Engineers
%% s defined as the ability of a vehicle to perform all driving tasks in certain circumstances -- 

One way such maps could be used is to constrain navigation in all free space to a set of legal ``rails'' on which a vehicle can travel. Maps may also be used to assist in planning beyond the sensor range and in harsh environments.  Besides providing routes for navigation, maps can ensure that the autonomous vehicle (AV) follows local driving laws when navigating through a city. They embody a representation of the world that the AV can understand, and contain valuable information about the environment. Building and validating maps represent an essential and general part of spatial artificial intelligence (AI) \cite{Davison18arxiv_FutureMapping}, embodied intelligence that enables safety-critical awareness of a robot's surroundings. \\

However, maps assume a static world, an assumption which is violated in practice; although these changes are rare, they certainly occur and will continue to occur, and can have serious implications. Level 4 autonomy is defined as sustained performance by an autonomous driving system within an operational design domain, without any expectation that a user will respond to a request to intervene \cite{Sae18tr_SelfDrivingTaxonomy}. Thus, constant verification that the represented world, expressed as a map, matches the real world, a task known as \emph{map change detection} \cite{Pannen19icra_ChangeDetectionParticleFilter, Pannen20icra_HDMapUpToDate, Karpathy20talk_CVPR}, is a clear requirement for L4 autonomy. This problem may be solved in the \emph{classification} setting, reasoning globally over the map and scene, or additionally in a \emph{localization} setting, where the spatial extent of changed map entities are locally identified. Because dedicated mapping vehicles cannot traverse the world frequently enough to keep maps up to date \cite{Pannen20icra_HDMapUpToDate}, high-definition (HD) maps become ``stale,'' with out of date information. If maps are used as hard priors, this could lead to confident but incorrect assumptions about the environment. \\
%potentially leading to situations of high confidence despite high danger. 
% FIND CITATION LATER ABOUT THE HIGH CONFIDENCE, re-read related work

% Wasn't there some paper that had contrived map changes, as in they moved around traffic cones or something? 

In this work, we present the first public dataset for urban map change detection based on actual, observed map changes, which we name \emph{TbV}. While researchers could use paired sensor data and HD maps from other datasets such as Argoverse \cite{Chang19cvpr_Argoverse} or nuScenes \cite{Caesar20cvpr_nuscenes} to hypothesize about map change detection performance in the real world based on synthetic data, these datasets include zero real map changes, meaning one could not know with any accuracy or degree of certainty about how well the system would actually operate in the real world. Concurrent work \cite{Heo20iros_HDMapChangeCrossDomain}  presents qualitative results on a handful of real-world map changes on a proprietary dataset, but relies upon synthetic test datasets for all quantitative evaluation. Not only does no comparable dataset to TbV exist, there also has not even been an attempt to characterize how often map changes occur and what form they take. Collecting data for map change detection is challenging since changes occur randomly and infrequently. In addition, in order to use data corresponding to real changes to train and evaluate models, identified changes must be manually localized in both space and time.\\

HD map change detection is a difficult task even for humans, as it requires the careful comparison of all nearby semantic entities in the real world with all nearby map elements in the represented world. In an urban scene, there can be dozens of such entities, many with extended shapes. The task is sufficiently difficult that several have even questioned the viability of HD maps for long-term autonomy, opting instead to pursue HD-map-free solutions \cite{Karpathy20talk_CVPR}. We concentrate on changes to two types of semantic entities -- lane geometry and pedestrian crosswalks. We define the task as correctly classifying whether a change occurred at evenly spaced intervals along a vehicle's trajectory.\\

 The task itself is relatively new, especially since HD maps were not made publicly available until the release of the Argoverse, nuScenes, Lyft Level5, and Waymo Open Motion datasets \cite{Chang19cvpr_Argoverse,Caesar20cvpr_nuscenes,Houston20_LyftLevel5,Waymo21_MotionDataset}. We present the first entirely learning-based formulation for solving the problem in either a bird's eye view (BEV), as well as a new formulation for the ego-view (i.e. front camera frustum), eliminating several heuristics that have defined prior work. We pose the problem as learning a representation of maps, sensor data, or the combination of the two.\\

 Our contributions are as follows:
\begin{itemize}
	\item We present a novel AV dataset, with 799 vehicle logs in our train and synthetic validation splits, and over 200 vehicle logs with real-world map-changes in our real val and test splits.
	%ICCV\vspace{-2mm}
	\item We implement various learning-based approaches as strong baselines to explore this task for the first time with real data. We also demonstrate how gradients flowing through our networks can be leveraged to localize map changes. % in 2d or 3d.
	%ICCV\vspace{-2mm}
	\item We analyze the advantages of various data viewpoint by training both models operating on the ego-view and others on a bird's eye view.
	%ICCV\vspace{-2mm}
	\item We show that synthetic training data is useful for detecting real map changes. At the same time, we identify a considerable domain gap between synthetic and real data, with significant performance consequences.
\end{itemize}

\section{Related Work}

\noindent\textbf{HD Maps.} HD maps include lane-level geometry, as well as other geometric data and semantic annotations \cite{Mattyus16cvpr_HDMaps,Mattyus17iccv_DeepRoadMapper,Wang17iccv_TorontoCity,Yang18corl_HDNet,Homayounfar18cvpr_HRAN,Chang19cvpr_Argoverse,Homayounfar19iccv_DAGMapper,Zeng19cvpr_NeuralMotionPlanner, Garnett19iccv_3dLaneNet,Gao20cvpr_VectorNet,Phan-Minh20cvpr_CoverNet, Liang20eccv_LaneGCN, Chen21cvpr_GeoSim,Tan21cvpr_SceneGen, Suo21cvpr_TrafficSim}. %Low-definition routing maps, such as Google Maps
The Argoverse \cite{Chang19cvpr_Argoverse}, nuScenes \cite{Caesar20cvpr_nuscenes}, Lyft Level 5 \cite{Houston20_LyftLevel5}, Waymo Open Motion \cite{Waymo21_MotionDataset}, and Argoverse 2.0 \cite{Wilson21neurips_Argoverse2} datasets are the only publicly available sources of HD maps today, all with different semantic entities. Argoverse \cite{Chang19cvpr_Argoverse} includes a ground surface height map, rasterized driveable area, lane centerline geometry, connectivity, and other attributes. nuScenes \cite{Caesar20cvpr_nuscenes} followed by also releasing centerline geometry, pedestrian crossing polygons, parking areas, and sidewalk polygons, along with rasterized driveable area. Lyft Level 5 \cite{Houston20_LyftLevel5} later provided a dataset with many more map entities, going beyond lane marking boundaries, crosswalks to provide traffic signs, traffic lights, lane restrictions, and speed bumps. The Waymo Open Motion Dataset \cite{Waymo21_MotionDataset} released motion forecasting scenario data with associated HD maps. Their yet-richer HD map representation includes crosswalk polygons, speed bump polygons, lane boundary polylines with marking type, lane speed limits, lane types, and stop sign positions and their corresponding lane associations; their map data is most comparable with our HD maps. HD maps are useful for a range of tasks, from perception \cite{Yang18corl_HDNet,Casas18corl_IntentNet,Chang19cvpr_Argoverse}, to motion forecasting, to motion planning \cite{Chen19corl_LearningByCheating,Zeng19cvpr_NeuralMotionPlanner}, to traffic scene simulation \cite{Tan21cvpr_SceneGen,Suo21cvpr_TrafficSim,Chen21cvpr_GeoSim}. All state-of-the-art motion forecasting methods for self-driving today use HD maps \cite{Gao20cvpr_VectorNet, Liang20eccv_LaneGCN, Phan-Minh20cvpr_CoverNet, Zhao20corl_TNT}.

\paragraph{HD Map Change Detection.} HD map change detection is a recent problem, with limited prior work.  Pannen \emph{et al.}  \cite{Pannen19icra_ChangeDetectionParticleFilter} introduce one of the first approaches; two particle filters are run simultaneously, with one utilizing only Global Navigation Satellite System (GNSS) and odometry measurements, and the other filter using only odometry with camera lane and road edge detections. These two distributions and sensor innovations are then fed to weak classifiers. Other prior work in the literature seeks to define hand-crafted heuristics for associating online lane and road detections with map entities \cite{Jo18sensors_SLAM_MapChangeUpdate,Silver16patent_ChangeDetCurveAlignment,Pannen20icra_HDMapUpToDate}. These methods are usually evaluated on a single vehicle log \cite{Jo18sensors_SLAM_MapChangeUpdate}.\\

%Jo \emph{et al.} \cite{Jo18sensors_SLAM_MapChangeUpdate} introduce a probabilistic framework to reason about new, deleted, and unchanged physical map features; online measurements are associated with map features. However, only a single vehicle log is evaluated for map change detection. In a similar spirit, Silver and Ferguson \cite{Silver16patent_ChangeDetCurveAlignment} propose a system that seeks to identify every nearby lane marking or other semantic object in real-time, and compute association costs with HD map entities. These systems require extensive heuristics and thresholds to account for the diversity of change types. Pannen \etal \cite{Pannen20icra_HDMapUpToDate}  also associate online lane and road edge detections with map features, defining an association threshold of 2.5 m. They partition a road into discrete ``linklets'' and feed features from associated entities to a gradient boosted tree regressor which outputs change probabilities.
Instead of comparing vector map elements, Ding \emph{et al.} \cite{Ding20icra_ChangingCityScenes} use 2d BEV raster representations of the world; first, IMU-motion-compensated LiDAR odometry is used to build a local online ``submap''. Afterwards, the submap is projected to 2d and overlaid onto a prebuilt map; the intensity mean, intensity variance, and altitude mean of corresponding cells are compared for change detection. Rather than pursuing this approach, which requires creating and storing high-resolution reflectance maps of a city, we pursue the alignment of \emph{vector maps} with sensor data. Vector maps can be encoded cheaply with low memory cost and are the more common representation, being used in all five public HD map datasets.\\

In concurrent work, Heo \emph{et al.} \cite{Heo20iros_HDMapChangeCrossDomain} introduce an adversarial metric learning-based formulation for HD map change detection, but access to their dataset is restricted to South Korean researchers and performance is measured on a synthetic dataset, rather than on real-world changes. They employ a forward-facing ego-view representation, and require training a second, separate U-Net model to localize changed regions in 2d, whereas we show changed entity localization can come for free via examination of the gradients of a single model.

\paragraph{Mapping Dynamic Environments.} While ``HD maps'' are a relatively new entity, dynamic map construction is a more mature field of study. Semi-static environments are not limited to urban streets; households, offices, warehouses, and parking lots are relatively fixed environments that a robot may navigate, with changing cars, furniture, and goods \cite{Tipaldi13ijrr_LifelongLoc}. Mapping dynamic environments has been an area of study within the SLAM community for decades
\cite{Shaik17ki_DynamicMapUpdate,Wang02icra_DetectTrackMovingObjects,Hahnel03icra_MapBuildingDynamic}.  However, we focus purely on change detection, rather than map updates. 

Recently, machine learning for online mapping has generated interest. An alternative to using an HD map prior is to rebuild the map on-the-fly during robot operation; however, such an approach cannot map occluded objects or entities. In addition, these methods are limited to producing raster map layers, such as a driveable area mask, with an output resembling semantic segmentation. Raster data is significantly less useful than vector data for path planning and generating vector map data with machine learning is generally an unsolved problem. Raster map layers may be generated from LiDAR \cite{Yang18corl_HDNet}, accumulated from networks operating on ego-view images over multiple cameras and timesteps \cite{Roddick20cvpr_SemanticMap,Philion20eccv_LiftSplatShoot,Pan19ral_crossview}, or from a single image paired with a depth map or LiDAR \cite{Mani20wacv_MonoLayout}. They all show that automatic mapping is quite challenging.

\paragraph{Image-to-Image Change Detection.}
Image-to-image scene change detection over the temporal axis is a well-studied problem \cite{Wang14iccvw_CDnet,Alcantarilla18ar_StreetviewChangeDet}.
Scenes are dynamic over time in numerous ways, and those ways are mostly nuisance variables for our purposes. We wish to develop models invariant to season, lighting, the fading of road markings, and occlusion because these variables don't actually change the lane geometry. Wang \emph{et al.} \cite{Wang14iccvw_CDnet} introduced the CDnet benchmark, a collection of videos with frame pixels annotated as static, shadow, non-ROI, unknown, or moving. Alcantarilla \cite{Alcantarilla18ar_StreetviewChangeDet} \emph{et al.} introduce the VL-CMU-CD street view change detection benchmark, from a subset of the Visual Localization CMU dataset.

\section{The TbV Dataset}

We curate a novel dataset of autonomous vehicle driving data comprising 1043 logs, over 200 of which contain map changes. The vehicle logs are on average 54 seconds in duration, collected in six North American cities: Austin, TX, Detroit, MI, Miami, FL, Palo Alto, CA, Pittsburgh, PA, and Washington, D.C.

\begin{figure}[htb]
 \vspace{-2mm}
 \centering
 %\begin{tabular}{c} %@{}c@{}}
 
\subfloat[crosswalk is removed]{
  \includegraphics[width=0.45\linewidth]{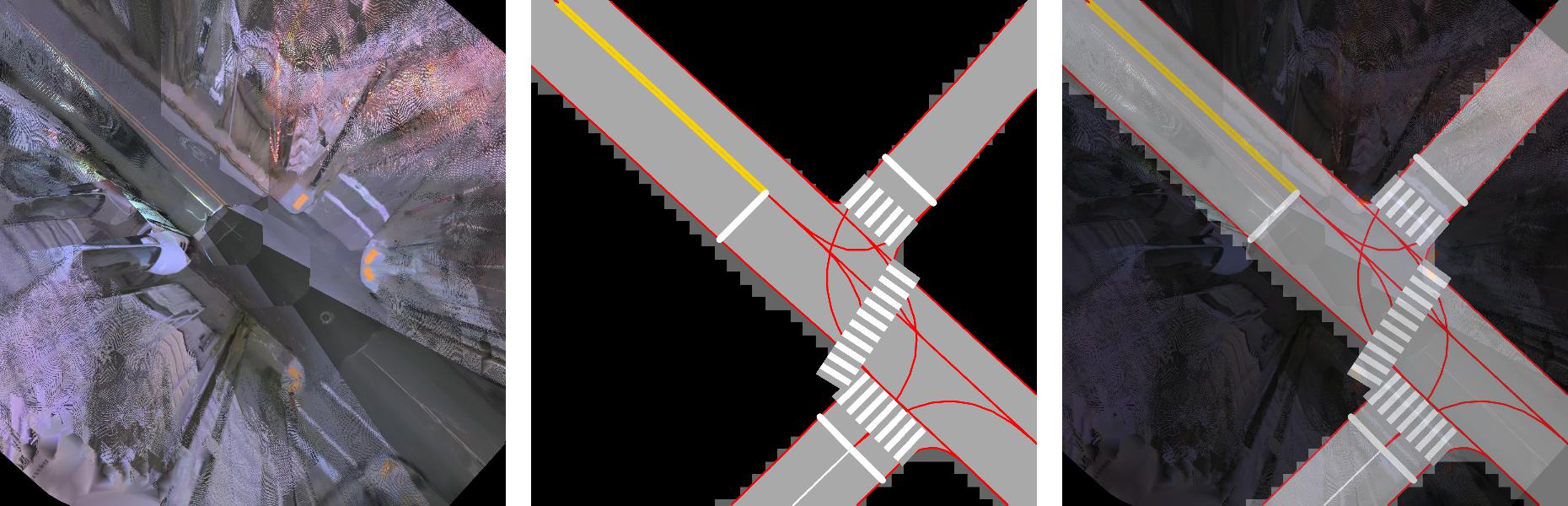}
}
\hspace{5mm}
\subfloat[dashed white repainted as double solid white ]{
   \includegraphics[width=0.45\linewidth]{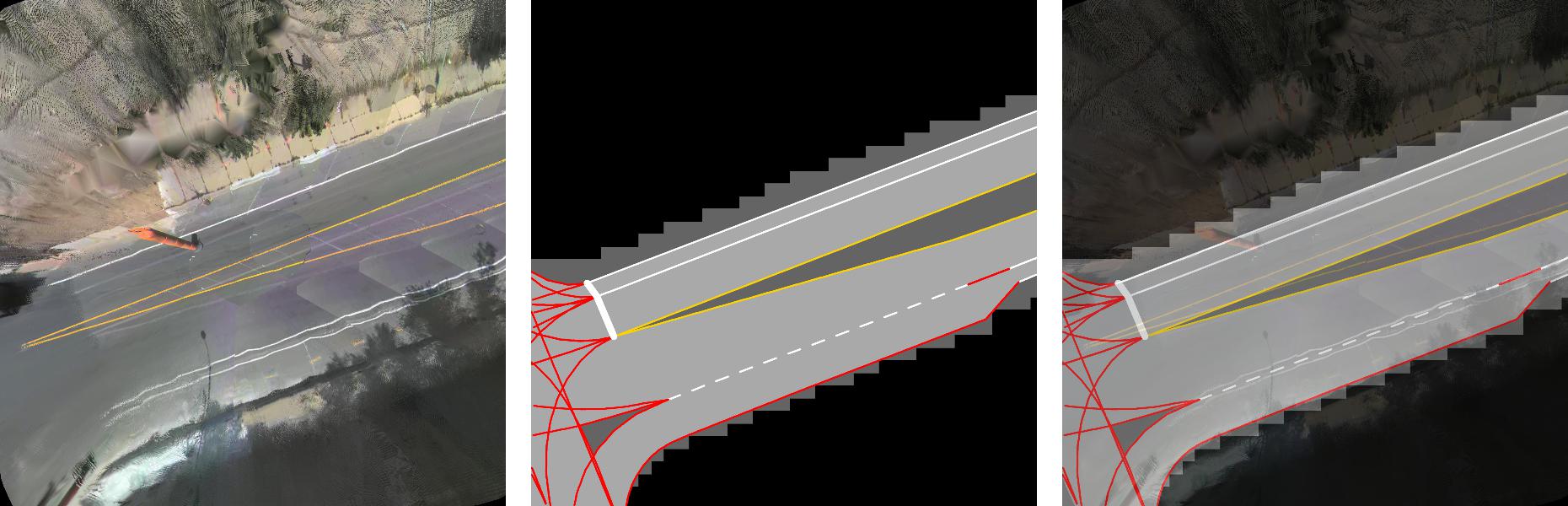}
}
\hspace{0mm}
\subfloat[solid yellow paint marking is added]{
  \includegraphics[width=0.45\linewidth]{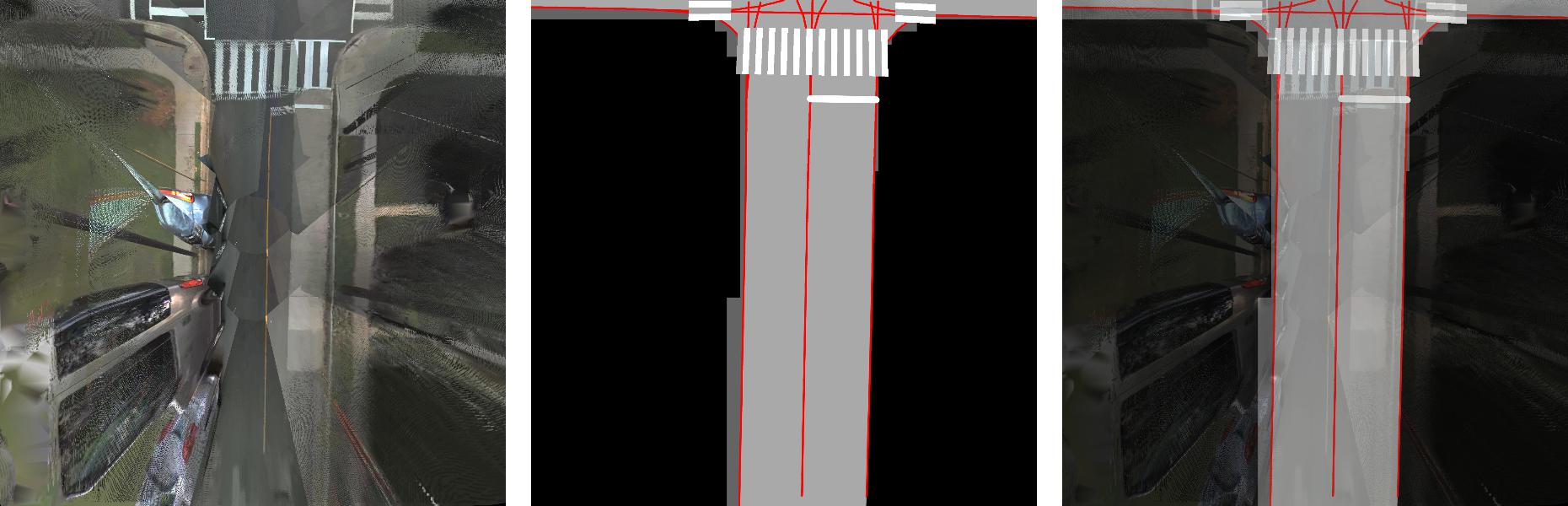}
}
\hspace{5mm}
\subfloat[double-solid yellow paint marking is removed]{
   \includegraphics[width=0.45\linewidth]{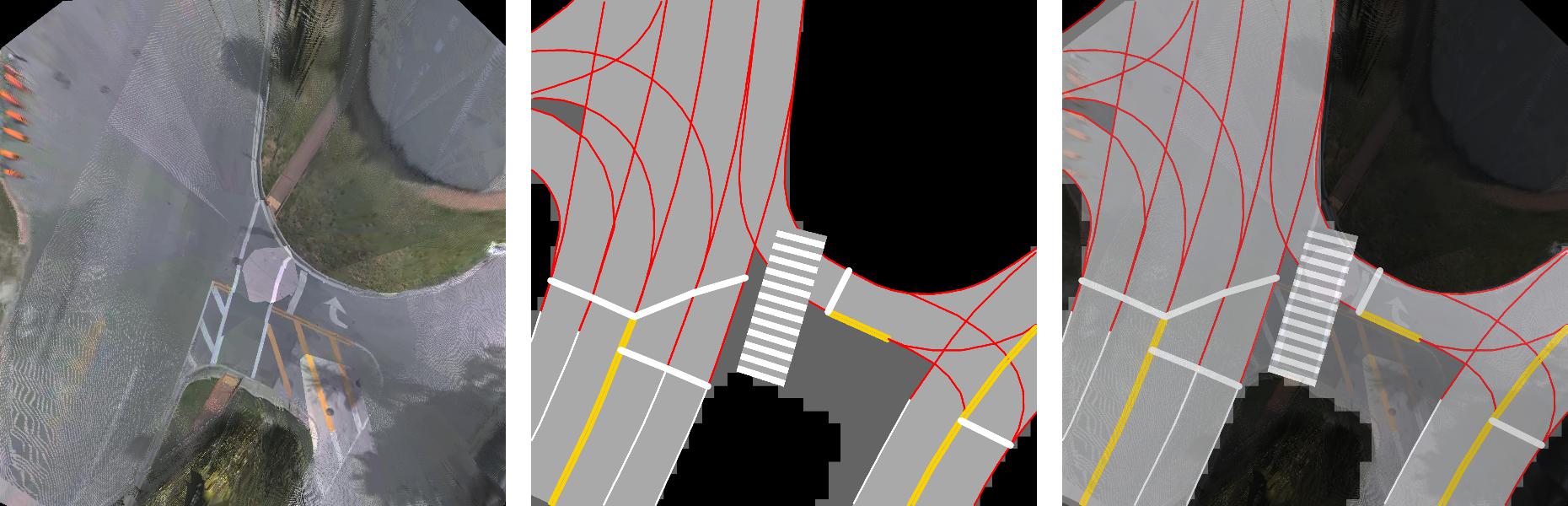}
}
  \caption{Examples from the validation and test splits of our TbV dataset. Left to right: BEV sensor representation, onboard map representation, blended map and sensor representations. Rows, from top to bottom: deleted crosswalk (top row), and painted lane geometry changes (bottom three rows). }
\vspace{-4mm}
\label{fig:testsetexamples}
 \end{figure}

Our training set consists of real data with accurate corresponding onboard maps (``positives''). Accordingly, synthetic perturbation of positives to create plausible ``negatives'' is required for training. We release the data, code and API to generate them. However, in the spirit of other datasets meant for testing only (i.e. not training) such as the influential WildDash dataset \cite{Zendel18eccv_WildDash}, we curate our validation and test splits from the real-world distribution. We do so since map changes are difficult to mine \cite{Heo20iros_HDMapChangeCrossDomain}, thus we save their limited quantity for testing and evaluation. We provide a few examples from our 244 validation and test logs in Figure \ref{fig:testsetexamples}. Statistics of the train, validation, and tests split are described in Table \ref{tab:testsetstatistics}. %We separate 10\% of the training data into the held-out synthetic validation split.

\begin{table}[!t]
    \centering
    %\vspace{-em}
    \caption{We describe the statistics of the \emph{TbV-1.0} dataset. We also provide statistics regarding map deviation data in our validation and test splits, and the types of deviations we observe. We define each BEV frame as a pose where the egovehicle has moved at least 5 meters since the previous pose. Lane geometry changes extend over far more frames than crosswalk changes. }
    %\vspace{-1em}

\begin{adjustbox}{max height=24mm}
\begingroup
\begin{tabular}{llll | c | cccc}
& \multicolumn{3}{c}{ \textsc{\textbf{TbV Data Split}}} & \textsc{\textbf{TbV}} & & \textsc{nuScenes} & \textsc{Argoverse} & \textsc{Waymo} \\
                                              & \textsc{Train}   & \textsc{Val}    & \textsc{Test}   & \textsc{Total}  & & &  1.0 & \textsc{Open}   \\
\toprule
\textsc{Num. Logs} &  799 & 111 & 133 & 1043 & & 1000 & 113 & 1150 \\
\textsc{Num. Images}                                    & 6.3M    & 0.7M  & 0.9M   & 7.8M   & & 1.4M & 490K & 1M  \\
\textsc{Avg. Number of images per log (@20 Hz)}        & 7.8K    & 6.4K   & 6.6K   & -      & &      & -   & -     \\
\textsc{Num. LiDAR sweeps}                             & 446.6K  & 50.7K  & 62.2K  & 559.4K & & 400K & 44K & 230K \\
\textsc{Avg. Number of LiDAR sweeps per log (@10 Hz)}  & 559     & 457    & 467    & -      & &  -   & -   & -  \\
\textsc{Avg. Log Duration}                                 & 55.9 s  & 45.7 s & 46.7 s & -      & & 20 s    & -   & 20 s       \\
\textsc{Hours of Driving (ego-vehicle)}                & 12.4 h  & 1.4 h  & 1.7 h  & 15.5 h & & 5.5 h  & 0.6 h & 6.4 h      \\
\textsc{Miles of Driving (ego-vehicle)}                & 145.6   & 14.7   & 19.7   & 180.0  & & -    & -   & -     \\
\textsc{Kilometers of Driving (ego-vehicle)}           & 234.4   & 23.7   & 31.7   & 289.8  & & -    & -   & -   \\
\arrayrulecolor{LightGrayForTableRule}
\hline
\arrayrulecolor{Black}
\textsc{Num. BEV Frames (once every 5 m of trajectory)} & 43.8K   & 4.5K   & 5.9K   & 54.2K & & -     & -   & - \\    
\arrayrulecolor{LightGrayForTableRule}
\hline
\arrayrulecolor{Black}
\textsc{\# BEV Frames w/ No Map Changes}  & 43.8K & 2.7K & 3.9K & 50.4K & & \textsc{All} & \textsc{All} & \textsc{All} \\
\textsc{\# BEV Frames w/ Any Map Changes} & 0     & 1.8K & 2.0K & 3.8K  & & 0 & 0 & 0 \\
\arrayrulecolor{LightGrayForTableRule}
\hline
\arrayrulecolor{Black}
\textsc{\# BEV Frames w/ Crosswalk Deletion Only}                   & 0 & 57   & 158  & 215  & & 0 & 0 & 0 \\
\textsc{\# BEV Frames w/ Crosswalk Addition Only}                   & 0 & 147  & 75   & 222  & & 0 & 0 & 0 \\
\textsc{\# BEV Frames w/ Lane Geometry Change Only}                 & 0 & 1523 & 1645 & 3168 & & 0 & 0 & 0 \\
\textsc{\# BEV Frames w/ Lane Geometry Change + Crosswalk Deletion} & 0 & 34   & 54   & 88   & & 0 & 0 & 0 \\
\textsc{\# BEV Frames w/ Lane Geometry Change + Crosswalk Addition} & 0 & 43   & 64   & 107  & & 0 & 0 & 0 \\
\bottomrule
\end{tabular}
\endgroup
\end{adjustbox}
\label{tab:testsetstatistics}
\end{table}

%\subsection{Temporal Analysis of Changes}
%
%In Figure \ref{fig:datasettemporalaxis}, we show a visual representation of the TbV dataset along the temporal axis.
%
%%  trim={<left> <lower> <right> <upper>}
%\begin{figure}[!h]
%\centering
%\includegraphics[width=0.65\columnwidth]{figs/dataset_temporal_axis.pdf}
%%\includegraphics[width=0.4\columnwidth,trim={1cm 0 0 1cm},clip]{figs/dataset_temporal_axis_finegrainedchanges.pdf}
%\caption{We show what portions of the logs are driving through changed parts of the world. Red represents timestamps where the AV encounters changes, and green represents alignment between the real and represented world.}
%\label{fig:datasettemporalaxis}
%\end{figure}

\begin{table}[!b]
    \centering
    \caption{Probability of a 30m $\times$ 30m region that has been visited at least 5 times in 5 months undergoing a lane geometry or crosswalk change within the same time period. These statistics apply
    only to surface-level urban streets, not highways.}
    %\vspace{-0.5em}
    \begin{adjustbox}{max width=\columnwidth}
	\begingroup
    \begin{tabular}{l ccc ccc}
        \toprule
        & \multicolumn{6}{c}{\textsc{\textbf{City Name}}} \\
         %& \textsc{PIT} & \textsc{DTW} & \textsc{WDC} & \textsc{MIA} & \textsc{ATX} & \textsc{PAO} \\
         & \textsc{Pittsburgh} & \textsc{Detroit} & \textsc{Washington, D.C.} & \textsc{Miami} & \textsc{Austin} & \textsc{Palo Alto} \\
        \midrule
        \textsc{Probability of Change} & 0.0068 & 0.0056 & 0.0046 & 0.0038 & 0.0009 & 0.0007 \\
\arrayrulecolor{LightGrayForTableRule}
\hline
\arrayrulecolor{Black}
        Up to \textsc{T / 1000 tiles}  & \multirow{2}{*}{7}  & \multirow{2}{*}{6}  & \multirow{2}{*}{5}  & \multirow{2}{*}{4}  & \multirow{2}{*}{0.9} & \multirow{2}{*}{0.7} \\
        \textsc{will change in 5 mo.} & & & & & &  \\
    \bottomrule
    \end{tabular}
    \endgroup
    \end{adjustbox}
    \vspace{-1.5em}
    \label{tab:city-tile-spatialchange-prob}
\end{table}

\subsection{Annotation}
\label{sec:annotation-and-change-freq}
In order to label map changes, we use three rounds of human filtering, where changes are identified, confirmed, and characterized by three independent reviewing panels. We assign spatial coordinates to each changed object within a city coordinate system. Crosswalk changes are denoted by a polygon, and lane geometry changes by polylines. We use egovehicle-to-changed-map-entity distance (point-to-polygon or point-to-line) to determine whether or not a sensor and map rendering should be classified as a map-sensor match or mismatch. % change of a particular type. In other words, we use these distances to classify which frames in a test sequence are map-sensor matches or mismatches.

\paragraph{Analysis of Map Change Frequency} We use our annotated map changes, along with 5 months of fleet data, to analyze the frequency of map changes on a city-scale across several cities. Two particular questions are of interest: (1) \emph{how often will an autonomous vehicle encounter a map change as part of its day-to-day operation?} and (2) \emph{what percentage of map elements in a city will change each month or each year}? For our analysis, we subdivide a city's map into square spatial units of dimension 30 meters $\times$ 30 meters, often referred to as ``tiles'' in the mapping community. We find the probability $p$ of an encounter at any given time with a tile with changed lane geometry or crosswalk to be $p \approx 5.5174 \times 10^{-5}$. Although the probability of a single event is low, we cannot ignore rare events, as doing so would be reckless. Given the 3.225 trillion miles driven in the U.S. per year \cite{UsDot19pr_DrivingMoreThanEverBefore}, this could amount to \emph{billions} of such encounters per year. We determine that up to 7 of every 1000 map tiles may change in a 5-month span (see Table \ref{tab:city-tile-spatialchange-prob}), a significant number. More details are provided in the Appendix.

% We use a series of validation steps from three separate teams of expert annotators to ensure . First, a team of expert annotators annotates each vehicle log for potential map changes. Afterwards, a separate, second team of expert annotators verifies each annotation from round 1. Finally, the co-authors,  manually inspect each annotation and

\subsection{Sensor Data}
Our TbV dataset includes LiDAR sweeps collected at 10 Hz, along with 20 fps imagery from 7 cameras positioned to provide a fully panoramic field of view. In addition, camera intrinsics, extrinsics and 6 d.o.f. AV pose in a global coordinate system are provided. LiDAR returns are captured by two 32-beam LiDARs, spinning at 10 Hz in the same direction (``in phase''), but separated in time by $180^\circ$. The cameras trigger in-sync with both of them, leading to a 20 Hz framerate. 
The 7 global shutter cameras are synchronized to the LiDAR to have their exposure centered on when the LiDAR sweeps through the middle of their fields of view. The top LiDAR spins clockwise in its frame, while the bottom LiDAR spins counter-clockwise in its frame; in the ego-vehicle frame, they both spin clockwise.

%Additional sensor details are omitted here for the sake of anonymity, but will be included in the final paper.
% Each LiDAR has 32 beams. The two LIDARs spin 180 degrees apart from one another at 10 Hz, and the cameras trigger in sync with both of them, leading to the 20 Hz framerate. \textcolor{red}{TODO: This may be a trade-secret}.
%\textcolor{red}{stereo pair? how are the cameras laid out exactly? FOV?  }

\subsection{Map Data}
In the Appendix, we list the semantic map entities we include in the TbV dataset. Previous AV datasets have released sensor data localized within a single map per city \cite{Chang19cvpr_Argoverse,Caesar20cvpr_nuscenes,Houston20_LyftLevel5}. This is not a viable solution for TbV, since the maps change over our long period of data gathering. We instead release local maps with all semantic entities within 100 meters of the egovehicle featured. Accordingly, single, incremental changes can be identified and tested. We release many maps, one per vehicle log; the corresponding map is the map used on-board at time of capture. Lane segments within our map are annotated with boundary annotations for both the right and left marking (including against curbs) and are marked as implicit if there is no corresponding paint. We use the same release format for our maps as Argoverse 2.0 \cite{Wilson21neurips_Argoverse2} uses. %Accordingly, we include annotations of any paint that occurs close to the curb, which can produce some of the most subtle types of map changes. we include sensor data where different maps are valid at separate times.

\subsection{Dataset Taxonomy}
Our dataset's taxonomy is intentionally oriented towards lane geometry and crosswalk changes. In general, we focus on permanent changes, which are far less frequent in urban areas than temporary map changes. Temporary map changes often arise due to construction and road blockades. % (which we find to be fairly common in urban areas). %. While construction is extremely common in urban areas, permanent map changes are less frequent. While the line is at times blurry between the two, we associate temporary changes with temporary construction and road blockades. 
% Make the approach the most important part, not the dataset.
%, that within days could be resolved? 

We postulate that temporary map changes -- temporarily closed lanes or roads, or temporary lanes indicated by barriers or cones, should be relegated to onboard object recognition and detection systems. Indeed, recent datasets such as nuScenes \cite{Caesar20cvpr_nuscenes} include 3d labeling for traffic cones and movable road barriers, such as Jersey barriers (see Appendix for examples). Even certain types of permanent changes are object-centered (e.g. changes to traffic signs). Accordingly, a natural division arises between ``things'' and ``stuff' in map change detection, just as in general scene understanding \cite{Adelson91book_PlenopticFn,Caesar18cvpr_COCOStuff}. We focus on the ``stuff'' aspect, corresponding to entities which are often spatially distributed in the BEV; we find lane geometry and crosswalk changes to be more frequent than other ``stuff''-related changes.

% (see Figure \ref{fig:dynamicobjs})
%\subfloat[img 1]{
%\includegraphics[width=0.25\linewidth]{figs/dynamic_obj_examples/dynamic_obj_example_road_barricade_79ApelcEgWdX7Rklwq5zJ3pZhBZX7Agj__2020-08-31-Z1F0069.png}
%}
%\subfloat[img 1]{
%\includegraphics[width=0.25\linewidth]{figs/dynamic_obj_examples/dynamic_obj_example_caterpillar1_WgTXheD91ksILruPxaHM0nywcZz4IiCJ__2020-08-27-Z1F0076.png}
%}
%\subfloat[img 1]{
%\includegraphics[width=0.25\linewidth]{figs/dynamic_obj_examples/dynamic_obj_example_caterpillar2_WgTXheD91ksILruPxaHM0nywcZz4IiCJ__2020-08-27-Z1F0076.png}
%}

 %training pair is required. Cannot use a single base-map.

\section{Approach}

\subsection{Learning Formulation}
We formulate the learning problem as predicting whether a map is stale by fusing local HD map representations and incoming sensor data.  We assume accurate pose is known. At training time, we assume access to training examples in the form of triplets $(x,x^*, y)$, where $x$ is a local region of the map, $x^*$ is an online sensor sweep, and $y$ is a binary label suggesting whether a ``significant'' map change occurred. $(x,x^*)$ should be captured in the same location.
% Proximity in the embedding space indicates agreement between the two data streams.

%One could think of the learning process as a minimax two-player game \cite{Goodfellow14nips_GAN}, although our generator consists of hand-crafted heuristics, and only our discriminator is learned from data. Rather than discriminating between real and synthesized images, the discriminator learns to recognize sensor-map alignment and misalignment.

\begin{figure}[]
\centering
\vspace{-2em}
\subfloat[Early Fusion (Sensor + Map)]{
\includegraphics[width=0.20\columnwidth]{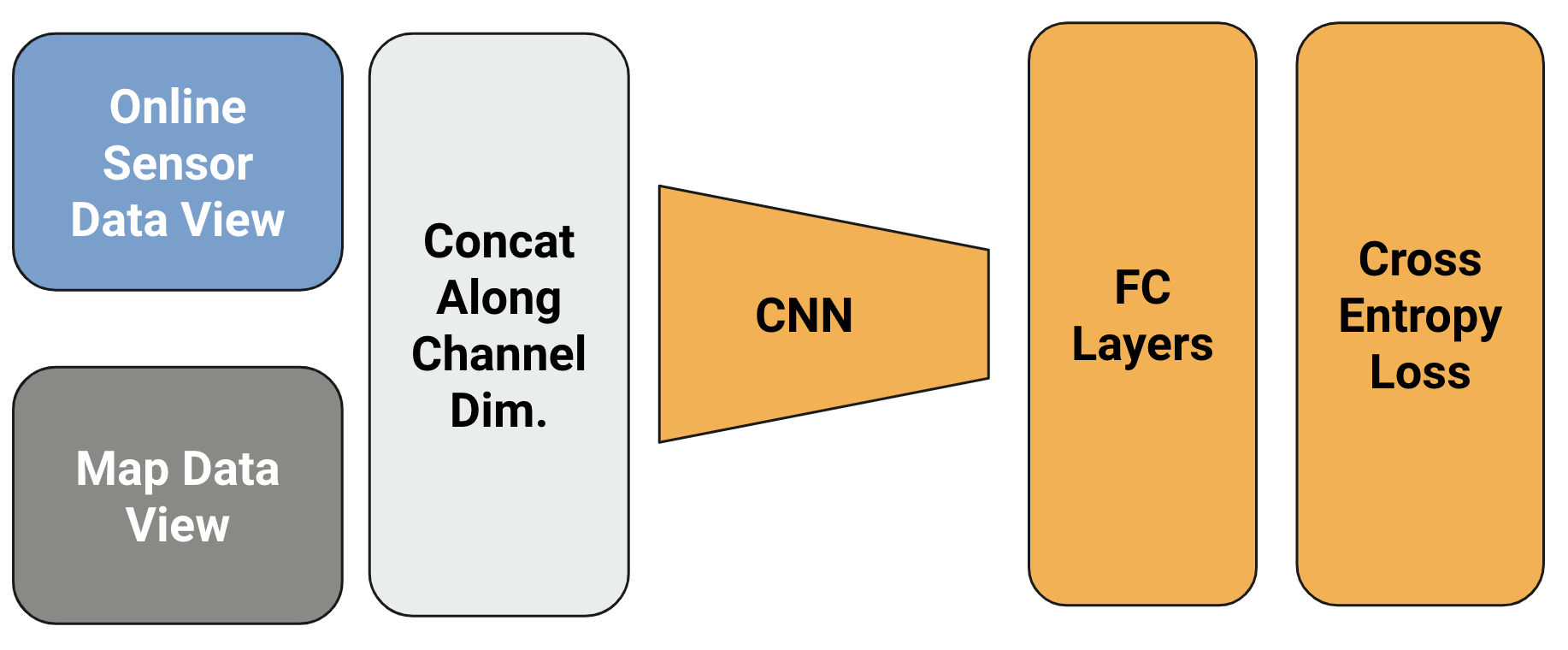}
}
\hspace{3mm}
\subfloat[Early Fusion (Sensor + Semantics + Map)]{
\includegraphics[width=0.20\columnwidth]{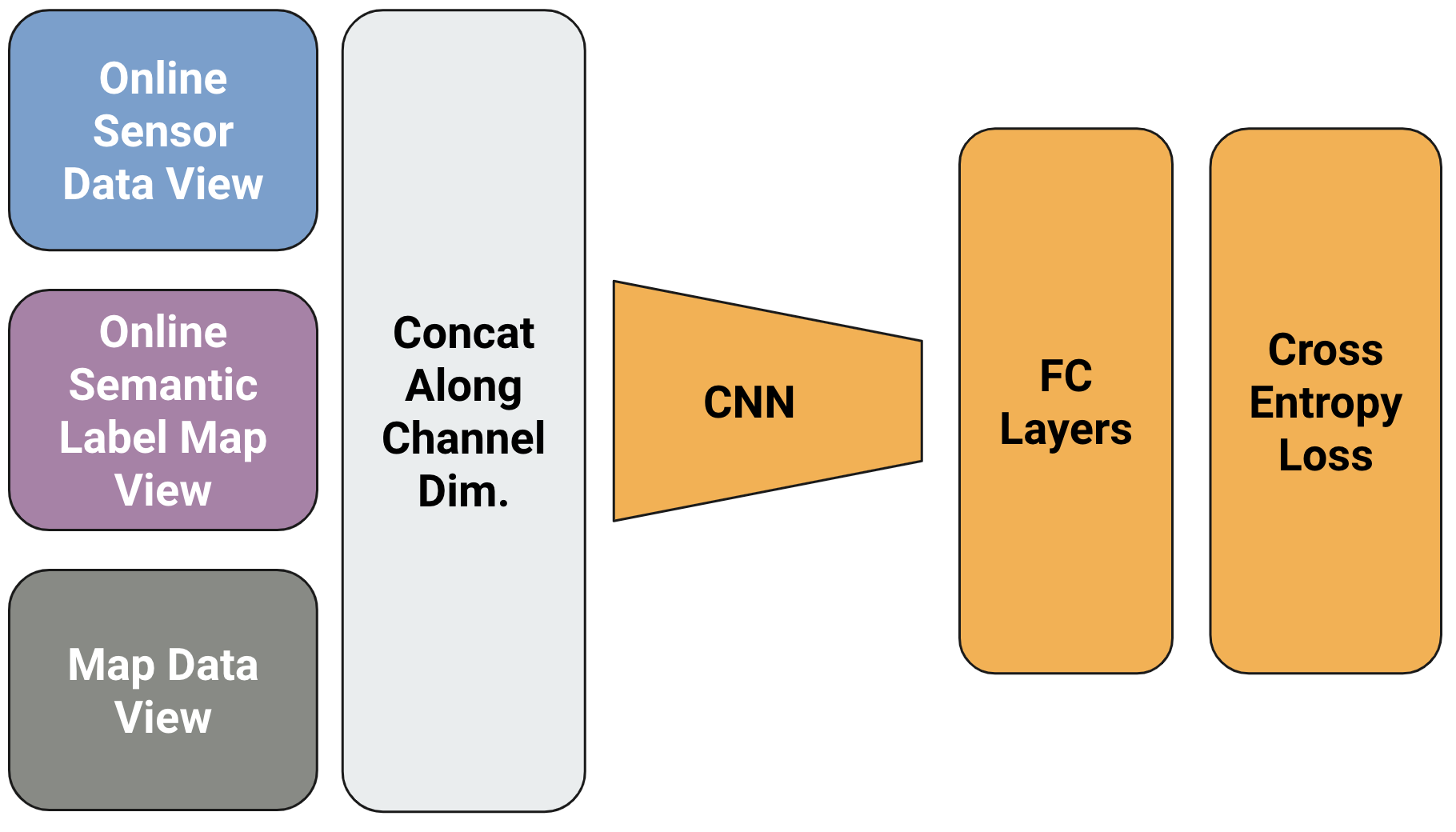}
}
\hspace{3mm}
\subfloat[Late Fusion (Siamese)]{
\includegraphics[width=0.20\columnwidth]{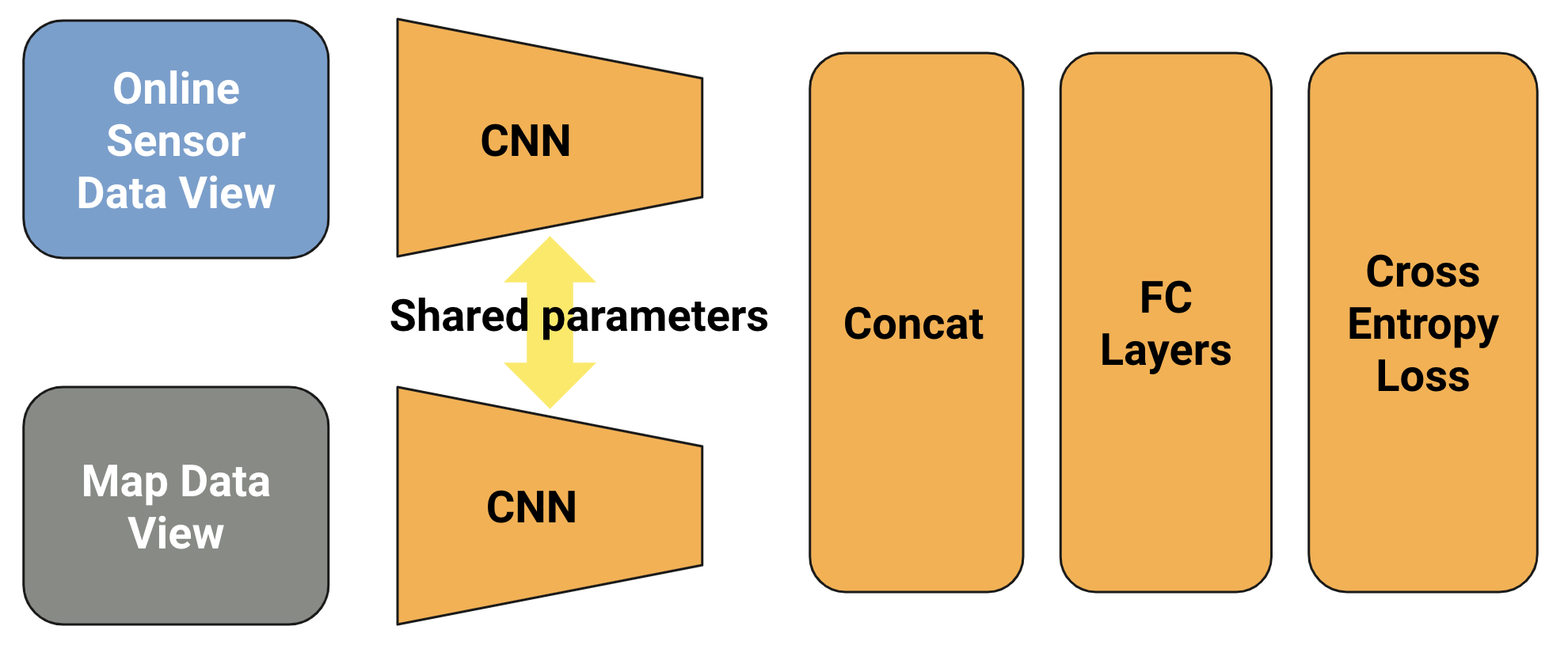}
}
\hspace{3mm}
\subfloat[Map-Only Input]{
\includegraphics[width=0.20\columnwidth]{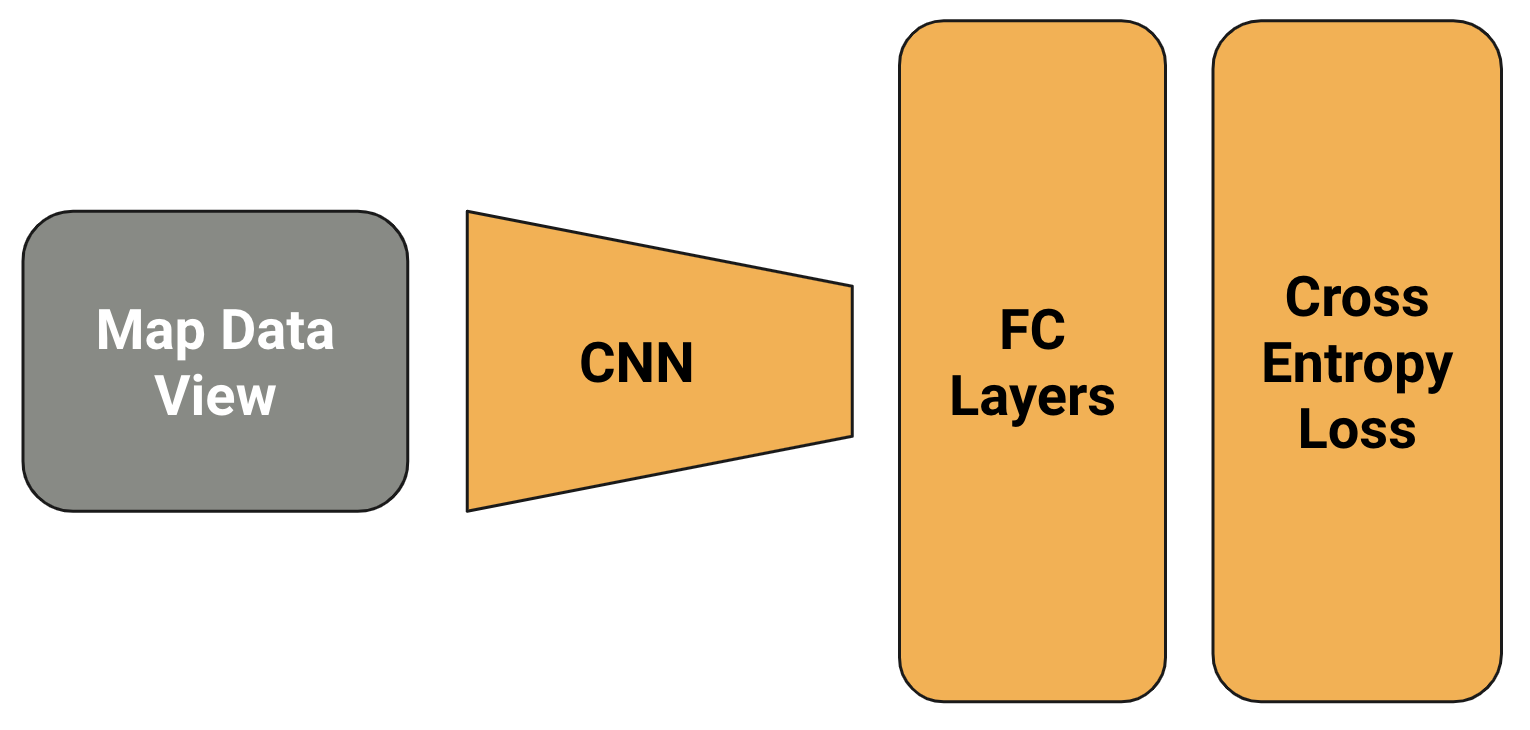}
}
\caption{Learning architectures we explore for the map change detection problem.}
\label{fig:learningarchitectures}
% \vspace{-1.2em}
\end{figure}

We explore a number of architectures to learn a shared map-sensor representation, including early fusion and late fusion (see Figure \ref{fig:learningarchitectures}). The late fusion model uses a siamese network architecture  with two input towers, and then a sequence of fully connected layers. We utilize a two-stream architecture \cite{Han15cvpr_MatchNet, Zbontar15cvpr_matchingcost} with shared parameters, which has been shown to still be effective even for multi-modal input \cite{Lambert18cvpr_DLUPI}. %Although triplet or contrastive metric learning is possible, we find a cross-entropy loss formulation to be already useful. 
We also explore an early-fusion architecture, where the map, sensor, and/or semantic segmentation data are immediately concatenated along the channel dimension before being fed to the network. We take no credit for these convnet architectures, which are well studied.

\begin{figure*}
  \vspace{-1em}
  \centering
  \subfloat[Lane marking color is changed from solid white to implicit (see bottom-center of image)]{
  \includegraphics[width=0.47\columnwidth]{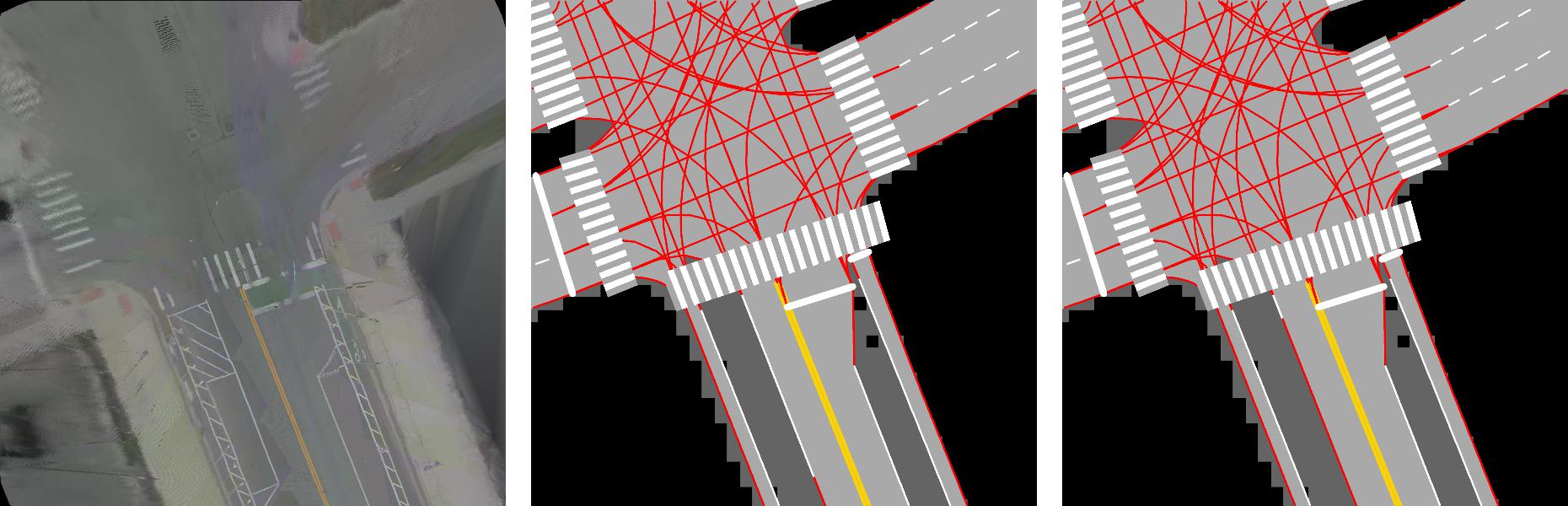}
  }
  \vspace{-2mm}
  \hspace{2mm}
  \subfloat[A crosswalk is deleted from the map. Reflections off of windows create illumination variation on the road surface.]{
  \includegraphics[width=0.47\columnwidth]{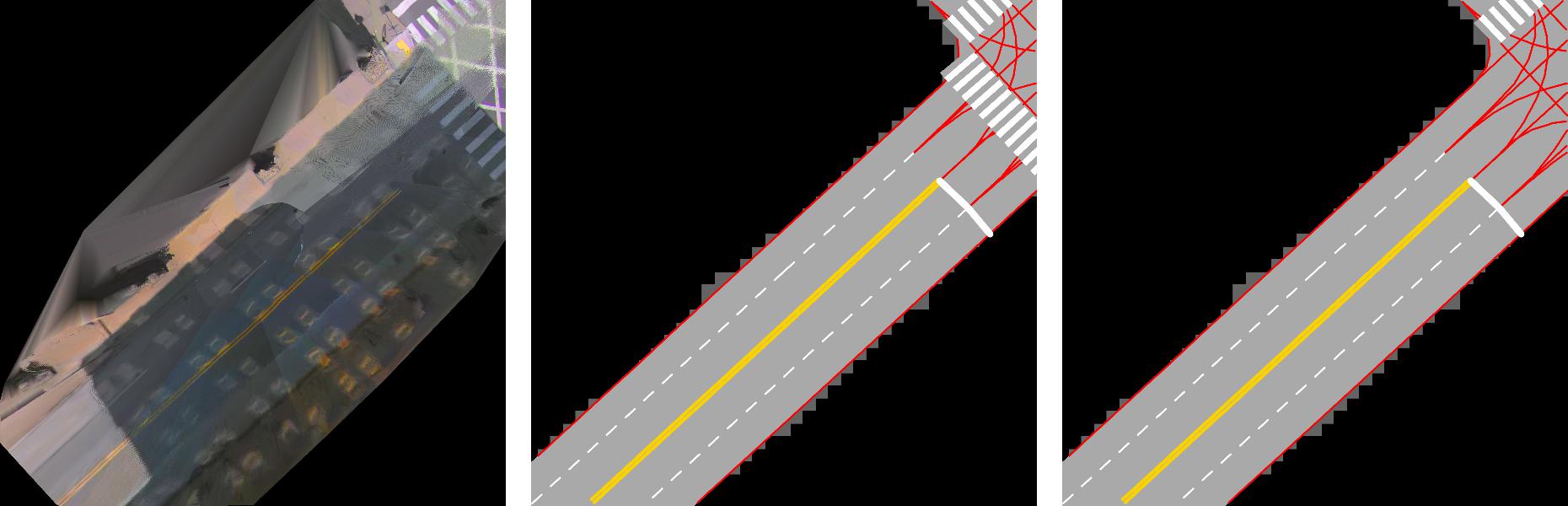}
  }
  %\vspace{-2mm}
  \hspace{2mm}
  \subfloat[A bike lane is added to the map (see center-right of image)]{
  \includegraphics[width=0.47\columnwidth]{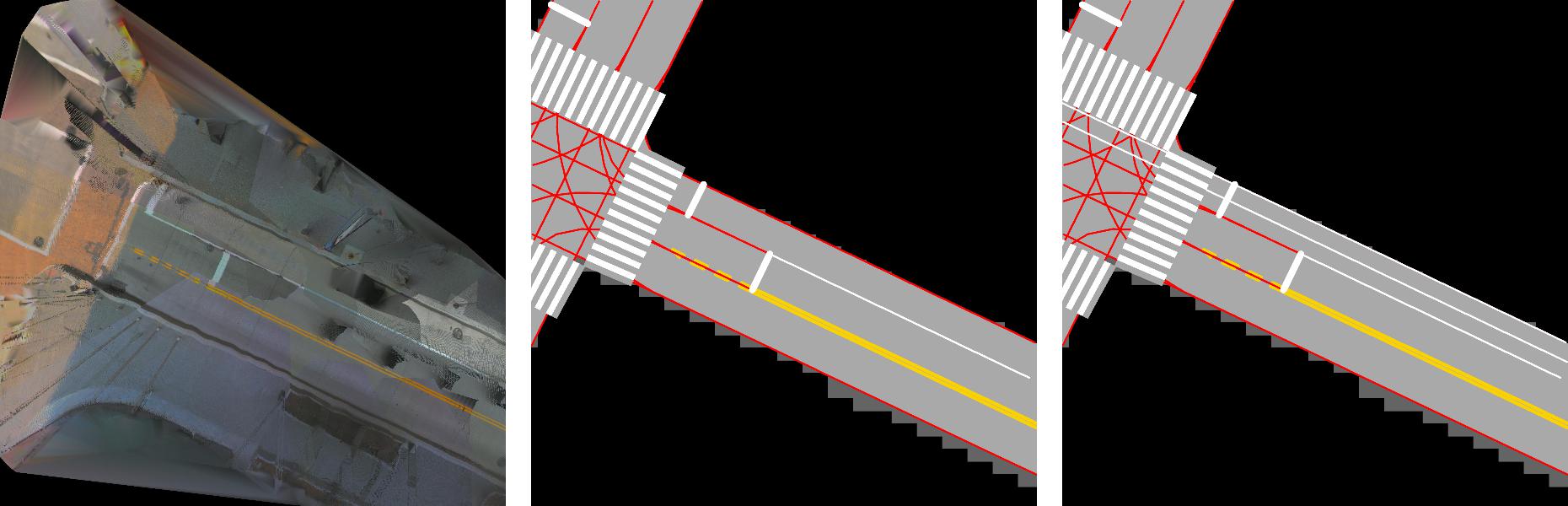}
  }
  \vspace{-2mm}
  \hspace{2mm}
  \subfloat[The structure of a lane boundary marking is changed, from double-solid yellow to single-solid yellow (see bottom-center of image). Its color is preserved.]{
  \includegraphics[width=0.47\columnwidth]{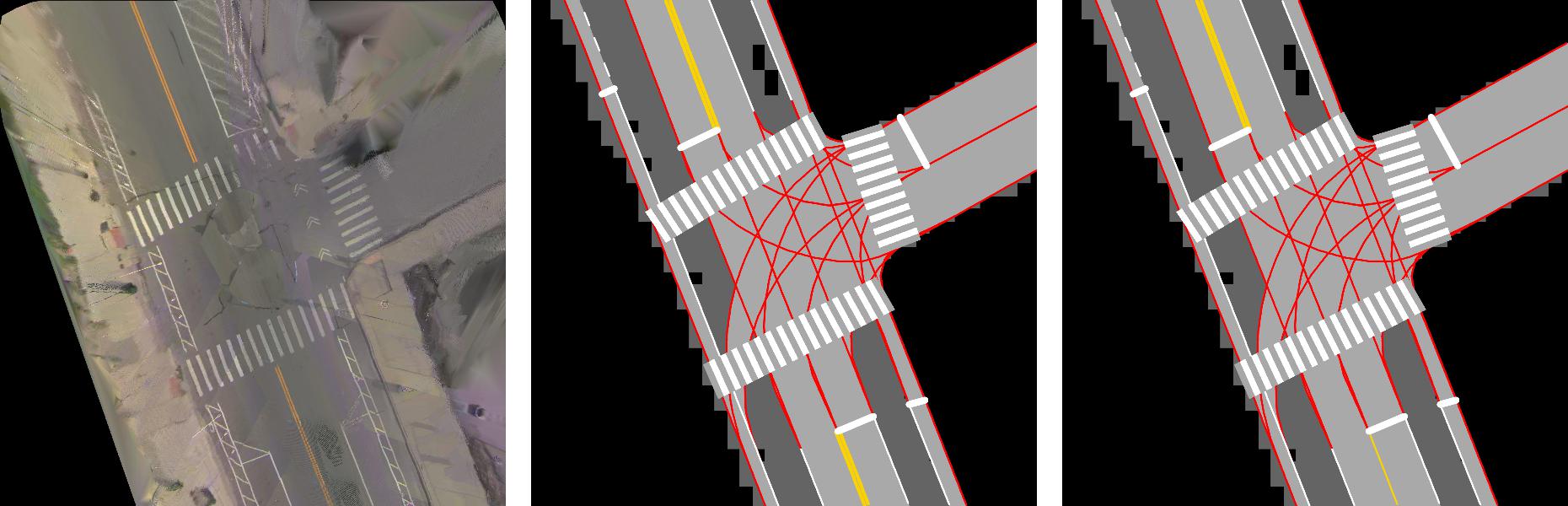}
  }
  %\vspace{-2mm}
  \hspace{2mm}
  \subfloat[A crosswalk is synthetically inserted into the map.]{
  \includegraphics[width=0.47\columnwidth]{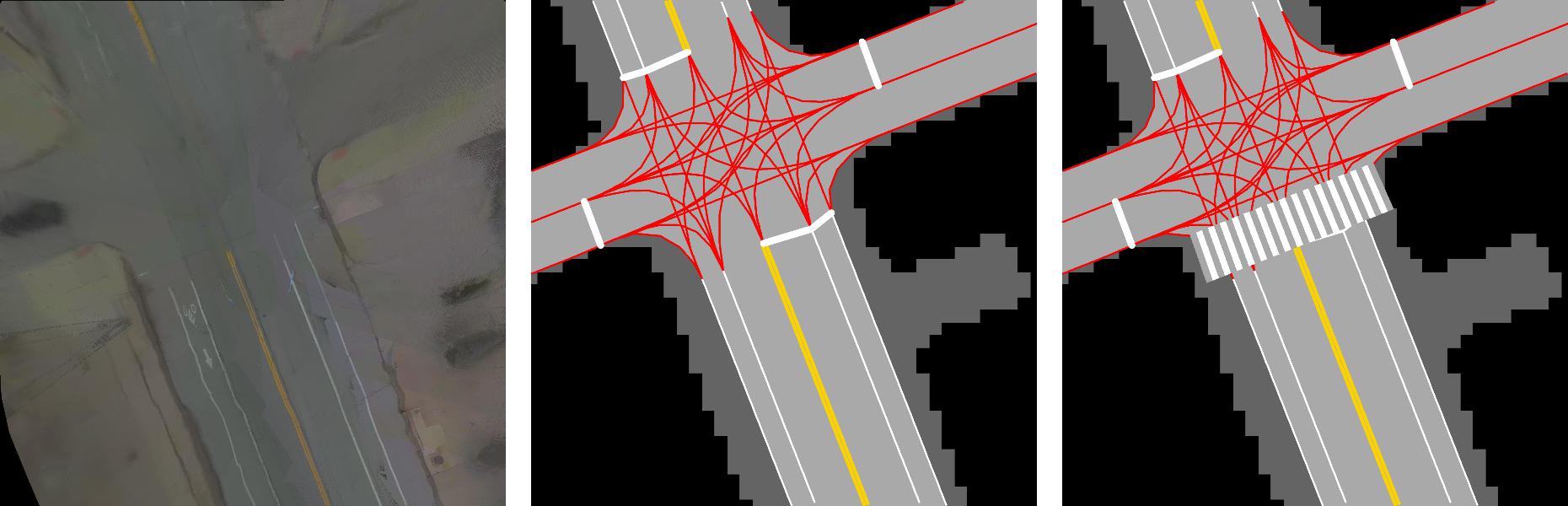}
  }
  %\vspace{-2mm}
  \hspace{2mm}
  \subfloat[A solid white lane boundary marking is deleted (see top-center of image).]{
  \includegraphics[width=0.47\columnwidth]{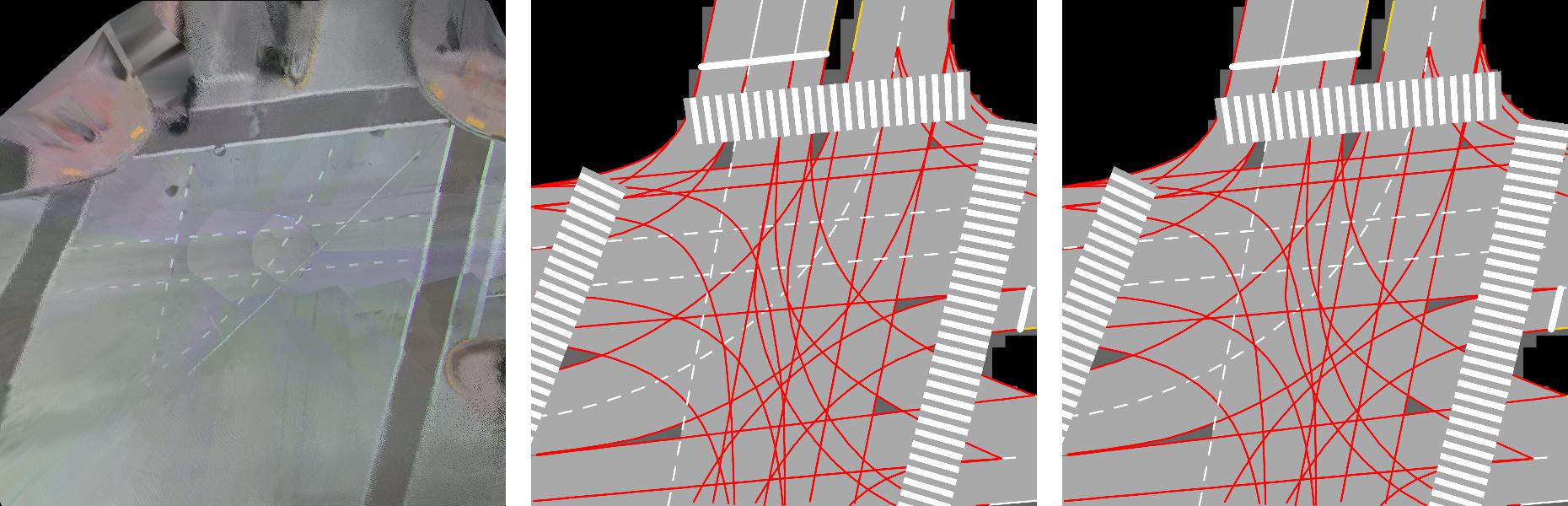}
  }
  \caption{Examples of our 6 types of synthetic map changes (zoom in for detail). Each row represents a single scene. \emph{Left:} bird's eye view (BEV) sensor data representation. \emph{Center:} rasterized onboard map representation (positive). \emph{Right:} synthetic perturbation of onboard map (negative). We use red to denote implicit lane boundaries. }
  \label{fig:6-types-synthetic-changes}
  \vspace{-1em}
  \end{figure*}

\subsection{Synthesis of Mismatched Data}
\label{sec:tbv-synthesis-mismatched-data}
Real negatives are difficult to obtain; because their location is difficult to predict a priori, they cannot be captured in a deterministic way by driving around an urban area on any particular day. Therefore, rather than using real negatives for training, we synthesize fake negatives. While sensor data is difficult to simulate, requiring synthesis of sensor measurements from the natural image manifold \cite{Yang20cvpr_SurfelGAN,Chen21cvpr_GeoSim}, manipulating vector maps is relatively straightforward.\\

 Synthetic data generation via randomized rendering pipelines can be highly effective for synthetic-to-real transfer \cite{Shotton11cvpr_RealtimePose}.
 In order to synthesize fake negatives from true positives, one must be able to trust the fidelity of labeled true positives. In other words, one must trust that for true positive logs, the map is completely accurate for the corresponding sensor data. %We use a fleet of operators to attest to this fidelity, and then Once we have verified this fidelity, 
We perturb the data in a number of ways (See Appendix). Once such fidelity is confirmed and assured, vector map manipulation is trivial because map elements are vector entities which can be perturbed, deleted, or added.  \\

While synthesizing random vector elements is trivial, sampling from a realistic distribution requires conformance to priors, including the lane graph, drivable area, and intersection. We aim for synthetic map/sensor deviations to resemble real world deviations, and real world deviations tend to be subtle, e.g. a single lane is removed or painted a different color, or a single crosswalk is added, while 90\% of the scene is still a match. In order to generate realistic-appearing synthetic map objects, we hand-design a number of priors that must be respected for a perturbed example to enter our training set as a valid training example (see Appendix). Figure \ref{fig:6-types-synthetic-changes} and Table \ref{tab:perturbationstatistics} of the Appendix enumerate a full list of the 6 types of synthetic changes we employ.
%Please refer to the Supplementary Material for a detailed discussion of our synthesis process and of the priors we employ.
 
%We render lane boundary markings as colored polylines; we use red to denote implicit boundaries, and yellow and white for lane markings of their respective color. Lane boundary markings are deleted by simply not rendering them in the rasterized image. 

\subsection{Sensor Data Representation}
\label{sec:tbv-sensor-data-repr}
We experiment with two sensor data representations -- \emph{ego-view} (the front center camera image) and \emph{bird's eye view} (BEV). Rather than using Inverse Perspective Mapping (IPM) \cite{Geiger09ivs_MonocularRoadMosaicing,Zhang14icrb_LaneLevelOrthophoto,Rapo18thesis_orthoimagery}, we generate the BEV representation (i.e. orthoimagery) by ray-casting image pixels to a ground surface triangle mesh. %While I is common in the literature for orthoimagery generation \cite{Geiger09ivs_MonocularRoadMosaicing,Zhang14icrb_LaneLevelOrthophoto,Rapo18thesis_orthoimagery}, it requires a planar ground assumption leading to significant distortion. 
For ray-casting, we use a set of camera sensors with a panoramic field of view, mounted onboard an autonomous vehicle. The temporal nature of the data is exploited as pixel values from 7 streams of ego-view images are aggregated to render each BEV image in order to reduce sparsity (see Appendix). 70 images, capturing 10 timesteps from each of 7 frustums, are used to create each rendering.

\subsection{Map Data Learning Representation}
We render our map inputs as rasterized images; Entities are layered from the back of the raster to the front in the following order: driveable area, lane segment polygons, lane boundaries, pedestrian crossings (i.e. crosswalks). We  release the API to generate and render these map images. Vector map entities are synthetically perturbed before rasterization for synthetic negative examples.

% We provide several thousand vehicle scenarios from vehicle logs, along with corresponding maps. Taxonomy is as follows, with Table XXX showing the frequency in our 

% In this section, we present two simple baselines to detect that an AV has encountered a scenario unlike its map.
% If map deviations are heavily skewed towards one particular type, simple baselines may be surprisingly effective. For example, if orange cones representing closed lanes represent the principal deviation, one could simply train an  object detection or segmentation network to recognize orange traffic cones, backproject the image-space pixels associated with such detections into the BEV scene, and verify whether they fall within a lane's area. If so, a change is nearly guaranteed.

%Each input image is a 31-channel BEV image consisting of reflectance and 30 z-slice occupancy channels is fed to a network. The cross-entropy loss can be used for training.

% fake map deviation

\section{Experimental Results}

We frame the map change detection task as follows: \emph{at timestamp} $t$, \emph{given a buffer of all past sensor data, including camera intrinsics and extrinsics, along with 6 d.o.f. egovehicle poses} ${}^{city}T_{egovehicle}$ which we denote as $T_{i=0\dots t}$, \emph{image data} $\{I_{i=0\dots t}^c\}_{c=1}^C$ \emph{ where} $c$ \emph{is a camera index}, \emph{lidar sweeps} $L_{i=0 \dots t}$, \emph{onboard map data}  $M$, \emph{estimate whether the map is in agreement with the sensor data}.

\subsection{Implementation Details}
\label{sec:impl-details}

\paragraph{Ego-view Models.} Our ego-view models that operate on front-center camera images leverage both LiDAR and RGB sensor imagery.  We use LiDAR information to filter out occluded map elements from the rendering. We linearly interpolate a dense depth map from sparse LiDAR measurements, and then compare the depth of individual map elements against the interpolated depth map; elements found behind the depth map are not rendered. In our early fusion architecture, we experiment with models that also have access to semantic label masks from the semantic head of a publicly-available \emph{seamseg} ResNet-50 panoptic segmentation model \cite{Porzi19cvpr_Seamseg}. For those models with access to the semantic label map modality, we append $224 \times 224$ binary masks for 5 semantic classes (`road', `bike- lane', `marking-crosswalk-zebra', `lane-marking-general', and `crosswalk-plain') as additional channels in early fusion.

\paragraph{Bird's Eye View Models.} We also implement camera-based models that accept orthoimagery as input, relying only upon RGB imagery and a pre-generated ground height field, without utilizing online LiDAR. We generate new orthoimagery each time the ego-vehicle moves at least 5 meters, and use each orthoimagery example as a training or test example.  We use 2 cm per pixel resolution for orthoimagery; all pixels corresponding to 3d points up to 20 meters in $\ell_{\infty}$-norm from the ego-vehicle are included, generating a $2000 \times 2000$ px image.

\paragraph{Training.} We use a ResNet-18 or ResNet-50 \cite{He16cvpr_ResNet} backbone, with ImageNet-pretrained weights, where a corresponding weight parameter's size is applicable. We use a crop size of $224 \times 224$ from images resized to $234 \times 234$ px. Please refer to the Appendix for additional implementation details and an ablation experiment on the influence of input crop size on performance.

\subsection{Evaluation}
We report results on an earlier, beta version of TbV, which we call \emph{TbV-Beta}, which includes slightly fewer logs. The publicly released version of TbV is \emph{TbV-1.0}.\\

Comparable evaluation of the ego-view and bird's eye view models is challenging since they operate on different portions of the scene. The ego-view model should not be penalized for ignoring changes outside of its field of view, especially those located behind the ego-vehicle. Thus, we provide results for visibility-based evaluation (when the change is visible in the ego-view), and a purely proximity-based comparison (when it is within 20 m. by $\ell_\infty$ norm). The area about which a model should reason is somewhat arbitrary; changes behind and to the side may matter for fleet operation, but changes directly ahead of the AV are arguably most important for path-planning \cite{Philion20cvpr_PlannerCentric}. In addition, changes visible to the rear at some timestamp are often visible directly in front of the AV at a prior timestamp. We consider the visibility-based evaluation to be most fair for ego-view models.\\

We use a mean of per-class accuracies to measure performance on a two-class problem: predicting whether the real world is changed (i.e. map and sensor data are mismatched), or unchanged (i.e. a valid match). This accounts for both precision and recall. If a confusion matrix is computed with predicted entries on the rows and actual classes as the columns, and normalized by dividing by the sum of each column, 2-class accuracy can be simply calculated as the mean of the diagonal of the confusion matrix. More formally, let $n_{cl}=2$ be the number of classes, $\hat{y}_i$ be the prediction for the $i$'th test example, and $y_i$ be the ground truth label for the $i$'th test example. We define per-class accuracy ($\text{Acc}_c$) and mean accuracy (mAcc) as:
\vspace{-4mm}
\footnotesize
\begin{equation}
\text{mAcc} = \nicefrac{1}{n_{cl}} \sum\limits_{c=0}^{n_{cl}}	 Acc_{c}, \hspace{3mm} \text{Acc}_c = \frac{\sum\limits_{i=0}^N \mathbbm{1}\{ \hat{y}_i = y_i\} \cdot \mathbbm{1}\{ y_i = c \}  }{\sum\limits_{i=0}^N \mathbbm{1}\{y_i = c \} } 
\end{equation}
\normalsize
\vspace{-4mm}

\begin{table*}[]
    %\vspace{-5em}
    \caption{Controlled evaluation of the influence of fusion architecture and scene rendering viewpoint (ego-view vs. BEV), on \emph{TbV-Beta}.  Rows marked with an asterisk represent an expected mean accuracy based on randomly flipped labels, rather than results from a trained model.}
    \centering
    %\vspace{-1em}
    \begin{adjustbox}{max width=\columnwidth} % ICCV 2\columnwidth
    \begingroup
    \begin{tabular}{llc ccc c ccc ccc}
        \toprule
        &    &     & \multicolumn{3}{|c|}{\textsc{Modalities}}    & \textsc{Visibility-based} & \multicolumn{3}{|c|}{\textsc{BEV proximity}}     & \multicolumn{3}{|c}{\textsc{Visibility-based}}    \\
        &    &     & \multicolumn{3}{|c|}{}   & \textsc{Eval. @ 20m}    & \multicolumn{3}{|c|}{\textsc{Eval. @20m}}     & \multicolumn{3}{|c}{\textsc{Eval. @20m}}    \\
 \textsc{Backbone} & \textsc{Arch.} & \textsc{Viewpoint} & \multicolumn{1}{|c}{\textsc{RGB}} & \textsc{Semantics} & \multicolumn{1}{c|}{\textsc{Map}} & \textsc{val}        & \multicolumn{1}{|c}{\textsc{test}} &      \textsc{Is Changed}     & \textsc{No Change}     & \multicolumn{1}{|c}{\textsc{test}}       & \textsc{Is Changed}     & \textsc{No Change} \\
                   &               &                    &  \multicolumn{1}{|c}{ }            &                   &  \multicolumn{1}{c|}{ }            & \textsc{mAcc}       & \multicolumn{1}{|c}{\textsc{mAcc}} &      \textsc{Acc}            & \textsc{Acc}           & \multicolumn{1}{|c}{\textsc{mAcc}}       & \textsc{Acc}            & \textsc{Acc}     \\ 
    \midrule
    -         & \textsc{Random Chance}* & -         & -          & -          & -          & 0.5000         & 0.5000          & 0.50                 & 0.50                & 0.5000                  & 0.50                         & 0.50                        \\
\arrayrulecolor{LightGrayForTableRule}
\hline
\arrayrulecolor{Black}
    \textsc{ResNet-18} & \textsc{Early Fusion}  & \textsc{Ego-View}  & \checkmark & \checkmark & \checkmark & 0.8417         & \textbf{0.6724}          & 0.57                 & 0.77                & \textbf{0.7234}                  & \textbf{0.67}                         & 0.78                        \\
    \textsc{ResNet-18} & \textsc{Late Fusion}   & \textsc{Ego-View}  & \checkmark &            & \checkmark & 0.8108         & 0.4930           & 0.13                 & \textbf{0.85}                & 0.4956                  & 0.13                         & \textbf{0.86}                        \\

\arrayrulecolor{LightGrayForTableRule}
\hline
\arrayrulecolor{Black}
    \textsc{ResNet-50} & \textsc{Early Fusion}  & \textsc{BEV}       & \checkmark & \checkmark & \checkmark & \textbf{0.9130}  & \textbf{0.6728}  & \textbf{0.58}                 & 0.77                & -                       & -                            & -                           \\
    \textsc{ResNet-50} & \textsc{Late Fusion}   & \textsc{BEV}       & \checkmark &            & \checkmark & 0.8697         & 0.5761          & 0.43                 & 0.72                & -                       & -                            & -  \\
    \bottomrule  
    \end{tabular}
    \endgroup
    \end{adjustbox}
    \label{tab:renderingviewpoint}         
    \vspace{-1.5em}
    \end{table*}

    \begin{table*}[]
    %\vspace{-1em}
    \caption{Controlled evaluation of the influence of data modalities, on \emph{TbV-Beta}. Rows marked with an asterisk represent an expected mean accuracy based on randomly flipped labels, rather than results from a trained model. }
        \centering
        %\vspace{-1em}
        \begin{adjustbox}{max width=\columnwidth} % ICCV 2\columnwidth
        \begingroup
        \begin{tabular}{llc ccc c ccc ccc}
            \toprule
            &    &     & \multicolumn{3}{|c|}{\textsc{Modalities}}    & \textsc{Visibility-based} & \multicolumn{3}{|c|}{\textsc{BEV proximity}}     & \multicolumn{3}{|c}{\textsc{Visibility-based}}    \\
            &    &     & \multicolumn{3}{|c|}{}   & \textsc{Eval. @ 20m}    & \multicolumn{3}{|c|}{\textsc{Eval. @20m}}     & \multicolumn{3}{|c}{\textsc{Eval. @20m}}    \\
     \textsc{Backbone} & \textsc{Arch.} & \textsc{Viewpoint} & \multicolumn{1}{|c}{\textsc{RGB}} & \textsc{Semantics} & \multicolumn{1}{c|}{\textsc{Map}} & \textsc{val}        & \multicolumn{1}{|c}{\textsc{test}} &      \textsc{Is Changed}     & \textsc{No Change}     & \multicolumn{1}{|c}{\textsc{test}}       & \textsc{Is Changed}     & \textsc{No Change} \\
                       &               &                    &  \multicolumn{1}{|c}{ }            &                   &  \multicolumn{1}{c|}{ }            & \textsc{mAcc}       & \multicolumn{1}{|c}{\textsc{mAcc}} &      \textsc{Acc}            & \textsc{Acc}           & \multicolumn{1}{|c}{\textsc{mAcc}}       & \textsc{Acc}            & \textsc{Acc}     \\ 
        \midrule
        -                   & \textsc{Random Chance}*         & -                   & -          & -          & -          & 0.5000     & 0.5000              & 0.50                     & 0.50                    & 0.5000              & 0.50                     & 0.50                      \\
\arrayrulecolor{LightGrayForTableRule}
\hline
\arrayrulecolor{Black}
        \textsc{ResNet-18 } & \textsc{Single Modality}* & \textsc{Ego-View}            & \checkmark &            &            & 0.5000     & 0.5000              & 0.50                     & 0.50                    & 0.5000              & 0.50                     & 0.50                      \\
        \textsc{ResNet-18 } & \textsc{Single Modality} & \textsc{Ego-View}            &            &            & \checkmark & 0.8444     & 0.5333              & 0.36                     & 0.70                    & 0.5431              & 0.38                     & 0.71                     \\
        \textsc{ResNet-18 } & \textsc{Early Fusion} & \textsc{Ego-View}            & \checkmark &            & \checkmark & 0.8599     & 0.6463              & 0.52                     & \textbf{0.77}                    & 0.6824              & 0.60                     & 0.77                     \\
        \textsc{ResNet-18 } & \textsc{Early Fusion} & \textsc{Ego-View}            &            & \checkmark & \checkmark & 0.8632     & 0.6082              & 0.36                     & 0.85                    & 0.6363              & 0.42                     & \textbf{0.85}                     \\
        \textsc{ResNet-18 } & \textsc{Early Fusion} & \textsc{Ego-View}            & \checkmark & \checkmark & \checkmark & 0.8417     & \textbf{0.6724}              & 0.57                     & \textbf{0.77}                    & \textbf{0.7234}              & \textbf{0.67}                     & 0.78                     \\
\arrayrulecolor{LightGrayForTableRule}
\hline
\arrayrulecolor{Black}
        \textsc{ResNet-50 } & \textsc{Single Modality}* & \textsc{BEV}                 & \checkmark &            &            & 0.5000     & 0.5000              & 0.50                     & 0.50                 & -                   & -                        & -                        \\
        \textsc{ResNet-50 } & \textsc{Single Modality} & \textsc{BEV}                 &            &            & \checkmark &  0.8900    & 0.5754	           &  0.50	                  & 0.65                 & -                   & -                        & -                        \\
        \textsc{ResNet-50 } & \textsc{Early Fusion} & \textsc{BEV}                 & \checkmark &            & \checkmark & 0.9007     & 0.6543              & 0.57                     & 0.74                    & -                   & -                        & -                        \\
        \textsc{ResNet-50 } & \textsc{Early Fusion} & \textsc{BEV}                 &            & \checkmark & \checkmark & 0.9153     & 0.6615	            & 0.60                     & 0.72                    & -                   & -                        & -                        \\
        \textsc{ResNet-50 } & \textsc{Early Fusion} & \textsc{BEV}                 & \checkmark & \checkmark & \checkmark & \textbf{0.9130}     & \textbf{0.6728}              & \textbf{0.58}                    & \textbf{0.77}       & -                        & -                        & -                       \\       
        \bottomrule  
    \end{tabular}
    \endgroup
    \end{adjustbox}    
    \label{tab:results-modalities}    
    %\vspace{-2em}              
    \end{table*}

    \begin{table*}[]
        %\vspace{-2em}
        \caption{Controlled evaluation of the benefit and influence of dropout of data modalities, on \emph{TbV-Beta}.  Rows marked with an asterisk represent an expected mean accuracy based on randomly flipped labels, rather than results from a trained model.}
        %\vspace{-1em}
            \centering
            \begin{adjustbox}{max width=\columnwidth} % ICCV 2\columnwidth
            \begingroup
            \begin{tabular}{llc ccc c ccc ccc}
                \toprule
                &    &     & \multicolumn{3}{|c|}{\textsc{Modalities}}    & \textsc{Visibility-based} & \multicolumn{3}{|c|}{\textsc{BEV proximity}}     & \multicolumn{3}{|c}{\textsc{Visibility-based}}    \\
                &    &     & \multicolumn{3}{|c|}{}   & \textsc{Eval @ 20m}    & \multicolumn{3}{|c|}{\textsc{Eval. @20m}}     & \multicolumn{3}{|c}{\textsc{Eval @20m}}    \\
         \textsc{Backbone} & \textsc{Arch.} & \textsc{Viewpoint} & \multicolumn{1}{|c}{\textsc{RGB}} & \textsc{Semantics} & \multicolumn{1}{c|}{\textsc{Map}} & \textsc{val}        & \multicolumn{1}{|c}{\textsc{test}} &      \textsc{Is Changed}     & \textsc{No Change}     & \multicolumn{1}{|c}{\textsc{test}}       & \textsc{Is Changed}     & \textsc{No Change} \\
                           &               &                    &  \multicolumn{1}{|c}{ }            &                   &  \multicolumn{1}{c|}{ }            & \textsc{mAcc}       & \multicolumn{1}{|c}{\textsc{mAcc}} &      \textsc{Acc}            & \textsc{Acc}           & \multicolumn{1}{|c}{\textsc{mAcc}}       & \textsc{Acc}            & \textsc{Acc}     \\ 
            \midrule
            -         & \textsc{Random Chance}* & -        & -          & -          & -          & 0.5000    & 0.5000    & 0.50  & 0.50  & 0.5000    & 0.50  & 0.50  \\
            \textsc{ResNet-18} & \textsc{Early Fusion}  & \textsc{Ego-View} & \checkmark & \checkmark & \checkmark & 0.8417 & 0.6724 & 0.57 & 0.77 & 0.7234 & 0.67 & 0.78 \\
            \textsc{ResNet-18} & \textsc{Early Fusion}  & \textsc{Ego-View} & dropout    & dropout    & \checkmark & \textbf{0.8605} & 0.6581 & 0.51 & 0.81 & 0.6926 & 0.58 & \textbf{0.81} \\
            \textsc{ResNet-18} & \textsc{Early Fusion}  & \textsc{Ego-View} & \checkmark & dropout    & dropout    & 0.8384 & \textbf{0.6850}  & \textbf{0.63} & 0.74 & 0.\textbf{7342} & \textbf{0.72} & 0.74 \\
            \textsc{ResNet-18} & \textsc{Early Fusion}  & \textsc{Ego-View} & \checkmark & dropout    & \checkmark & 0.8474 & 0.6483 & 0.51 & 0.78 & 0.6914 & 0.6  & 0.79 \\
            \textsc{ResNet-18} & \textsc{Early Fusion}  & \textsc{Ego-View} & \checkmark & \checkmark & dropout    & 0.8429 & 0.6617 & 0.51 & \textbf{0.82} & 0.6994 & 0.58 & \textbf{0.81} \\
    \bottomrule  
    \end{tabular}
    \endgroup
    \end{adjustbox}  
    \label{tab:effectdropout}            
    \vspace{-5mm}
    \end{table*}

%, along with per-class precision and recall. Other works used...

\paragraph{Which data viewpoint is most effective?} Models that operate on an ego-view scene perspective are more effective than those operating in the bird's eye view (5\% more effective over their own respective field of view), achieving 72.3\% mAcc (see Table \ref{tab:renderingviewpoint}). We found a simpler architecture (ResNet-18) to outperform ResNet-50 in the ego-view.

\paragraph{Which modality fusion method is most effective?} Early fusion. For both BEV and ego-view, the early fusion models significantly outperform the late fusion models (+22.8\% mAcc in the ego-view and +9.7\% mAcc in BEV). This may be surprising, but we attribute this to the benefit of early alignment of map and sensor image channels for the early fusion models. Instead, the late-fusion model performs alignment with greatly reduced spatial resolution in a higher-dimensional space, and is forced to make decisions about both data streams \emph{independently}, which may be suboptimal. While the map and sensor images represent different modalities, a shared feature extractor is useful for both.

\paragraph{Which input modalities are most effective?} A combination of RGB sensor data, semantics, and the map. We compare validation and test set performance of various input modalities in Table 
\ref{tab:results-modalities}. Early fusion of map and sensor data is compared with models that have access to only sensor or map data, or a combination of the two, with or without semantics. All models suffer a significant performance drop on the test set compared to the validation set. While a gap between validation and test performance is undesirable, better synthesis heuristics and better machine learning models can close that gap. %This validates our hypothesis that a new dataset of real-world map changes is required, rather than relying upon evaluation upon synthetic data \cite{Heo20iros_HDMapChangeCrossDomain}. 
% "This validates our hypothesis that a new dataset
% of real-world map changes is required, rather than relying upon evaluation upon synthetic data [15]."
% I think we can change this a bit. We don't need to argue about the relative value of our dataset any more. But I do want to emphasize that while a gap between validation and test performance is bad, better synthesis heuristics and better ML models can close that gap.
% I think a reader might see the large gap and say "Oh, your dataset is flawed", but I guess I want people to think "Hmm, maybe there's a better way to fake map changes for training".
We find semantic segmentation label map input to be quite helpful, although it places a dependency upon a separate model at inference time, increasing latency for a real-time system. Mean accuracy improves by 4\% in the ego-view and 2\% in the BEV when sensor and map data is augmented with semantic information in early fusion. In fact, early fusion of the map with the semantic stream alone (without sensor data) is 1\% more effective than using corresponding sensor data for the BEV.

\paragraph{Is sensor data necessary?} Yes. The ``map-only'' model we trained couldn’t meaningfully identify map changes over random chance, achieving mean accuracies on the test set of just 0.5754 in the BEV and 0.533 in an ego-view (see Table \ref{tab:results-modalities}). This model can also be thought of as a binary classifier which is trained to identify whether the HD map is real or synthetically manipulated (i.e. classify the source domain). Inspection via Guided GradCAM demonstrates that the map-only baseline attends to onboard map areas that are not in compliance with real-world priors, such as identifying asymmetry in crosswalk layout, paint patterns, and lane subdivisions.

%The prevailing prior is to feature 4 crosswalks at a 4-way intersection, and most map changes for crosswalks seem to occur when that prior is not respected

%While sensor-only measurements might seem to be doomed to fail at map change detection, we raise the hypothesis that for certain classes of changes, one may not require a map to determine that a change has recently occurred. Telltale clues could be a worker painting the road, recent paint or recent repaving of road asphalt.

\paragraph{Ablation on Modality Dropout.}
We find random drop-out of certain combinations of modalities to regularize the training in a beneficial way, improving accuracy by more than 1\% of our best model (See Table \ref{tab:effectdropout}). Given the wide array of modalities available to solve the task, from RGB sensor data, semantic label maps, rendered maps, and LiDAR reflectance data, we experiment with methods to force the network to learn to extract useful information from multiple modalities. Specifically, we perform random dropout of modalities, an approach developed in the self-supervised learning literature \cite{Vincent08icml_DenoisingAutoencoders,Doersch15iccv_ContextPrediction,Zhang17cvpr_SplitBrainAutoencoder}.\\
%This approach has been employed before in the literature; Denoising Autoencoders \cite{Vincent08icml_DenoisingAutoencoders} randomly zeros out a fixed number  of components. Half of the input (half of the color channels) to the Split-Brain Autoencoder \cite{Zhang17cvpr_SplitBrainAutoencoder} are zero-ed out, while replacing color channels with Gaussian noise \cite{Doersch15iccv_ContextPrediction} has been shown to prevent finding trivial solutions to semi-supervised learning pretext tasks.

 Perhaps the most intuitive approach would be to apply modality dropout to one of the sensor or semantic streams, forcing the network to extract useful features from both modalities during training. However, we find this is in fact detrimental. More effective, we discover, is to randomly drop out either the map or semantic streams. In theory, meaningful learning should be impossible without access to the map; however, since we drop-out each example in a batch with 50\% probability, in expectation 50\% of the examples should yield useful gradients in each batch. This approach improves accuracy by more than 1\% of our best model. We zero out all data from a specific modality as our drop out technique.

\begin{figure*}[]
    \vspace{-1.5em}
    \centering
    \subfloat[New paint constricting the intersection and bollards are added. ]{
        \includegraphics[width=0.3\columnwidth]{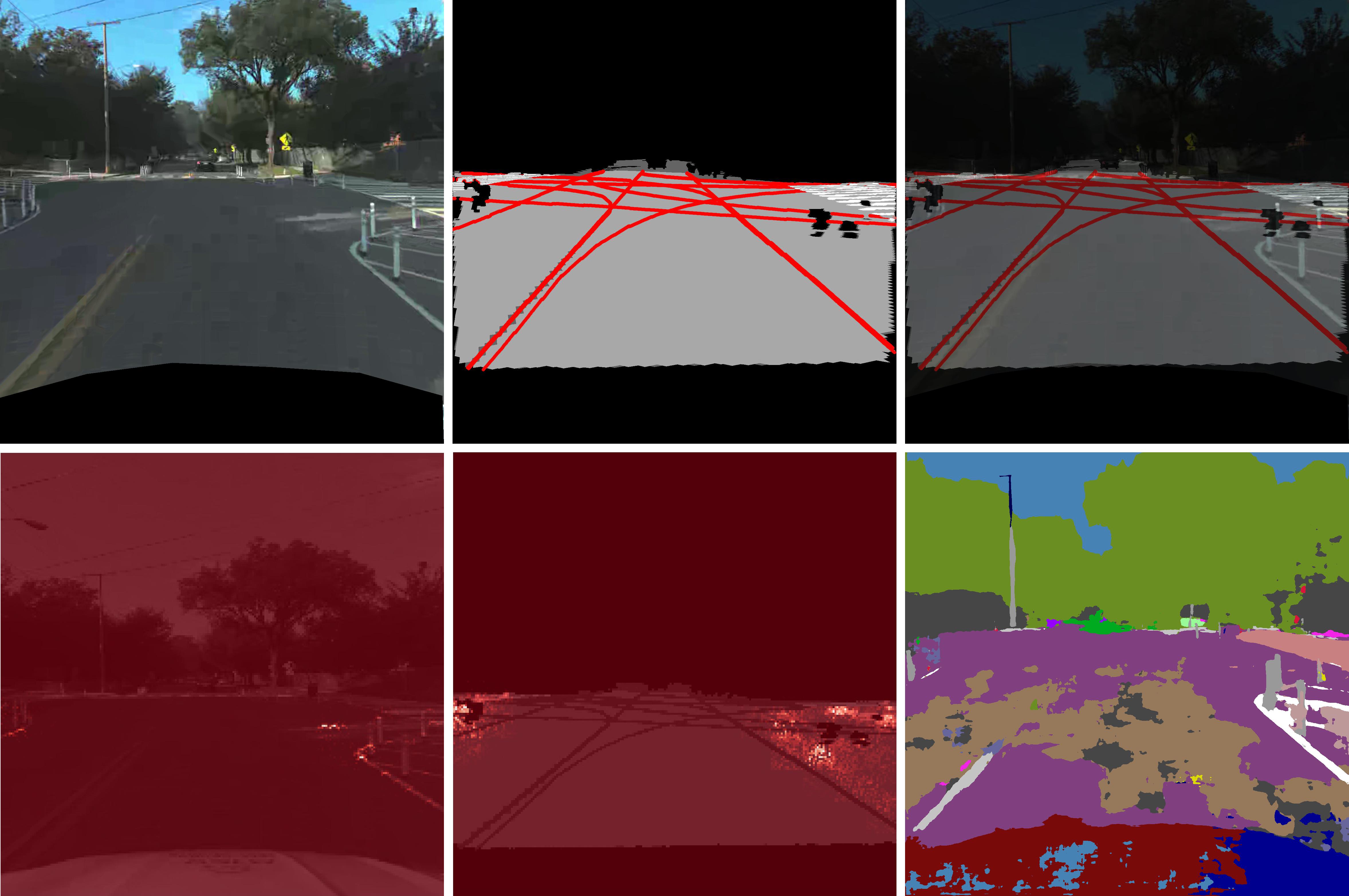}
    }
    %\vspace{-2mm}
    \hspace{2mm}
    \subfloat[A 3-lane road has been converted to a 2-lane road. ]{
    \includegraphics[width=0.3\columnwidth]{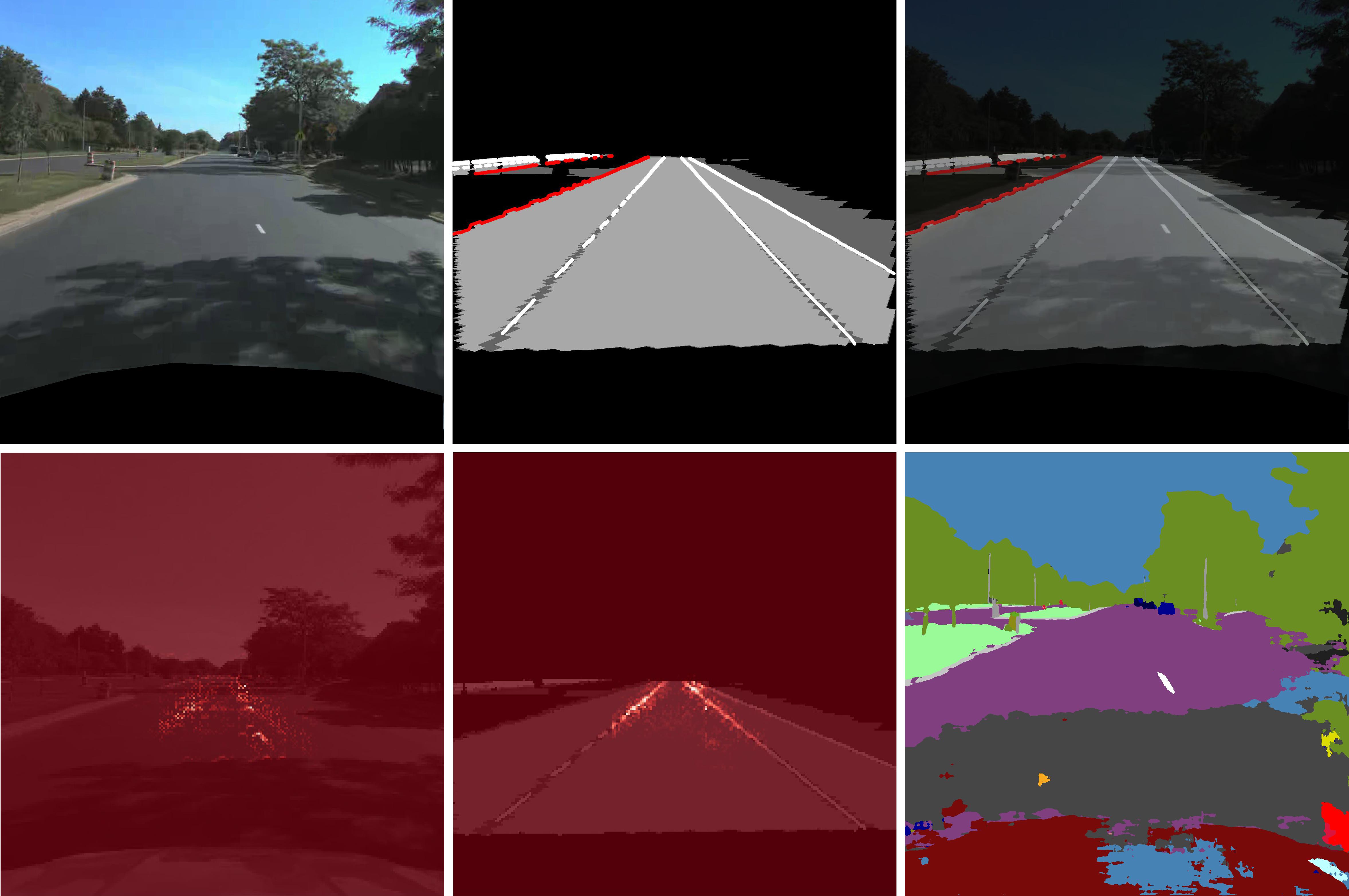}
    }
    %\vspace{-2mm}
    \hspace{2mm}
    \subfloat[Crosswalk paint has been removed. ]{
        \includegraphics[width=0.3\columnwidth]{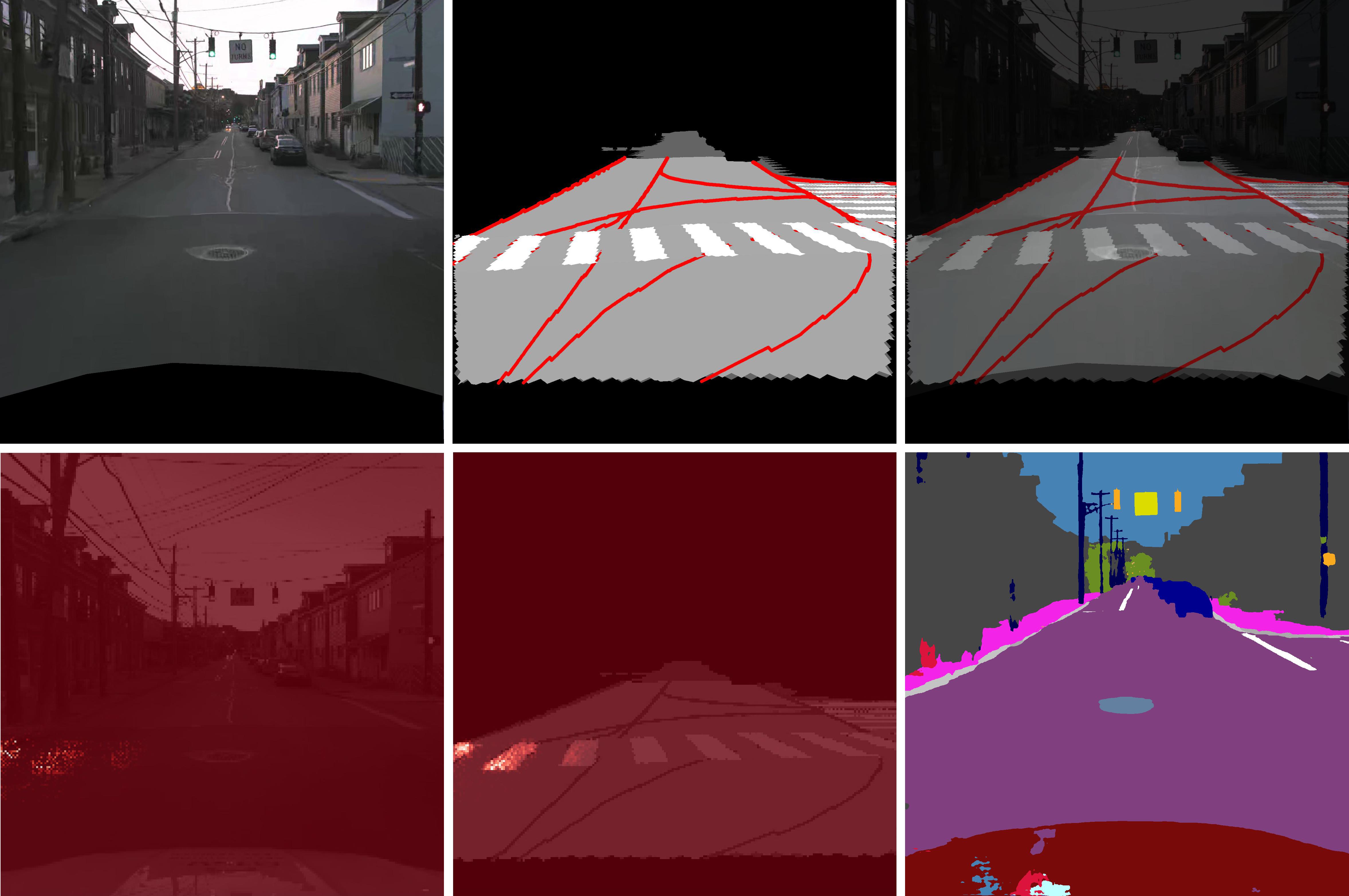}
    }
    %\hspace{0mm}
    % \includegraphics[width=\columnwidth]{figs/gradcam_examples/VrKB4vQ6rn05E6lC7C6KNSWXRbqImakD__2020-09-14-Z1F0061__315984051249927218.jpg}
    % \includegraphics[width=\columnwidth]{figs/gradcam_examples/5egFz1KKT87Q8h42m5cTm2vrzgSYordr__2020-06-08-Z1F0043__315966674549927220.jpg}
    % \includegraphics[width=\columnwidth]{figs/gradcam_examples/8zeNElNT6mNvKxUAzB0IooE0qo8I75e9__2020-06-25-Z1F0046__315970593499927223.jpg}
    % \includegraphics[width=\columnwidth]{figs/gradcam_examples/79TuaBnpYGOFKZAT1ZcmkUNEjHJtiiup__2020-10-07-Z1F0018__315969356399927212.jpg}
    % \includegraphics[width=\columnwidth]{figs/gradcam_examples/EP9PjbeBPfKRiTfbvKhzzyzjh8KoHDxE__2020-07-13-Z1F0045__315973148149927222.jpg}
    % \includegraphics[width=\columnwidth]{figs/gradcam_examples/IfDKJGa3z4LEFCYj5JSLGEuw9LI4UIlI__2020-08-14-Z1F0043__315980260399927223.jpg}
    % \includegraphics[width=\columnwidth]{figs/gradcam_examples/Jp6Jnwfg3lmF9h1lRugGBf6EV4cxFUlU__2020-07-17-Z1F0025__315971527349927211.jpg}
    % \includegraphics[width=\columnwidth]{figs/gradcam_examples/ouQqWBO1kAyPrbkIx7LcBVQgInxdntTq__2020-06-23-Z1F0066__315973307149927221.jpg}
    % \includegraphics[width=\columnwidth]{figs/gradcam_examples/Sek2BIMukaeBarKaxPqPXS4FiKTkrxNB__2020-07-17-Z1F0045__315972754099927217.jpg}
    % \includegraphics[width=\columnwidth]{figs/gradcam_examples/SP1puuZR5pn6JY3eT7qIU9XQbw3qh2iK__2020-07-29-Z1F0017__315972184199927219.jpg}
    % \includegraphics[width=\columnwidth]{figs/gradcam_examples/TOQaS7eCZB2ns552jjhmQoPs2MbdkgC6__2020-08-03-Z1F0037__315971187799927217.jpg}
    \caption{Guided GradCAM. 6 figures are shown for frames from various test set logs. Clockwise, from top-left: ego-view sensor image, rendered map in ego-view, blended combination of sensor and map, \emph{seamseg} label map, GradCAM activations for the map input, GradCAM activations for the sensor input. White color shows maximal activation, and red color shows zero activation in the heatmap palette. Label maps from \emph{seamseg} are at times quite noisy. }
    \label{fig:gradcamexamples}
    \vspace{-1.5em}
\end{figure*}

\paragraph{Interpretability and Localizing Changes.} While accurately perceiving changes is important, the ability to localize them would also be helpful. Bounding boxes are often unsuitable for a compact localization of a map change because most changed map regions in TbV relate to ``stuff'' classes, for example, linear, extended lane boundary markings. Instead, pixelwise localization is often more appropriate. We use Guided Grad-CAM \cite{Selvaraju17iccv_GradCAM} to identify which regions of the sensor, map, and semantic input are most relevant to the prediction of the `is-changed' class. In Figure \ref{fig:gradcamexamples}, we show qualitative results on frames for which our best model predicts real-world map changes have occurred.
%Change detection is the first step towards identifying specific map entities which are out of date, and we introduce a method to do so by identifying portions of the input that maximally affect the ``is-changed'' logit in our trained classification networks, via Grad-CAM \cite{Selvaraju17iccv_GradCAM}. Our dataset's centimeter-level annotations of changed entities supports the changed-entity identification task.
%Guided GradCAM effectively 
%  While Grad-CAM is also useful, we rely upon Guided Grad-CAM for its ability to reason at a higher resolution.
% uses the gradients of any target concept, flowing into the final convolutional layer to produce a coarse localization map highlighting important regions in the image for predicting the concept.
% \subsection{Qualitative Results}
% \textcolor{red}{TODO: add common success cases versus common failure modes}
%
%TbV use acronym in paper 
%Say pretrained helps

%\subsection{Map-Only 3d Point Filtering}
%However, entities can easily be occluded at any given time
%Network should learn to ignore dynamic objects and map elements occluded by dynamic objects

% our method with several ablations that include:  5) ray-traced ground-surface pixel correspondence; 
% 5) LiDAR-based pixel correspondence;
% 7) reflectance-based imagery.

    \begin{table}[b]
        %\vspace{-2em}
        \caption{Analysis of computational runtime for a single feedforward pass through the network (in milliseconds), for our various map change detection models.}
        %\vspace{-0.9em}
            \centering
            \begin{adjustbox}{max width=\columnwidth} % ICCV 2\columnwidth
            \begingroup
            \begin{tabular}{l ccc }
                \toprule
                \textsc{\textbf{Model (Input Modalities)}} & \textsc{\textbf{\# Input Channels}}   & \multicolumn{2}{c}{\textsc{\textbf{Runtime (ms)}}}  \\
          &  & \multicolumn{1}{c}{\textsc{ResNet-18 Backbone}}  & \multicolumn{1}{c}{\textsc{ResNet-50 Backbone}} \\
            \midrule
             \textsc{No Fusion (Map-only)}               &     3   &      2.09     &    6.48    \\
             \textsc{No Fusion (Semantic labelmap only)} &     5   &     2.10     &     6.64   \\
            \textsc{Early Fusion (Sensor + Map)}  & 3 + 3 = 6 &  2.19  &   6.51 \\
            \textsc{Early Fusion (Sensor + Map + Semantics)} & 3 + 3 + 5 = 11 & 2.24  &   6.61  \\
            \textsc{Late Fusion (Sensor + Map)} & 3, 3 &    4.29   & 12.99  \\
    \bottomrule  
    \end{tabular}
    \endgroup
    \end{adjustbox}  
    \label{tab:computationalruntime}            
    %\vspace{-5mm}
    \end{table}

\paragraph{Computational Runtime.} In order to demonstrate that our models can be used in practice without introducing a heavy computational burden, we report the time to complete a feedforward pass through a network, for our various models (see Table \ref{tab:computationalruntime}). The network architectures we employ are computationally lightweight, using ResNet-18 or ResNet-50 backbones. We report feedforward runtime (in milliseconds) for each network, averaged over 100 forward passes, after a warm-up period of 20 forward passes. The hardware used for the analysis is a Quadro P5000 GPU and Intel(R) Xeon(R) W-2145 CPU @ 3.70GHz processor, running the Ubuntu 20.04 operating system. Our best performing map change model (operating in the ego-view) requires just 2.24 milliseconds to complete a forward pass. However, this model does assume access to a readily-available semantic labelmap produced by a semantic segmentation network, which would increase the system latency. In Table \ref{tab:results-modalities}, we show that for a 4\% drop in accuracy for the ego-view models, the semantic labelmap can be excluded from the inputs, in which case the total runtime is just 2.19 milliseconds, well within real-time performance, introducing minimal latency for perception or planning modules which might utilize this information.

\section{Conclusion}

\noindent\textbf{Discussion.} In this work, we have introduced the first dataset for a new and challenging problem, map change detection. Our dataset is one of the largest AV datasets at the present time, featuring 1043 logs with an average duration of 54 seconds. We implement various approaches as strong baselines to explore this task for the first time with real data. Perhaps surprisingly, we find that comparing maps in a metric representation (a bird's eye view) is inferior to operating directly in the ego-view. We attribute this to a loss of texture during the projection process, and to a more difficult task of reasoning about a much larger spatial area ($85^{\circ}$ f.o.v. instead of $360^{\circ}$ f.o.v.). In addition, we provide a new method for localizing changed map entities, thereby facilitating efficient updates to HD maps.\\

We identify a significant gap between validation accuracy and test accuracy -- 10-20\% less on the test split -- which supports the importance of testing on real data. If performance is only measured on fake changes that resemble one's training distribution, performance can appear much better than what occurs in reality. Real changes can be subtle, and we hope the community will use this dataset to further push the state-of-the-art we introduce. We make publicly available our data, models, and code to generate our dataset and reproduce our results.

\label{sec:limitations}
\paragraph{Limitations.} \textbf{Rendering time.} A second limitation of our work is that real-time rendering requires GPU hardware; in the ego-view, map entity tesselation and rasterization are costly, whereas in the BEV, ray-casting is computationally intensive. \textbf{Perturbation diversity.} In our work, we introduce just 6 types of possible map perturbations, of which far more types are possible; nonetheless, we prove that they are surprisingly useful. \textbf{Accuracy.} Perhaps last of all, although our baselines have reasonable performance and by inspection we demonstrate they are learning to attend to meaningful regions, a large gap still exists before such a model would be accurate enough to be used on-vehicle.

%\textbf{Sensor quality}  Accordingly, we compress the images to a low bitrate so that the dataset doesn't become computationally unwieldy. None-the-less, the images are still interpretable, as shown in the figures. 

%%%% For Acknowledgements
% \begin{ack}

% \end{ack}

\medskip
\small
{
\bibliographystyle{plain}
\bibliography{egbib.bib}

\begin{thebibliography}{10}

\bibitem{Adelson91book_PlenopticFn}
Edward~H. Adelson and James~R. Bergen.
\newblock The plenoptic function and the elements of early vision.
\newblock In {\em Computational Models of Visual Processing}, pages 3--20. MIT
  Press, Cambridge, MA, 1991.

\bibitem{Alcantarilla18ar_StreetviewChangeDet}
Pablo~F Alcantarilla, Simon Stent, German Ros, Roberto Arroyo, and Riccardo
  Gherardi.
\newblock Street-view change detection with deconvolutional networks.
\newblock {\em Autonomous Robots}, 42(7):1301--1322, 2018.

\bibitem{Bacha08jfr_VictorTangoDARPA}
Andrew Bacha, Cheryl Bauman, Ruel Faruque, Michael Fleming, Chris Terwelp,
  Charles Reinholtz, Dennis Hong, Al~Wicks, Thomas Alberi, David Anderson,
  Stephen Cacciola, Patrick Currier, Aaron Dalton, Jesse Farmer, Jesse Hurdus,
  Shawn Kimmel, Peter King, Andrew Taylor, David~Van Covern, and Mike Webster.
\newblock Odin: Team victortango's entry in the darpa urban challenge.
\newblock {\em J. Field Robot.}, 25(8):467--492, August 2008.

\bibitem{Caesar20cvpr_nuscenes}
Holger Caesar, Varun Bankiti, Alex~H Lang, Sourabh Vora, Venice~Erin Liong,
  Qiang Xu, Anush Krishnan, Yu~Pan, Giancarlo Baldan, and Oscar Beijbom.
\newblock nuscenes: A multimodal dataset for autonomous driving.
\newblock In {\em CVPR}, pages 11621--11631, 2020.

\bibitem{Caesar18cvpr_COCOStuff}
Holger Caesar, Jasper Uijlings, and Vittorio Ferrari.
\newblock Coco-stuff: Thing and stuff classes in context.
\newblock In {\em CVPR}, June 2018.

\bibitem{Casas18corl_IntentNet}
Sergio Casas, Wenjie Luo, and Raquel Urtasun.
\newblock Intentnet: Learning to predict intention from raw sensor data.
\newblock In {\em 2nd Annual Conference on Robot Learning, CoRL 2018,
  Z{\"{u}}rich, Switzerland, 29-31 October 2018, Proceedings}, volume~87 of
  {\em Proceedings of Machine Learning Research}, pages 947--956. {PMLR}, 2018.

\bibitem{Chang19cvpr_Argoverse}
Ming-Fang Chang, John Lambert, Patsorn Sangkloy, Jagjeet Singh, Slawomir Bak,
  Andrew Hartnett, De~Wang, Peter Carr, Simon Lucey, Deva Ramanan, and James
  Hays.
\newblock Argoverse: 3d tracking and forecasting with rich maps.
\newblock In {\em CVPR}, June 2019.

\bibitem{Chen19corl_LearningByCheating}
Dian Chen, Brady Zhou, Vladlen Koltun, and Philipp Kr\"ahenb\"uhl.
\newblock Learning by cheating.
\newblock In {\em Conference on Robot Learning (CoRL)}, 2019.

\bibitem{Chen21cvpr_GeoSim}
Yun Chen, Frieda Rong, Shivam Duggal, Shenlong Wang, Xinchen Yan, Sivabalan
  Manivasagam, Shangjie Xue, Ersin Yumer, and Raquel Urtasun.
\newblock Geosim: Realistic video simulation via geometry-aware composition for
  self-driving.
\newblock In {\em CVPR}, June 2021.

\bibitem{Wang02icra_DetectTrackMovingObjects}
{Chieh-Chih Wang} and C.~{Thorpe}.
\newblock Simultaneous localization and mapping with detection and tracking of
  moving objects.
\newblock In {\em ICRA}, 2002.

\bibitem{Davison18arxiv_FutureMapping}
Andrew~J. Davison.
\newblock Futuremapping: The computational structure of spatial {AI} systems.
\newblock {\em CoRR}, abs/1803.11288, 2018.

\bibitem{Ding20icra_ChangingCityScenes}
W.~{Ding}, S.~{Hou}, H.~{Gao}, G.~{Wan}, and S.~{Song}.
\newblock Lidar inertial odometry aided robust lidar localization system in
  changing city scenes.
\newblock In {\em ICRA}, 2020.

\bibitem{Doersch15iccv_ContextPrediction}
Carl Doersch, Abhinav Gupta, and Alexei~A. Efros.
\newblock Unsupervised visual representation learning by context prediction.
\newblock In {\em ICCV}, December 2015.

\bibitem{Waymo21_MotionDataset}
Scott Ettinger, Shuyang Cheng, Benjamin Caine, Chenxi Liu, Hang Zhao, Sabeek
  Pradhan, Yuning Chai, Ben Sapp, Charles~R. Qi, Yin Zhou, Zoey Yang,
  Aur\'elien Chouard, Pei Sun, Jiquan Ngiam, Vijay Vasudevan, Alexander
  McCauley, Jonathon Shlens, and Dragomir Anguelov.
\newblock Large scale interactive motion forecasting for autonomous driving:
  The waymo open motion dataset.
\newblock In {\em ICCV}, October 2021.

\bibitem{Gao20cvpr_VectorNet}
Jiyang Gao, Chen Sun, Hang Zhao, Yi~Shen, Dragomir Anguelov, Congcong Li, and
  Cordelia Schmid.
\newblock Vectornet: Encoding hd maps and agent dynamics from vectorized
  representation.
\newblock In {\em CVPR}, June 2020.

\bibitem{Garnett19iccv_3dLaneNet}
Noa Garnett, Rafi Cohen, Tomer Pe'er, Roee Lahav, and Dan Levi.
\newblock 3d-lanenet: End-to-end 3d multiple lane detection.
\newblock In {\em ICCV}, October 2019.

\bibitem{Gebru21_datasheets}
Timnit Gebru, Jamie Morgenstern, Briana Vecchione, Jennifer~Wortman Vaughan,
  Hanna Wallach, Hal~Daum\'{e} III, and Kate Crawford.
\newblock Datasheets for datasets.
\newblock {\em Commun. ACM}, 64(12):86–92, nov 2021.

\bibitem{Geiger09ivs_MonocularRoadMosaicing}
Andreas Geiger.
\newblock Monocular road mosaicing for urban environments.
\newblock In {\em 2009 IEEE Intelligent Vehicles Symposium}, pages 140--145.
  IEEE, 2009.

\bibitem{Hahnel03icra_MapBuildingDynamic}
Dirk Hahnel, Rudolph Triebel, Wolfram Burgard, and Sebastian Thrun.
\newblock Map building with mobile robots in dynamic environments.
\newblock In {\em ICRA}, volume~2, pages 1557--1563. IEEE, 2003.

\bibitem{Han15cvpr_MatchNet}
Xufeng Han, Thomas Leung, Yangqing Jia, Rahul Sukthankar, and Alexander~C.
  Berg.
\newblock Matchnet: Unifying feature and metric learning for patch-based
  matching.
\newblock In {\em CVPR}, June 2015.

\bibitem{He16cvpr_ResNet}
Kaiming He, Xiangyu Zhang, Shaoqing Ren, and Jian Sun.
\newblock Deep residual learning for image recognition.
\newblock In {\em CVPR}, 2016.

\bibitem{Heo20iros_HDMapChangeCrossDomain}
Minhyeok Heo, Jiwon Kim, and Sujung Kim.
\newblock Hd map change detection with cross-domain deep metric learning.
\newblock In {\em 2020 IEEE/RSJ International Conference on Intelligent Robots
  and Systems (IROS)}, pages 10218--10224. IEEE, 2020.

\bibitem{Homayounfar18cvpr_HRAN}
Namdar Homayounfar, Wei-Chiu Ma, Shrinidhi~Kowshika Lakshmikanth, and Raquel
  Urtasun.
\newblock Hierarchical recurrent attention networks for structured online maps.
\newblock In {\em CVPR}, June 2018.

\bibitem{Homayounfar19iccv_DAGMapper}
Namdar Homayounfar, Wei-Chiu Ma, Justin Liang, Xinyu Wu, Jack Fan, and Raquel
  Urtasun.
\newblock Dagmapper: Learning to map by discovering lane topology.
\newblock In {\em ICCV}, October 2019.

\bibitem{Houston20_LyftLevel5}
John Houston, Guido Zuidhof, Luca Bergamini, Yawei Ye, Ashesh Jain, Sammy
  Omari, Vladimir Iglovikov, and Peter Ondruska.
\newblock One thousand and one hours: Self-driving motion prediction dataset.
\newblock {\em arXiv preprint arXiv:2006.14480}, 2020.

\bibitem{Sae18tr_SelfDrivingTaxonomy}
SAE International.
\newblock Taxonomy and definitions for terms related to driving automation
  systems for on-road motor vehicles.
\newblock Technical Report J3016, June 2018.

\bibitem{Jo18sensors_SLAM_MapChangeUpdate}
Kichun Jo, Chansoo Kim, and Myoungho Sunwoo.
\newblock Simultaneous localization and map change update for the high
  definition map-based autonomous driving car.
\newblock {\em Sensors}, 18(9):3145, 2018.

\bibitem{Karpathy20talk_CVPR}
Andrej Karpathy.
\newblock Cvpr 2020 workshop on scalability in autonomous driving: Keynote
  talk, 2020.
\newblock \url{https://www.youtube.com/watch?v=g2R2T631x7k}.

\bibitem{Kingma14arxiv_Adam}
Diederik~P Kingma and Jimmy Ba.
\newblock Adam: A method for stochastic optimization.
\newblock {\em arXiv preprint arXiv:1412.6980}, 2014.

\bibitem{Lambert18cvpr_DLUPI}
John Lambert, Ozan Sener, and Silvio Savarese.
\newblock Deep learning under privileged information using heteroscedastic
  dropout.
\newblock In {\em CVPR}, June 2018.

\bibitem{Liang20eccv_LaneGCN}
Ming Liang, Bin Yang, Rui Hu, Yun Chen, Renjie Liao, Song Feng, and Raquel
  Urtasun.
\newblock Learning lane graph representations for motion forecasting.
\newblock In {\em ECCV}, 2020.

\bibitem{Mani20wacv_MonoLayout}
Kaustubh Mani, Swapnil Daga, Shubhika Garg, Sai~Shankar Narasimhan, Madhava
  Krishna, and Krishna~Murthy Jatavallabhula.
\newblock Monolayout: Amodal scene layout from a single image.
\newblock In {\em WACV}, March 2020.

\bibitem{Mattyus17iccv_DeepRoadMapper}
Gellert Mattyus, Wenjie Luo, and Raquel Urtasun.
\newblock Deeproadmapper: Extracting road topology from aerial images.
\newblock In {\em ICCV}, Oct 2017.

\bibitem{Mattyus16cvpr_HDMaps}
Gellert Mattyus, Shenlong Wang, Sanja Fidler, and Raquel Urtasun.
\newblock Hd maps: Fine-grained road segmentation by parsing ground and aerial
  images.
\newblock In {\em CVPR}, June 2016.

\bibitem{Moller97jgt_FastRayTriangle}
Tomas M{\"o}ller and Ben Trumbore.
\newblock Fast, minimum storage ray-triangle intersection.
\newblock {\em Journal of graphics tools}, 2(1):21--28, 1997.

\bibitem{Montemerlo08jfr_JuniorUrbanChallenge}
Michael Montemerlo, Jan Becker, Suhrid Bhat, Hendrik Dahlkamp, Dmitri Dolgov,
  Scott Ettinger, Dirk Haehnel, Tim Hilden, Gabe Hoffmann, Burkhard Huhnke,
  Doug Johnston, Stefan Klumpp, Dirk Langer, Anthony Levandowski, Jesse
  Levinson, Julien Marcil, David Orenstein, Johannes Paefgen, Isaac Penny, Anna
  Petrovskaya, Mike Pflueger, Ganymed Stanek, David Stavens, Antone Vogt, and
  Sebastian Thrun.
\newblock Junior: The stanford entry in the urban challenge.
\newblock {\em J. Field Robot.}, 25(9):569--597, September 2008.

\bibitem{Pan19ral_crossview}
B.~{Pan}, J.~{Sun}, H.~Y.~T. {Leung}, A.~{Andonian}, and B.~{Zhou}.
\newblock Cross-view semantic segmentation for sensing surroundings.
\newblock {\em IEEE Robotics and Automation Letters}, 5(3):4867--4873, 2020.

\bibitem{Pannen20icra_HDMapUpToDate}
D.~{Pannen}, M.~{Liebner}, W.~{ Hempel}, and W.~{Burgard}.
\newblock How to keep hd maps for automated driving up to date.
\newblock In {\em ICRA}, 2020.

\bibitem{Pannen19icra_ChangeDetectionParticleFilter}
D.~{Pannen}, M.~{Liebner}, and W.~{Burgard}.
\newblock Hd map change detection with a boosted particle filter.
\newblock In {\em ICRA}, 2019.

\bibitem{Phan-Minh20cvpr_CoverNet}
Tung Phan-Minh, Elena~Corina Grigore, Freddy~A. Boulton, Oscar Beijbom, and
  Eric~M. Wolff.
\newblock Covernet: Multimodal behavior prediction using trajectory sets.
\newblock In {\em CVPR}, June 2020.

\bibitem{Philion20eccv_LiftSplatShoot}
Jonah Philion and Sanja Fidler.
\newblock Lift, splat, shoot: Encoding images from arbitrary camera rigs by
  implicitly unprojecting to 3d.
\newblock In {\em ECCV}, 2020.

\bibitem{Philion20cvpr_PlannerCentric}
Jonah Philion, Amlan Kar, and Sanja Fidler.
\newblock Learning to evaluate perception models using planner-centric metrics.
\newblock In {\em CVPR}, June 2020.

\bibitem{Porzi19cvpr_Seamseg}
Lorenzo Porzi, Samuel Rota~Bul\`o, Aleksander Colovic, and Peter Kontschieder.
\newblock Seamless scene segmentation.
\newblock In {\em CVPR}, June 2019.

\bibitem{Rapo18thesis_orthoimagery}
Lauri Rapo.
\newblock Generating road orthoimagery using a smartphone.
\newblock Master's thesis, Lappeenranta University of Technology, 2018.

\bibitem{Roddick20cvpr_SemanticMap}
Thomas Roddick and Roberto Cipolla.
\newblock Predicting semantic map representations from images using pyramid
  occupancy networks.
\newblock In {\em CVPR}, June 2020.

\bibitem{Selvaraju17iccv_GradCAM}
Ramprasaath~R. Selvaraju, Michael Cogswell, Abhishek Das, Ramakrishna Vedantam,
  Devi Parikh, and Dhruv Batra.
\newblock Grad-cam: Visual explanations from deep networks via gradient-based
  localization.
\newblock In {\em ICCV}, Oct 2017.

\bibitem{Shaik17ki_DynamicMapUpdate}
Nayabrasul Shaik, Thomas Liebig, Christopher Kirsch, and Heinrich M{\"u}ller.
\newblock Dynamic map update of non-static facility logistics environment with
  a multi-robot system.
\newblock In {\em Joint German/Austrian Conference on Artificial Intelligence
  (K{\"u}nstliche Intelligenz)}, pages 249--261. Springer, 2017.

\bibitem{Shotton11cvpr_RealtimePose}
Jamie Shotton, Andrew Fitzgibbon, Mat Cook, Toby Sharp, Mark Finocchio, Richard
  Moore, Alex Kipman, and Andrew Blake.
\newblock Real-time human pose recognition in parts from single depth images.
\newblock In {\em CVPR}, 2011.

\bibitem{Silver16patent_ChangeDetCurveAlignment}
David~Harrison Silver and David Ian~Franklin Ferguson.
\newblock Change detection using curve alignment, April~26 2016.
\newblock US Patent 9,321,461.

\bibitem{Suo21cvpr_TrafficSim}
Simon Suo, Sebastian Regalado, Sergio Casas, and Raquel Urtasun.
\newblock {TrafficSim}: Learning to simulate realistic multi-agent behaviors.
\newblock In {\em CVPR}, June 2021.

\bibitem{Tan21cvpr_SceneGen}
Shuhan Tan, Kelvin Wong, Shenlong Wang, Sivabalan Manivasagam, Mengye Ren, and
  Raquel Urtasun.
\newblock Scenegen: Learning to generate realistic traffic scenes.
\newblock In {\em CVPR}, June 2021.

\bibitem{Tipaldi13ijrr_LifelongLoc}
Gian~Diego Tipaldi, Daniel Meyer-Delius, and Wolfram Burgard.
\newblock Lifelong localization in changing environments.
\newblock {\em The International Journal of Robotics Research},
  32(14):1662--1678, 2013.

\bibitem{Urmson07tr_TartanRacing}
Christopher Urmson, Joshua Anhalt, J.~Andrew~(Drew) Bagnell, Christopher~R.
  Baker, Robert~E. Bittner, John~M. Dolan, David Duggins, David Ferguson,
  Tugrul Galatali, Hartmut Geyer, Michele Gittleman, Sam Harbaugh, Martial
  Hebert, Thomas Howard, Alonzo Kelly, David Kohanbash, Maxim Likhachev, Nick
  Miller, Kevin Peterson, Raj Rajkumar, Paul Rybski, Bryan Salesky, Sebastian
  Scherer, Young-Woo Seo, Reid Simmons, Sanjiv Singh, Jarrod~M. Snider,
  Anthony~(Tony) Stentz, William (Red)~L. Whittaker, and Jason Ziglar.
\newblock Tartan racing: A multi-modal approach to the darpa urban challenge.
\newblock Technical report, Carnegie Mellon University, Pittsburgh, PA, April
  2007.

\bibitem{UsDot19pr_DrivingMoreThanEverBefore}
{U.S. Department of Transportation}.
\newblock Strong economy has americans driving more than ever before.
\newblock Press Release, March 2019.
\newblock \url{https://www.fhwa.dot.gov/pressroom/fhwa1905.cfm}.

\bibitem{Vincent08icml_DenoisingAutoencoders}
Pascal Vincent, Hugo Larochelle, Yoshua Bengio, and Pierre-Antoine Manzagol.
\newblock Extracting and composing robust features with denoising autoencoders.
\newblock In {\em ICML}, page 1096–1103, 2008.

\bibitem{Wang17iccv_TorontoCity}
Shenlong Wang, Min Bai, Gellert Mattyus, Hang Chu, Wenjie Luo, Bin Yang, Justin
  Liang, Joel Cheverie, Sanja Fidler, and Raquel Urtasun.
\newblock Torontocity: Seeing the world with a million eyes.
\newblock In {\em ICCV}, Oct 2017.

\bibitem{Wang14iccvw_CDnet}
Yi~Wang, Pierre-Marc Jodoin, Fatih Porikli, Janusz Konrad, Yannick Benezeth,
  and Prakash Ishwar.
\newblock Cdnet 2014: An expanded change detection benchmark dataset.
\newblock In {\em Proceedings of the IEEE conference on computer vision and
  pattern recognition workshops}, pages 387--394, 2014.

\bibitem{Wilson21neurips_Argoverse2}
Benjamin Wilson, William Qi, Tanmay Agarwal, John Lambert, Jagjeet Singh,
  Siddhesh Khandelwal, Bowen Pan, Ratnesh Kumar, Andrew Hartnett,
  Jhony~Kaesemodel Pontes, Deva Ramanan, Peter Carr, and James Hays.
\newblock Argoverse 2: Next generation datasets for self-driving perception and
  forecasting.
\newblock In {\em Proceedings of the Neural Information Processing Systems
  Track on Datasets and Benchmarks (NeurIPS Datasets and Benchmarks 2021)},
  2021.

\bibitem{Yang18corl_HDNet}
Bin Yang, Ming Liang, and Raquel Urtasun.
\newblock Hdnet: Exploiting hd maps for 3d object detection.
\newblock In {\em Proceedings of The 2nd Conference on Robot Learning},
  volume~87 of {\em Proceedings of Machine Learning Research}, pages 146--155.
  PMLR, 29--31 Oct 2018.

\bibitem{Yang20cvpr_SurfelGAN}
Zhenpei Yang, Yuning Chai, Dragomir Anguelov, Yin Zhou, Pei Sun, Dumitru Erhan,
  Sean Rafferty, and Henrik Kretzschmar.
\newblock Surfelgan: Synthesizing realistic sensor data for autonomous driving.
\newblock In {\em CVPR}, June 2020.

\bibitem{Zbontar15cvpr_matchingcost}
Jure Zbontar and Yann LeCun.
\newblock Computing the stereo matching cost with a convolutional neural
  network.
\newblock In {\em CVPR}, June 2015.

\bibitem{Zendel18eccv_WildDash}
Oliver Zendel, Katrin Honauer, Markus Murschitz, Daniel Steininger, and Gustavo
  Fernandez~Dominguez.
\newblock Wilddash - creating hazard-aware benchmarks.
\newblock In {\em ECCV}, 2018.

\bibitem{Zeng19cvpr_NeuralMotionPlanner}
Wenyuan Zeng, Wenjie Luo, Simon Suo, Abbas Sadat, Bin Yang, Sergio Casas, and
  Raquel Urtasun.
\newblock End-to-end interpretable neural motion planner.
\newblock In {\em CVPR}, June 2019.

\bibitem{Zhang14icrb_LaneLevelOrthophoto}
H.~{Zhang}, M.~{Yang}, C.~{Wang}, X.~{Weng}, and L.~{Ye}.
\newblock Lane-level orthophoto map generation using multiple onboard cameras.
\newblock In {\em 2014 IEEE International Conference on Robotics and
  Biomimetics (ROBIO 2014)}, pages 855--860, 2014.

\bibitem{Zhang17cvpr_SplitBrainAutoencoder}
Richard Zhang, Phillip Isola, and Alexei~A. Efros.
\newblock Split-brain autoencoders: Unsupervised learning by cross-channel
  prediction.
\newblock In {\em CVPR}, July 2017.

\bibitem{Zhao20corl_TNT}
Hang Zhao, Jiyang Gao, Tian Lan, Chen Sun, Benjamin Sapp, Balakrishnan
  Varadarajan, Yue Shen, Yi~Shen, Yuning Chai, Cordelia Schmid, Congcong Li,
  and Dragomir Anguelov.
\newblock Tnt: Target-driven trajectory prediction.
\newblock In {\em 4th Annual Conference on Robot Learning, CoRL 2020}, 2020.

\bibitem{Zhao17cvpr_PSPNet}
Hengshuang Zhao, Jianping Shi, Xiaojuan Qi, Xiaogang Wang, and Jiaya Jia.
\newblock Pyramid scene parsing network.
\newblock In {\em CVPR}, July 2017.

\end{thebibliography}
}

%%%%%%%%%%%%%%%%%%%%%%%%%%%%%%%%%%%%%%%%%%%%%%%%%%%%%%%%%%%%
% \section*{Checklist}
% \input{neurips-21/neurips-checklist.tex}

%%%%%%%%%%%%%%%%%%%%%%%%%%%%%%%%%%%%%%%%%%%%%%%%%%%%%%%%%%%%

\appendix

\section*{Appendix}

%\begin{abstract}
In this appendix, we provide additional details about our dataset and experiments. In Section (A), we provide an ablation study on the influence of input crop size on model performance. In Section (B), we discuss additional implementation details about our training, data augmentation, and occlusion-based map rendering process. In Section (C), we discuss the paired positive-negative logs we include. In Section (D), we describe our evaluation metric. In Section (E), we provide additional experimental analysis of different models and rendering viewpoints. In Section (F), we provide additional details about how we generate orthoimagery. In Section (G), we offer additional examples from our test set.  In Section (H), we give examples of other types of temporary map changes which we do not annotate or evaluate within our dataset. In Section (I) we provide further analysis of the frequency of map changes. In Section (J), we give  additional details about our synthetic map perturbation protocol. In Section (K), we provide a datasheet for the dataset. %, and in Section (L), we address required aspects of the submission for creation of a new dataset in the NeurIPS 2021 Datasets \& Benchmarks Track.
%\end{abstract}

%In Section (2), we show visualizations along the temporal dimension of logs with map changes. 
 %about training parameters and histogram matching. 
% across city per unit area throughout time.

% 3D point locations are quantized to float16.
% 257 Ground height maps are quantized to .3 meter resolution from their full resolution. HD map polygon
% 258 vertex locations are quantized to .01 meter resolution.

\section*{Appendix A: Influence of Input Crop Size}
In this section, we perform an ablation on input crop size, as discussed in Section \ref{sec:impl-details} of the main text. In the main paper, we set our input crop size to $224 \times 224$ px for all experiments mentioned therein. In this section, we present an ablation to measure the influence of input crop size. Again, we find the ego-view model is the best-performing model, as measured on its own field of view. Perhaps surprisingly, we find that an RGB image at $234 \times 234$ px resolution ($\sim$ 164K pixel values/image) is sufficient to capture significant detail. In Table  \ref{tab:resolutionablation}, we present an ablation where we find that for BEV models, higher resolution (i.e. $468 \times 468$ px) does improve mAcc by $2\%$ mAcc, although requiring almost 4x the GPU memory during training and significantly longer training times. However, for ego-view models, a higher crop size is quite detrimental, reducing visibility-based mAcc by around 7\%. 

\begin{table*}[h]
    %\vspace{-2em}
    \caption{Controlled evaluation of the influence of input crop size (for ego-view and BEV), on \emph{TbV-Beta}.} %.  Rows marked with an asterisk represent an expected mean accuracy based on randomly flipped labels, rather than results from a trained model.}
    \centering
    %\vspace{-1em}
    \begin{adjustbox}{max width=\columnwidth}
    \begingroup
    \begin{tabular}{c llc ccc c ccc ccc}
        \toprule
     &                &               &                  & \multicolumn{3}{|c|}{\textsc{Modalities}}                                                      & \textsc{Visibility-based} & \multicolumn{3}{|c|}{\textsc{BEV proximity}}     & \multicolumn{3}{|c}{\textsc{Visibility-based}}    \\
 &                   &               &                   & \multicolumn{3}{|c|}{}                                                                        & \textsc{Eval. @ 20m}    & \multicolumn{3}{|c|}{\textsc{Eval. @20m}}     & \multicolumn{3}{|c}{\textsc{Eval. @20m}}    \\
 \textsc{Resolution} &\textsc{Backbone} & \textsc{Arch.} & \textsc{Viewpoint} & \multicolumn{1}{|c}{\textsc{RGB}}    & \textsc{Semantics} & \multicolumn{1}{c|}{\textsc{Map}} & \textsc{val}        & \multicolumn{1}{|c}{\textsc{test}} &      \textsc{Is Changed}     & \textsc{No Change}     & \multicolumn{1}{|c}{\textsc{test}}       & \textsc{Is Changed}     & \textsc{No Change} \\
    &               &               &                    &  \multicolumn{1}{|c}{ }               &                   &  \multicolumn{1}{c|}{ }            & \textsc{mAcc}       & \multicolumn{1}{|c}{\textsc{mAcc}} &      \textsc{Acc}            & \textsc{Acc}           & \multicolumn{1}{|c}{\textsc{mAcc}}       & \textsc{Acc}            & \textsc{Acc}     \\ 
    \midrule
    % -         & \textsc{Random Chance}* & -         & -          & -          & -          & 0.5000         & 0.5000          & 0.50                 & 0.50                & 0.5000                  & 0.50                         & 0.50                        \\
    % \hline
        224x224 & ResNet-18 & Early Fusion & Ego-View & \checkmark & dropout & dropout    & 0.8384 & \textbf{0.6850}  & \textbf{0.63} & 0.74 & \textbf{0.7342} & \textbf{0.72} & 0.74 \\
        448x448 & ResNet-18 & Early Fusion & Ego-View & \checkmark & dropout & dropout    & 0.8713 & 0.6331 & 0.38 & \textbf{0.88} & 0.6644 & 0.45 & \textbf{0.88} \\
        \hline
        224x224 & ResNet-50 & Early Fusion & BEV      & \checkmark & no      & \checkmark & 0.9007 & 0.6543 & 0.57 & 0.74 &        &      &      \\
        448x448 & ResNet-50 & Early Fusion & BEV      & \checkmark & no      & \checkmark & \textbf{0.9072} & 0.6749 & \textbf{0.63} & 0.72 &        &      &     \\
    % \textsc{ResNet-18} & \textsc{Early Fusion}  & \textsc{Ego-View}  & \checkmark & \checkmark & \checkmark & 0.8417         & \textbf{0.6724}          & 0.57                 & 0.77                & \textbf{0.7234}                  & \textbf{0.67}                         & 0.78                        \\
    % \textsc{ResNet-18} & \textsc{Late Fusion}   & \textsc{Ego-View}  & \checkmark &            & \checkmark & 0.8108         & 0.4930           & 0.13                 & \textbf{0.85}                & 0.4956                  & 0.13                         & \textbf{0.86}                        \\
    % \hline
    % \textsc{ResNet-50} & \textsc{Early Fusion}  & \textsc{BEV}       & \checkmark & \checkmark & \checkmark & \textbf{0.9130}  & \textbf{0.6728}  & \textbf{0.58}                 & 0.77                & -                       & -                            & -                           \\
    % \textsc{ResNet-50} & \textsc{Late Fusion}   & \textsc{BEV}       & \checkmark &            & \checkmark & 0.8697         & 0.5761          & 0.43                 & 0.72                & -                       & -                            & -  \\
    \bottomrule  
    \end{tabular}
    \endgroup
    \end{adjustbox}
    \label{tab:resolutionablation}                   
    \end{table*}

\section*{Appendix B: Additional Implementation Details}
\subsection*{B.1. Training}
We train our models for 90 epochs with the Adam \cite{Kingma14arxiv_Adam} optimizer. We use a polynomial learning rate decay strategy, starting at $1 \times 10^{-3}$. We use a batch size of 1024 examples. We start with pretrained ImageNet weights for ResNet-18 or ResNet-50 \cite{He16cvpr_ResNet}.

We train with multiple negative examples per sensor image, which we found to be more beneficial than randomly sampling a single negative example (i.e. a synthetically perturbed map). In other words, we perform multiple types of perturbations for a given scene, and feed them to the network as separate negative examples (not necessarily in the same mini-batch).

\subsection*{B.2. Data Augmentation}
We employ a number of data augmentation techniques to improve the generalization of our models and prevent overfitting. Input images are of dimension $2048 \times 1550$ for the front-center camera, and $1550 \times 2048$ for all other 6 cameras. For the ego-view models, we first take a square crop from the bottom $1550 \times 1550$ of an ego-view image. Afterwards, we resize to $234 \times 234$, perform a random horizontal flip with 50\% probability, take a random $224 \times 224$ crop, divide pixel intensities by 255, and then normalize both sensor and map RGB channels by the ImageNet mean $(\mu_r,\mu_g,\mu_b) = (0.485, 0.456, 0.406)$ and standard deviation $(\sigma_r,\sigma_g,\sigma_b)=(0.229, 0.224, 0.225)$

For BEV models, we resize input images from $2000 \times 2000$ px to $234 \times 234$ px, perform a random horizontal and/or vertical flip with 50\% probability each (independently), choose a random $224 \times 224$ crop, and normalize as described above. 

We find other traditional data augmentation techniques from the semantic segmentation literature \cite{Zhao17cvpr_PSPNet}, such as applying a random rotation to the input or randomly blurring the input with a small kernel, to be ineffective.
    %First, we convolve the resized image with a $5 \times 5$ Gaussian kernel with $\sigma=1.1$ for blurring. Blurring of RGB images w/ 50\% probability has proven effective to be an effective data augmentation technique for ``doubling'' the number of training examples in the semantic segmentation literature, but we omit it in our latest models.

\subsection*{B.3. Occlusion Reasoning}

\begin{figure}[htb]
\vspace{-2mm}
\centering
\subfloat[RGB Image]{
    \includegraphics[width=0.25\columnwidth]{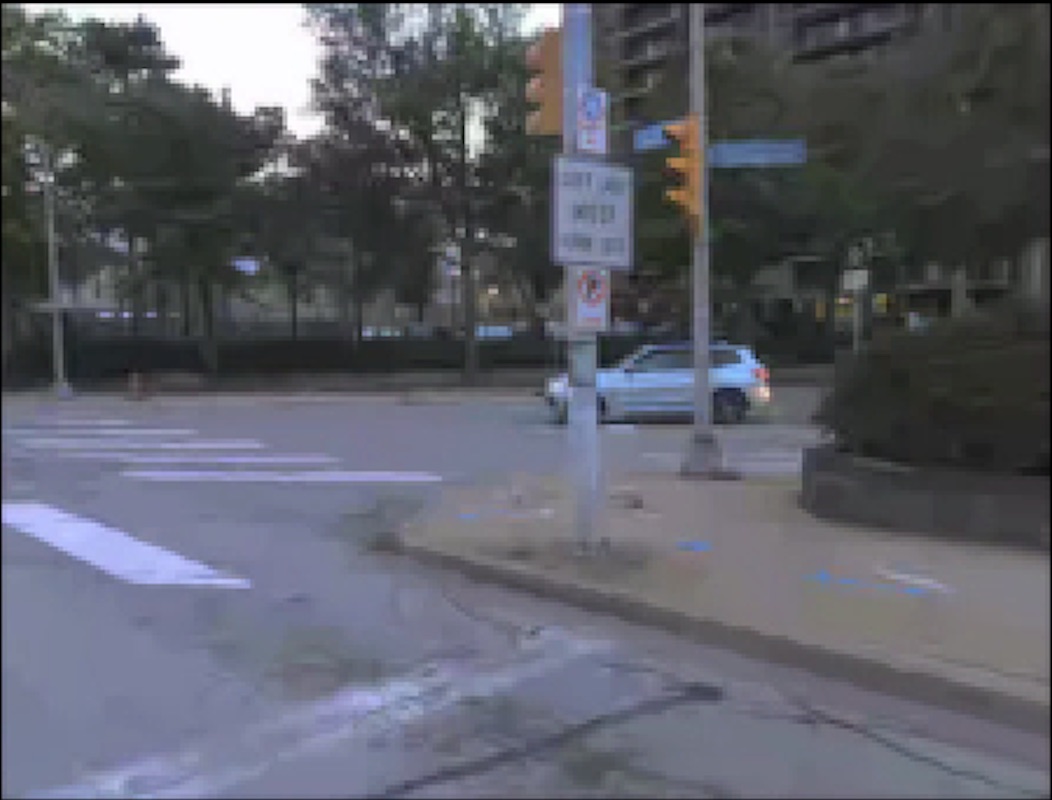}
}
\subfloat[Interpolated Depth Map]{
    \includegraphics[width=0.25\columnwidth]{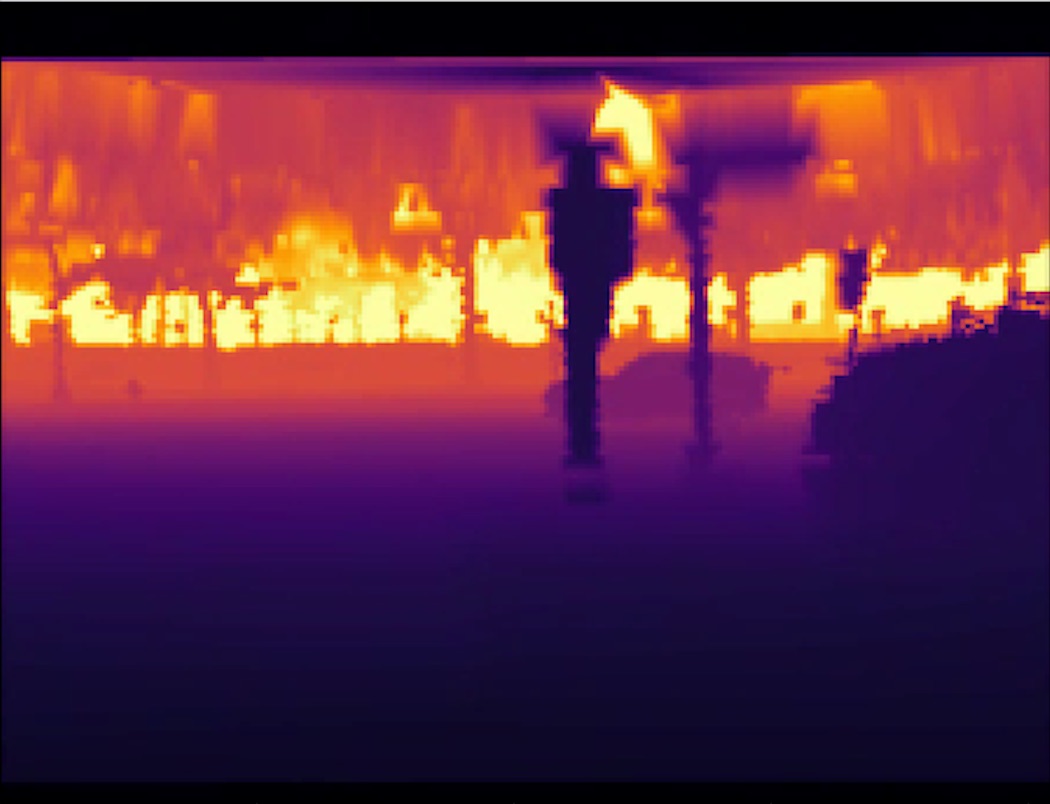}
}
%\vspace{-1em}
\caption{Example of a dense depth map interpolated from sparse LiDAR returns. }
\label{fig:depth-map}
\end{figure}

As discussed in Section \ref{sec:impl-details} of the main text, we use map occlusion reasoning when generating the input for our ego-view models. Occluded map elements and map elements that have been removed in the real world (``deleted'') are both not visible in camera imagery. While the former is an expected everyday occurrence, and the latter is of interest to us, we use occlusion reasoning in order to separate the two phenomena. We generate a dense depth map from sparse LiDAR returns (see Figure \ref{fig:depth-map}) and the depth of map entities is compared against the corresponding depth of its projection in the depth map. 

\subsection*{B.4. Details about Semantic Label Map Input}
As discussed in Section \ref{sec:impl-details} of the main text, we use semantic label maps generated from the semantic head of a publicly-available seamseg ResNet-50 panoptic segmentation model \cite{Porzi19cvpr_Seamseg} \footnote{Available at \url{https://github.com/mapillary/seamseg}.}. We create 5 binary mask channels from the semantic label map, for the `road', `bike-lane', `marking-crosswalk-zebra', `lane-marking-general', and `crosswalk-plain' classes. These are optionally provided as additional channels to the 3 RGB sensor channels and 3 RGB map channels via early fusion.  Seamseg's semantic label maps on their own do not capture sufficient granularity for the map change detection task we define, since the Mapillary Vistas public dataset's taxonomy does not differentiate between lane color and or different marking types (e.g. double-solid, solid, dashed-solid), which are of interest to autonomous vehicle operation.

\noindent\textbf{Unsuitability of Per-Pixel Semantic Comparison.} Directly comparing rendered map and semantic label maps at a \emph{per-pixel} level is not always useful since our HD map representation does not provide paint annotation for every single dashed longitudinal lane marking, but rather provides a description lane marking pattern, polyline boundary, and other corresponding attributes (See Table \ref{tab:mapattributes} of the main text). % \ref{tab:mapattributes}
Thus, we can simulate the pattern of dashed lane markings, but not their exact, pixel-perfect location. As the main text shows, the network can abstract away the per-pixel details to provide more meaningful features. %\cite{Lambert21neurips_TrustButVerifyHDMapChangeDetection}

% \subsection{Guided GradCAM Implementation}

% GuidedGradCAM, elementwise multiplication of 

% ResNet conv4 has shape $\mathbbm{R}^{1 \times C_l \times H_l \times W_l}$ for layer output (Activations) and gradient w.r.t. those activations for a single example.

% GuidedBackProp -- replace ReLU backward by function that passes the upstream gradient where both the layer input is positive, and the upstream gradient is also positive (CITE All-convolutional-net.)

% GradCAM -- cache layer activations and loss gradient w.r.t layer inputs for a specific layer (resnet conv4), generate feature map weights as the mean gradient for each $H_l \times W_l$ feature map, then linear combination of layer activations with aforementioned feature map weights.

\section*{Appendix C: Data Selection}
\begin{figure*}
\centering
\subfloat[Before (WDC)]{
    \includegraphics[width=0.21\columnwidth]{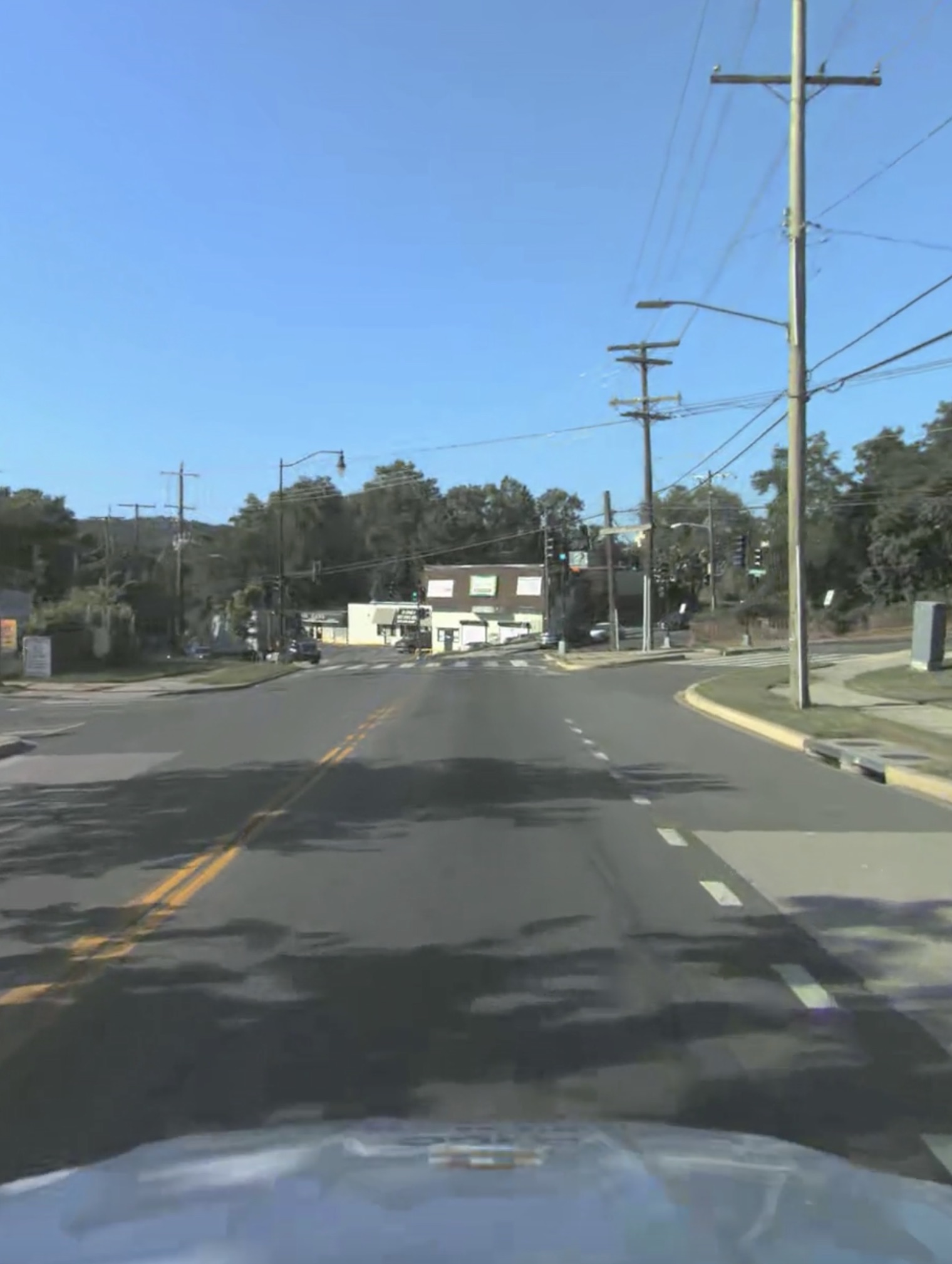} % July 2020
}
\subfloat[After (WDC)]{
    \includegraphics[width=0.21\columnwidth]{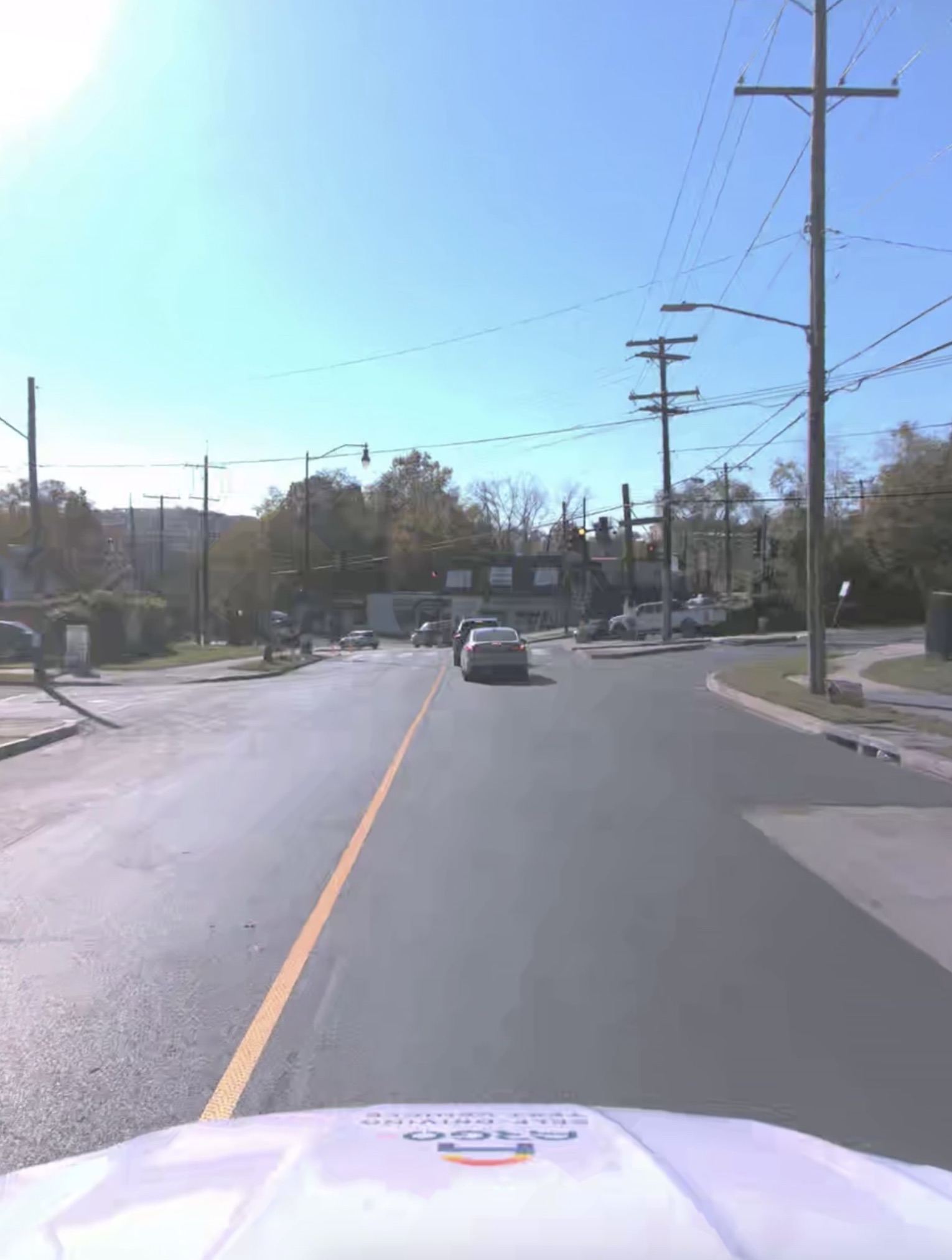}  % Nov. 2020 
}\hspace{10mm}
\subfloat[Before (PIT)]{
    \includegraphics[width=0.21\columnwidth]{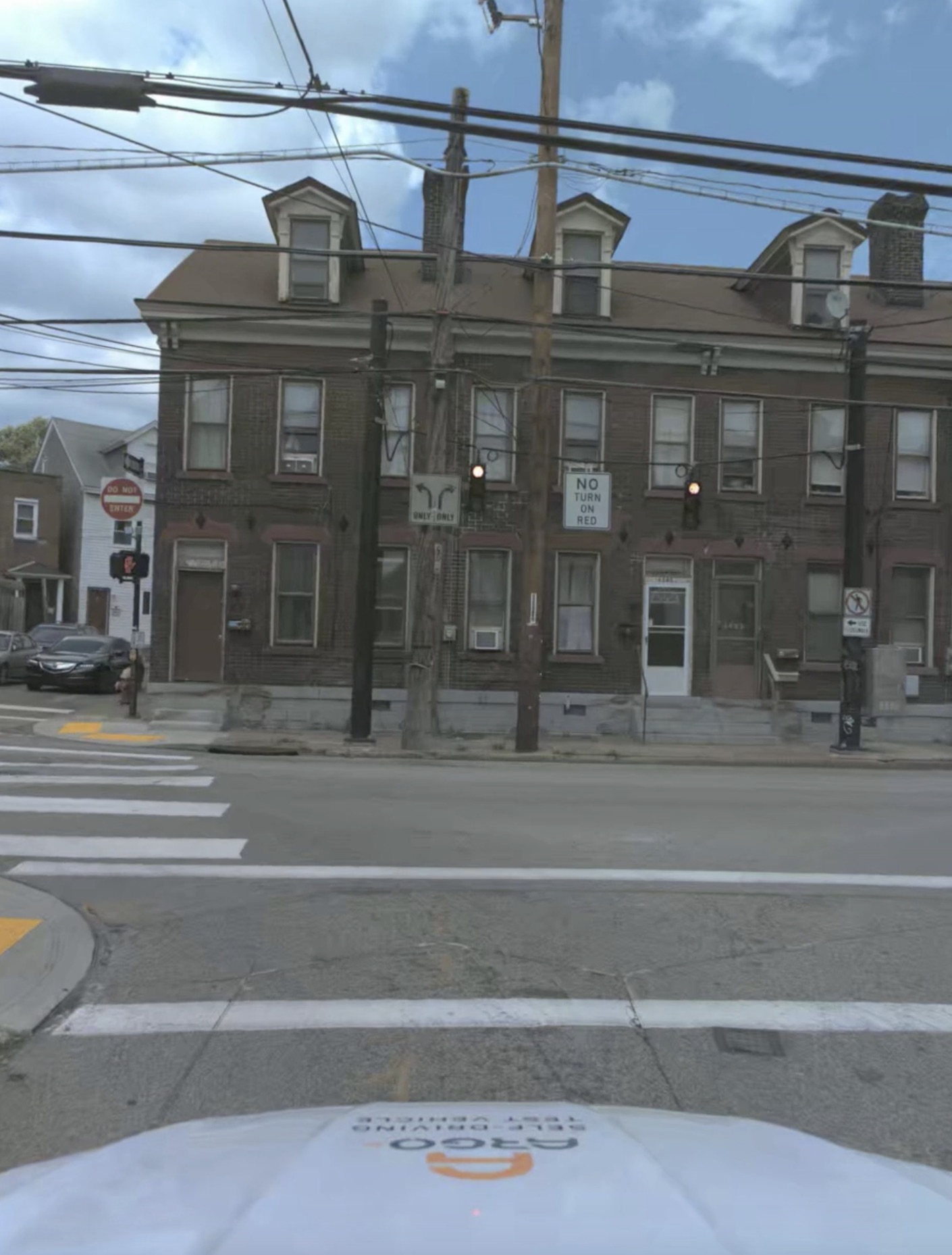}
}
\subfloat[After (PIT)]{
    \includegraphics[width=0.226\columnwidth]{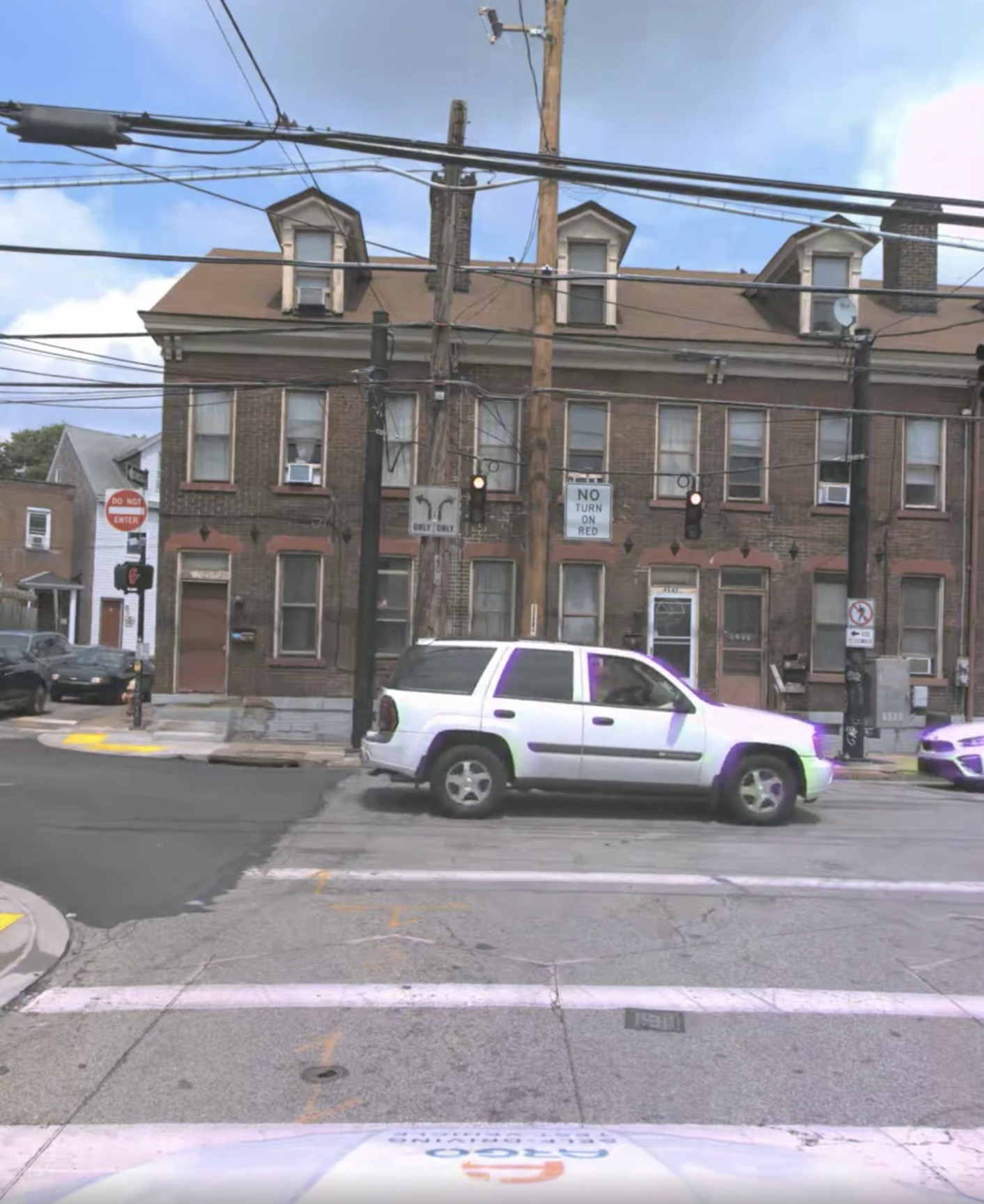}
}
\hspace{2mm}
\subfloat[Before (MIA)]{
    \includegraphics[width=0.202\columnwidth]{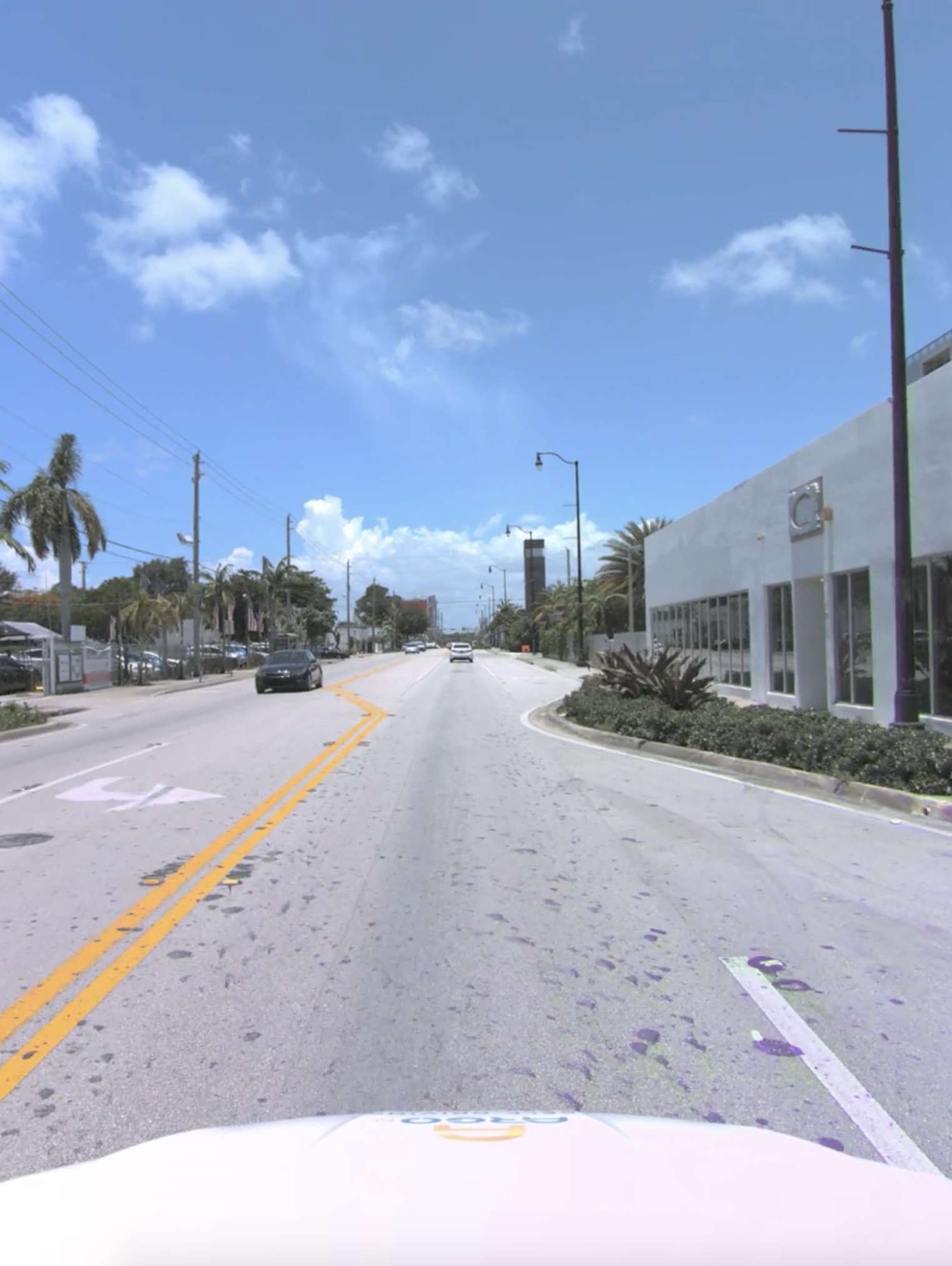}
}
\subfloat[After (MIA)]{
    \includegraphics[width=0.21\columnwidth]{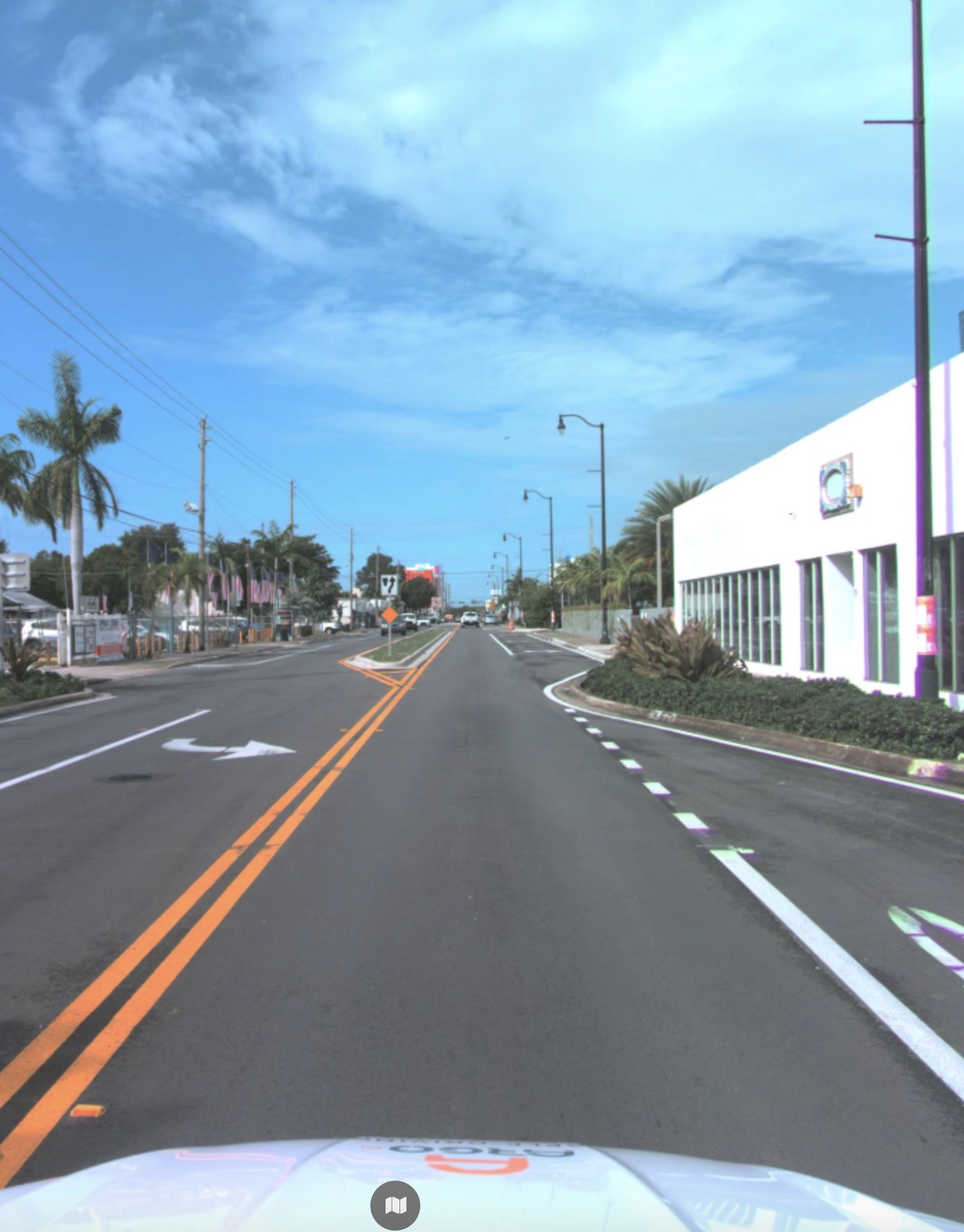}
}\hspace{10mm}
\subfloat[Before (PIT)]{
    \includegraphics[width=0.21\columnwidth]{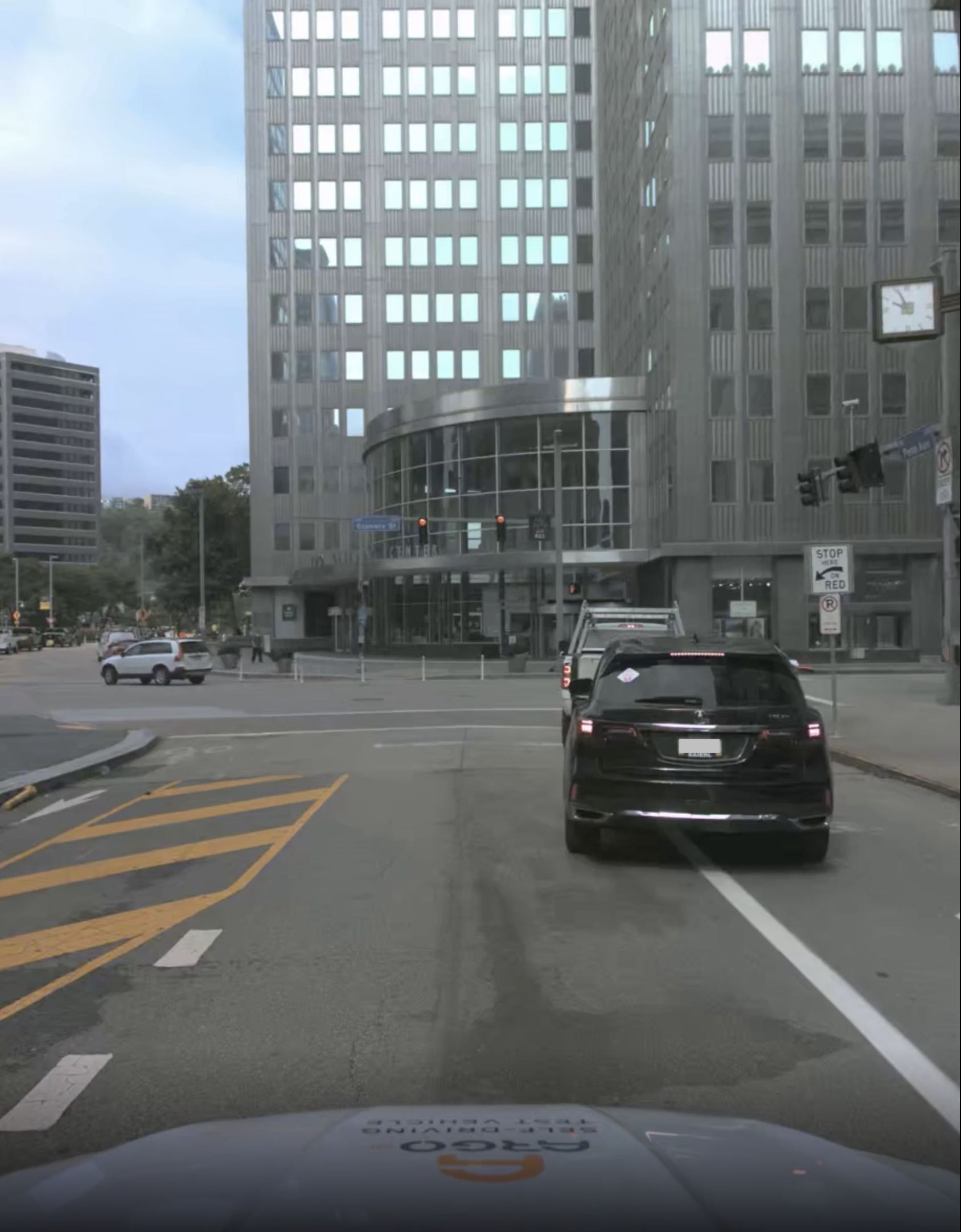}
}
\subfloat[After (PIT)]{
    \includegraphics[width=0.205\columnwidth]{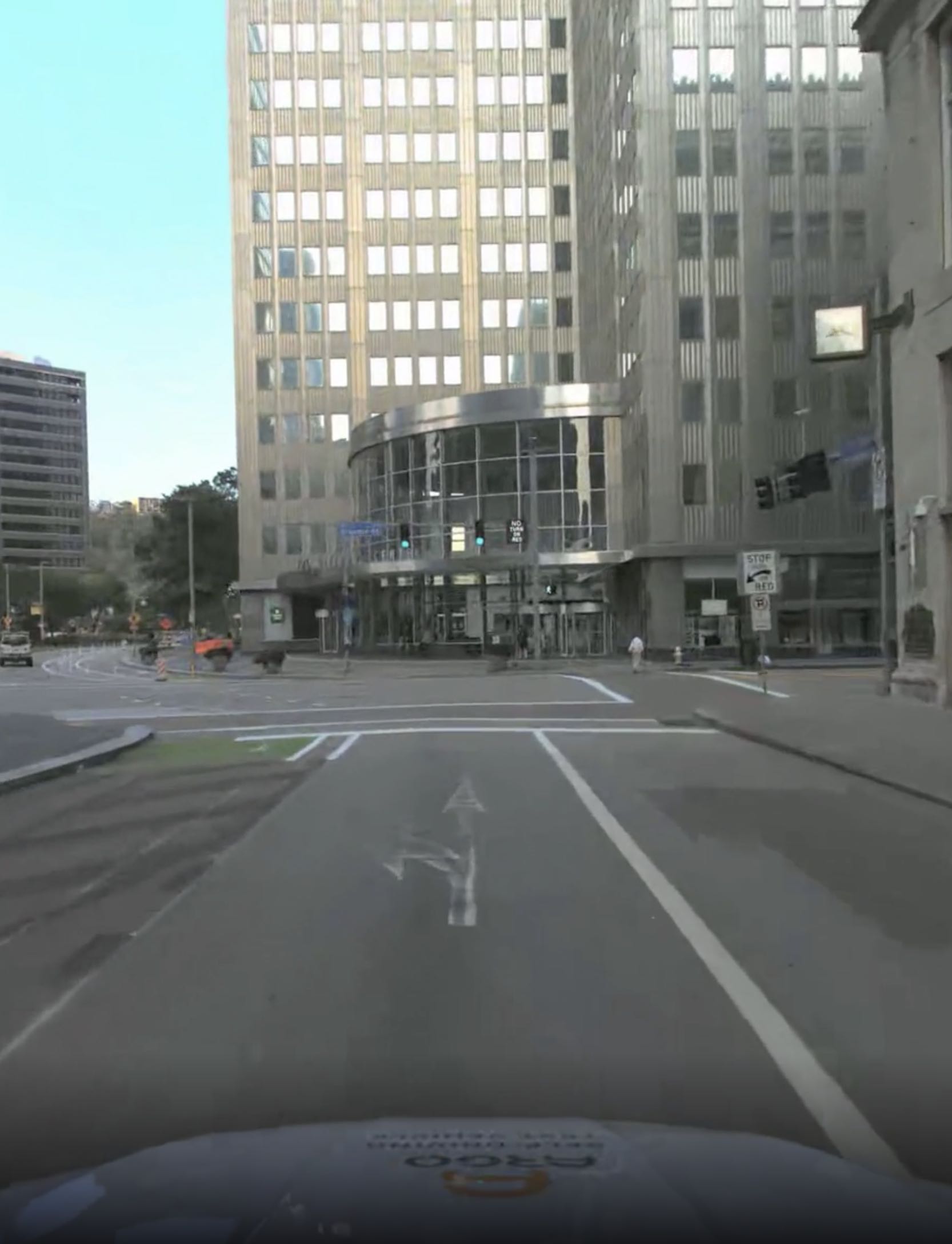}
}
\hspace{2mm}
\subfloat[Before (PAO)]{
    \includegraphics[width=0.21\columnwidth]{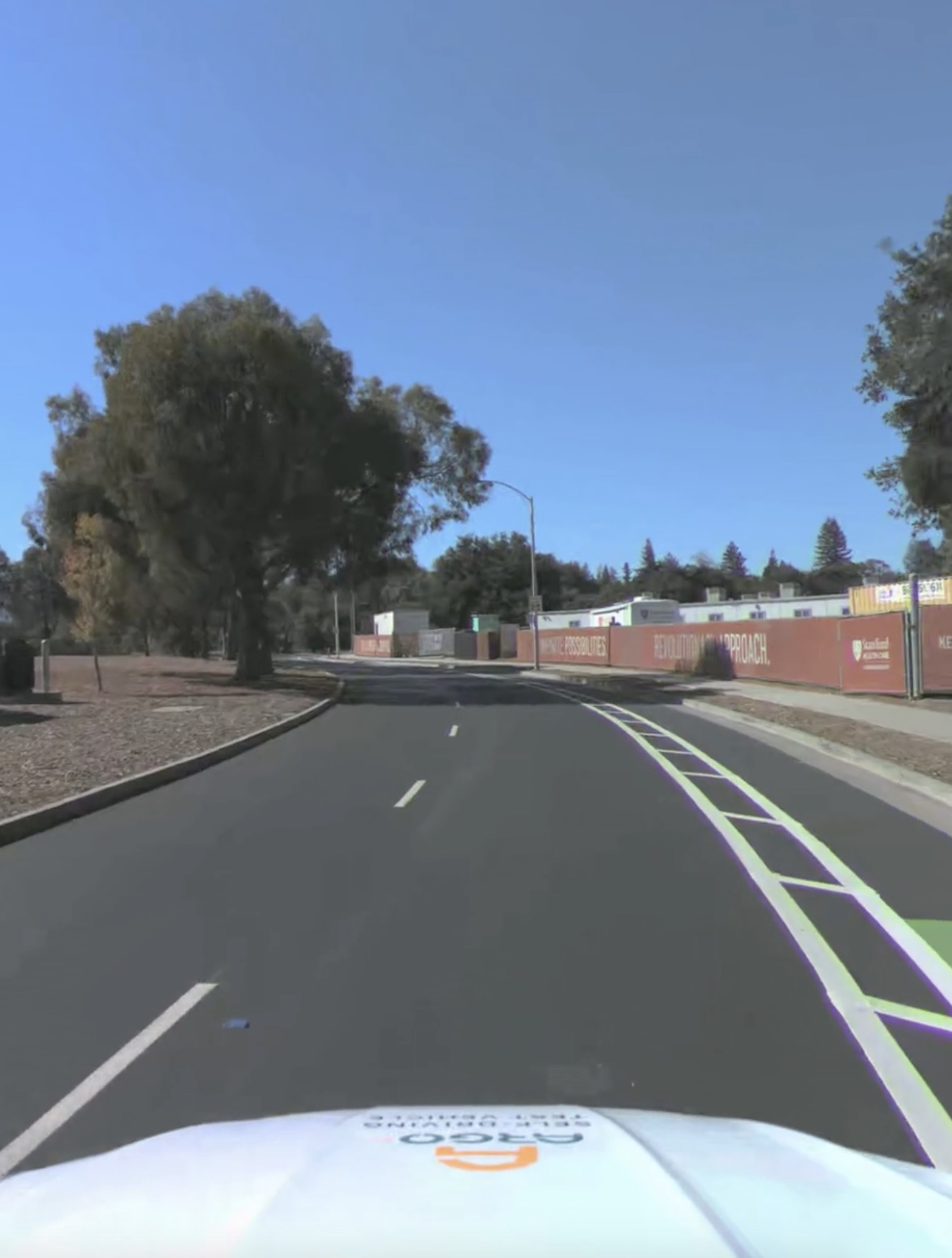}
}
\subfloat[After (PAO)]{
    \includegraphics[width=0.21\columnwidth]{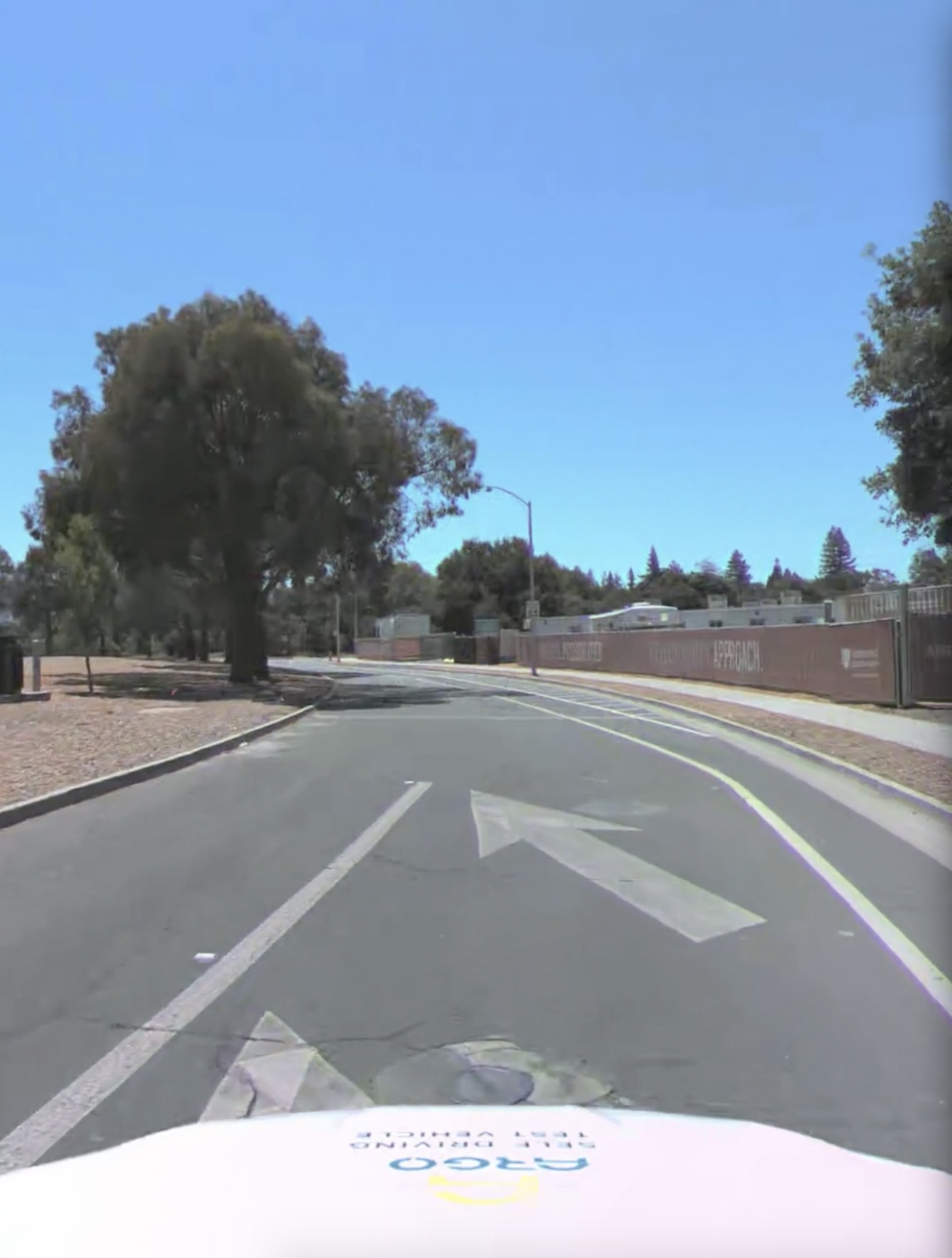}
}\hspace{10mm}
\subfloat[Before (PIT)]{
    \includegraphics[width=0.21\columnwidth]{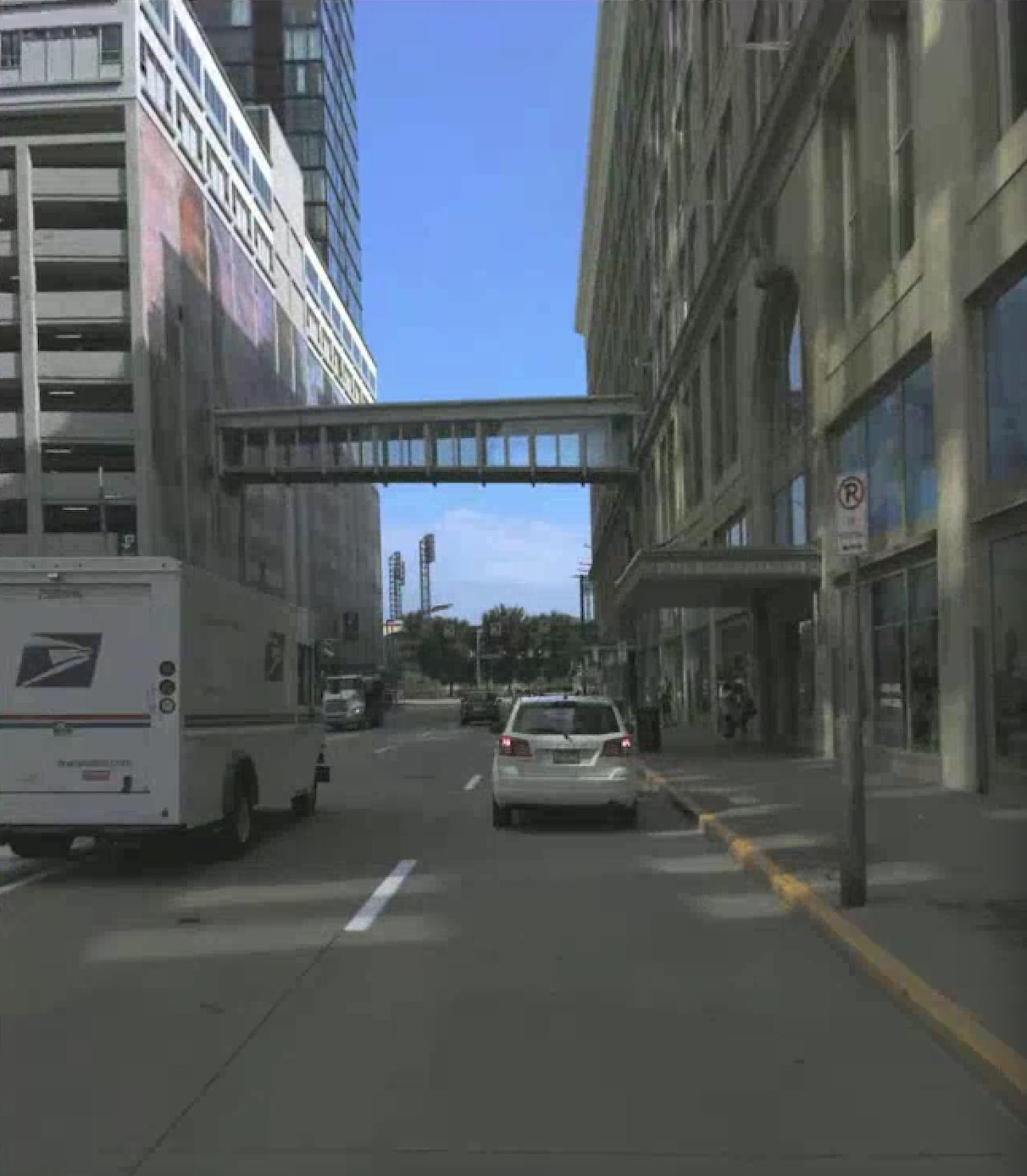}
}
\subfloat[After (PIT)]{
    \includegraphics[width=0.21\columnwidth]{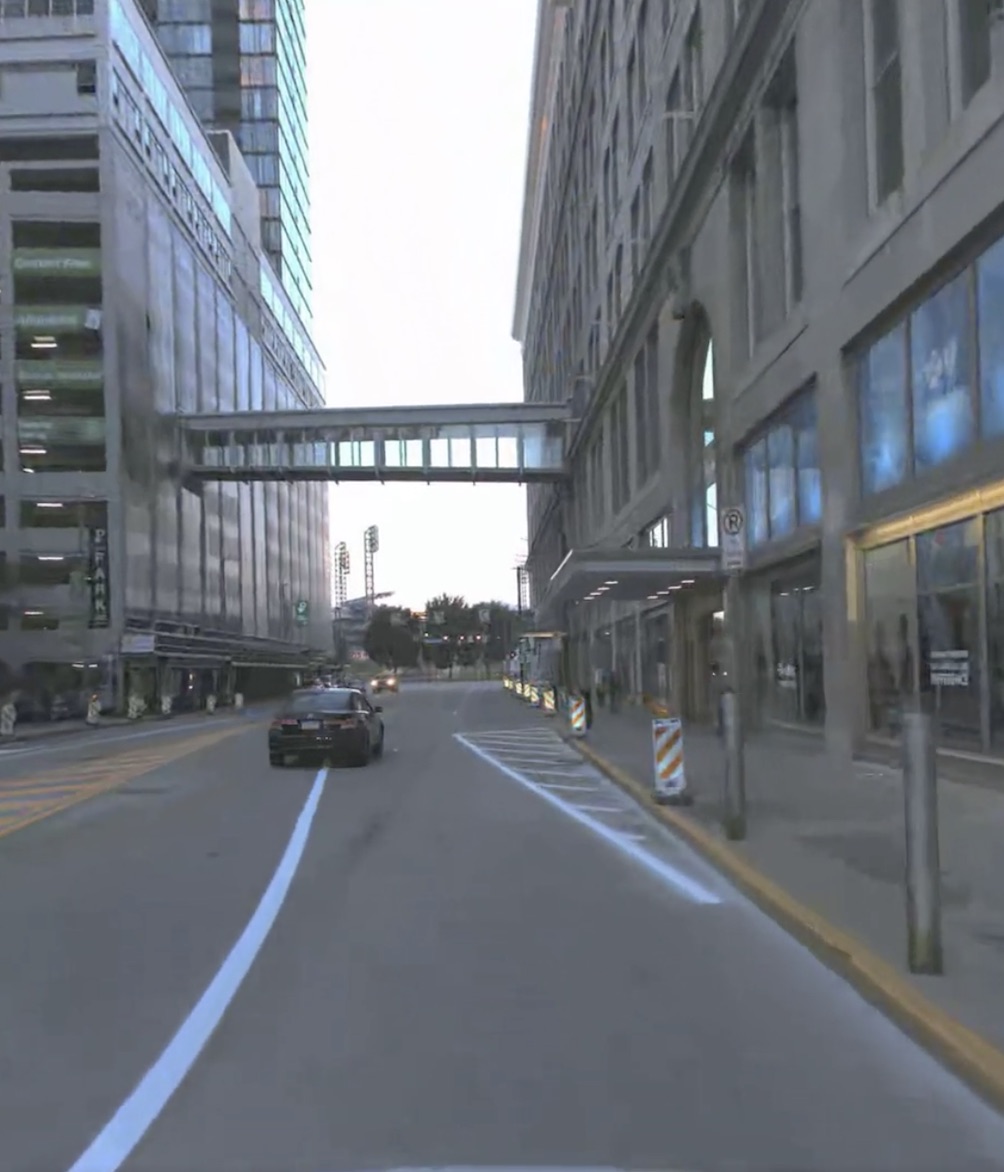}
}
\hspace{2mm}
\subfloat[Before (PIT)]{
    \includegraphics[trim=0 0 0 500,clip,width=0.16\columnwidth]{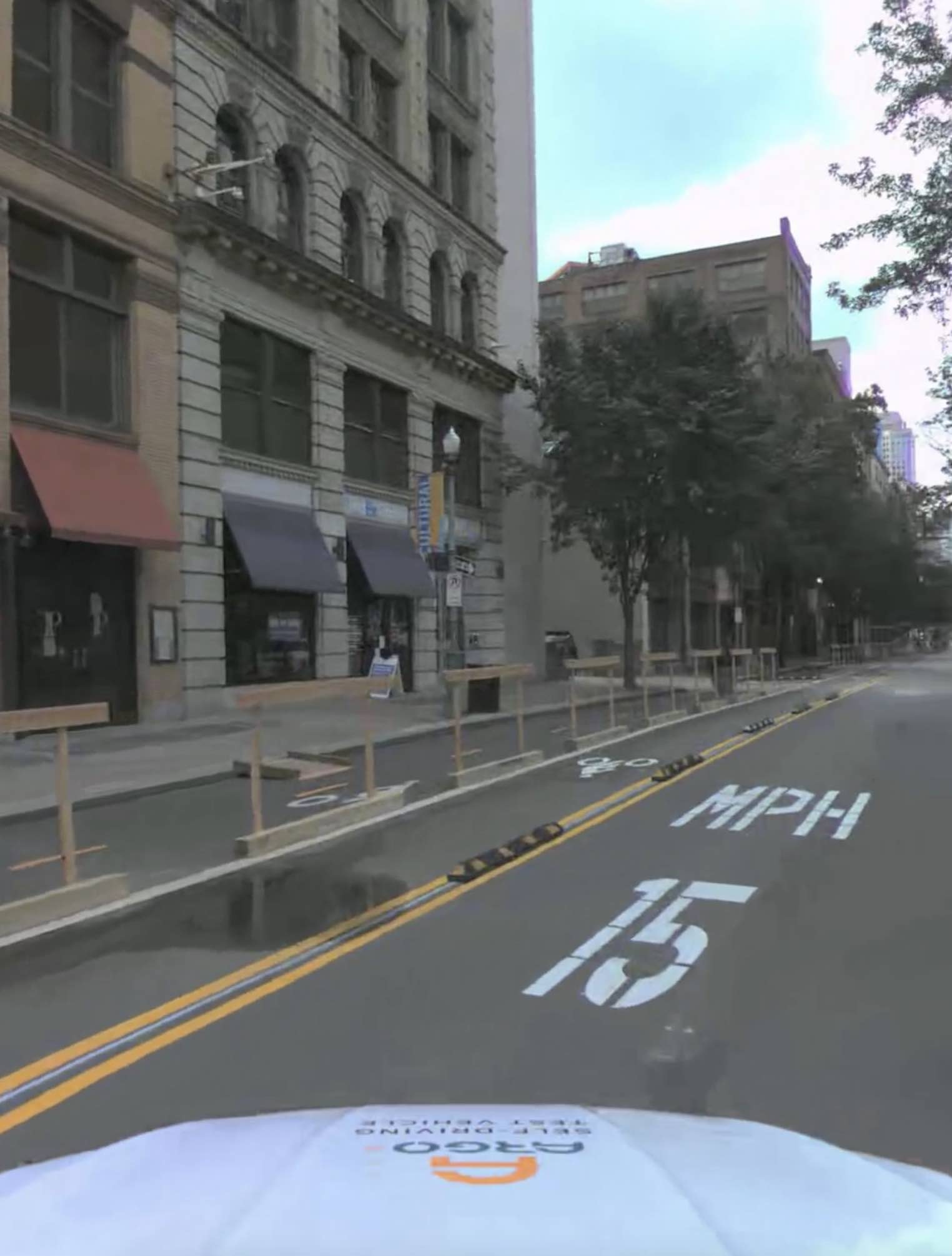}
}
\subfloat[After (PIT)]{
    \includegraphics[width=0.28\columnwidth]{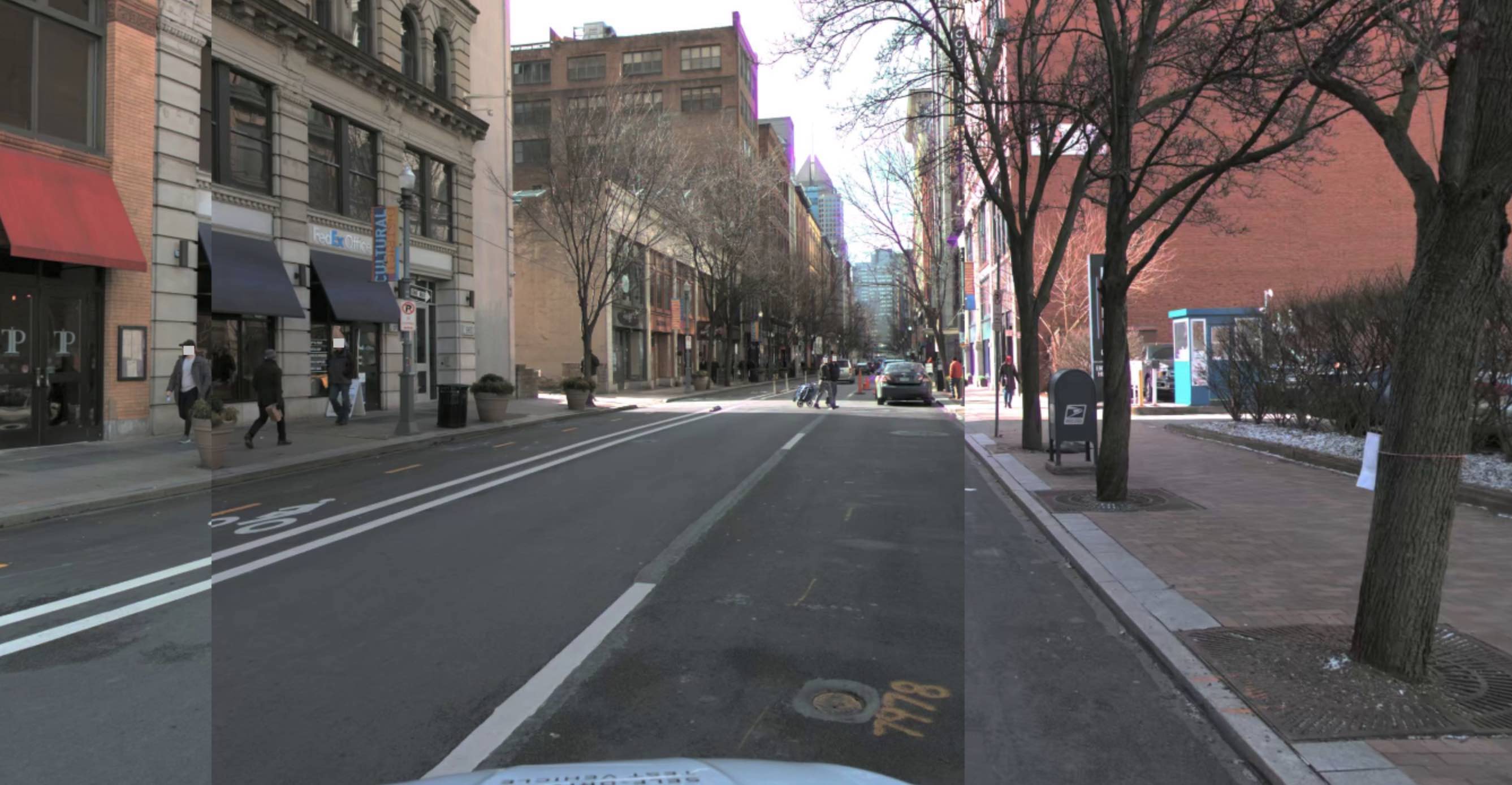}
}\hspace{10mm}
\subfloat[Before (MIA)]{
    \includegraphics[width=0.28\columnwidth]{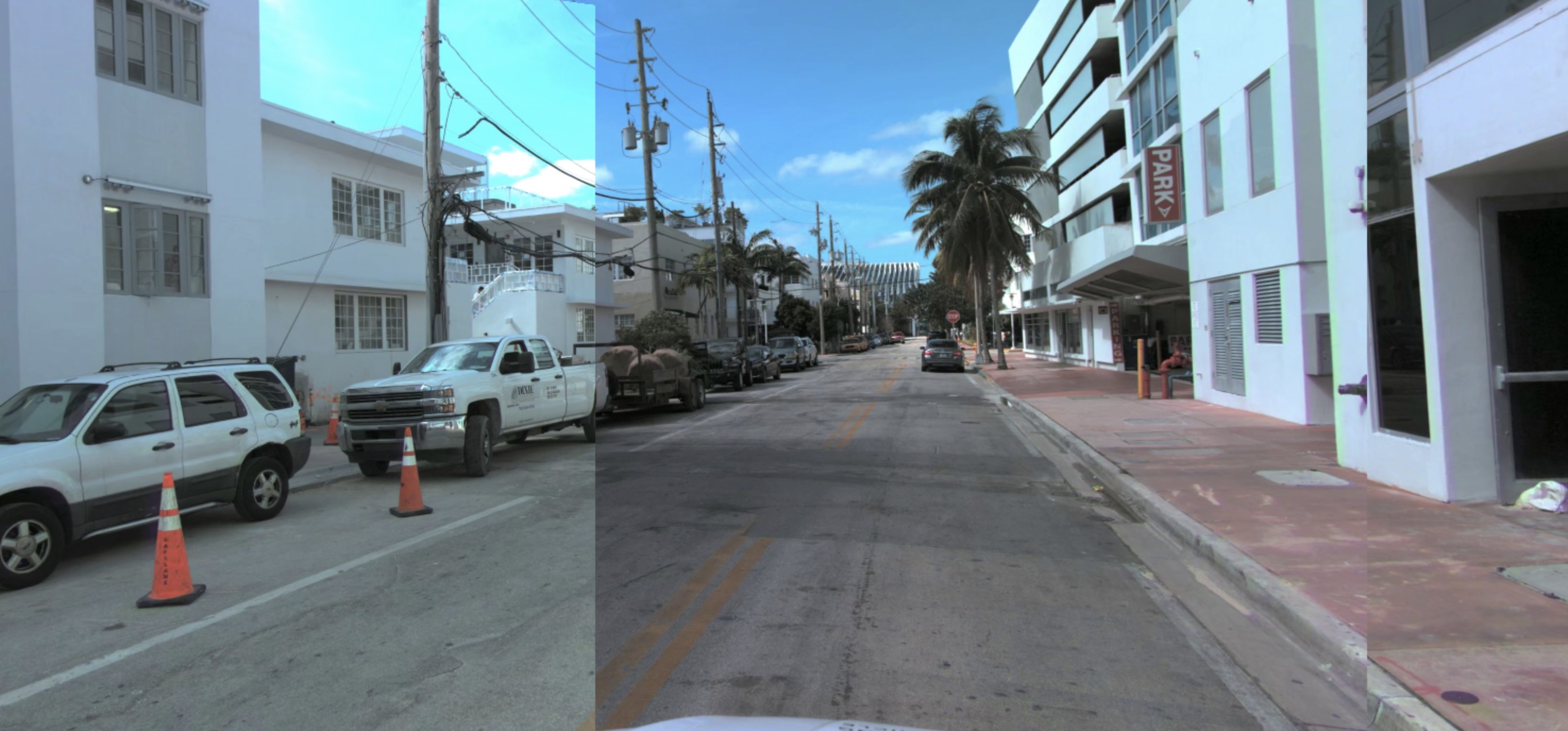}
}
\subfloat[After (MIA)]{
    \includegraphics[trim=0 0 0 500,clip,width=0.13\columnwidth]{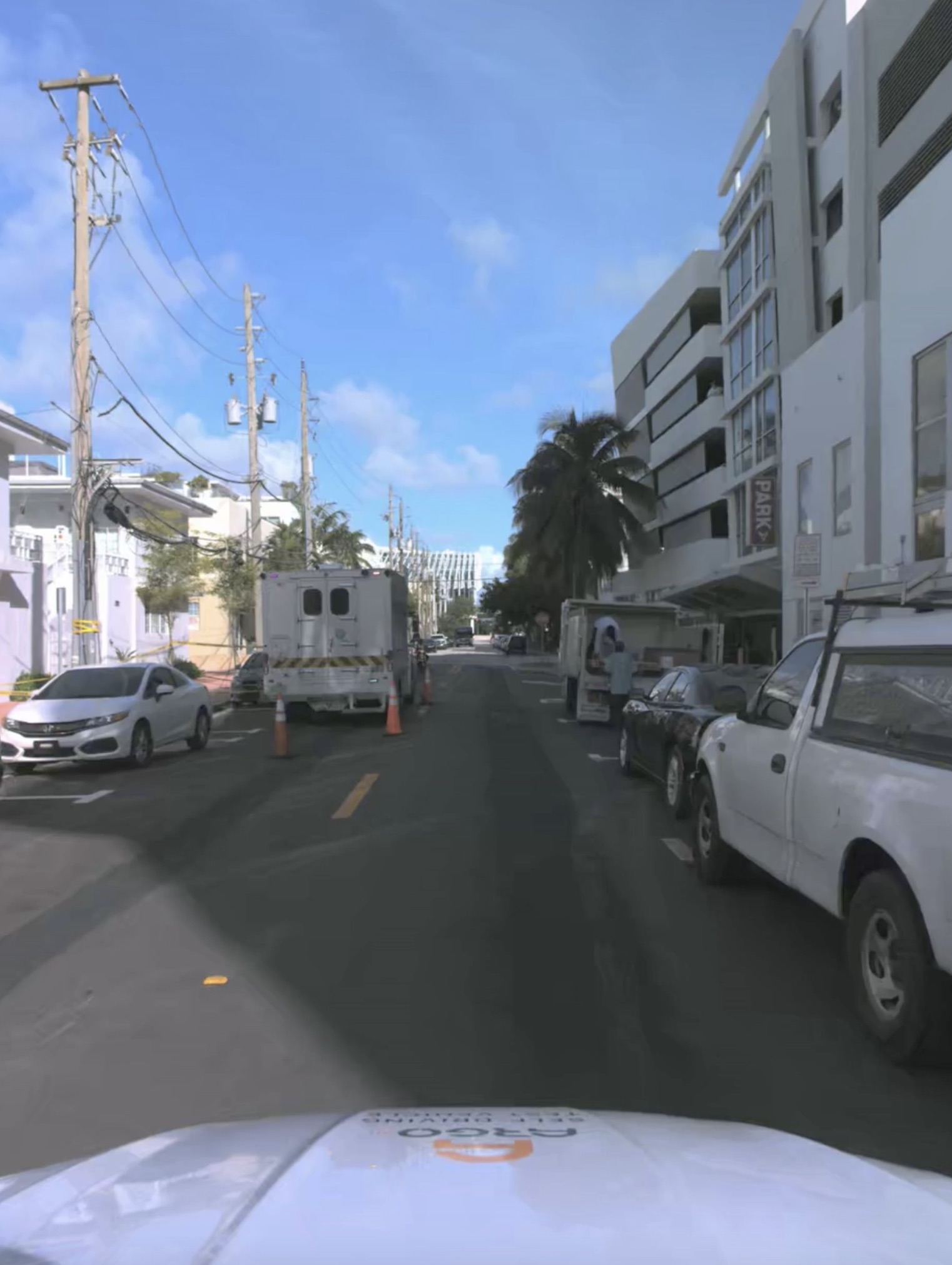}
}
\hspace{2mm}
\subfloat[Before (WDC)]{
    \includegraphics[trim=0 0 0 500,clip,width=0.16\columnwidth]{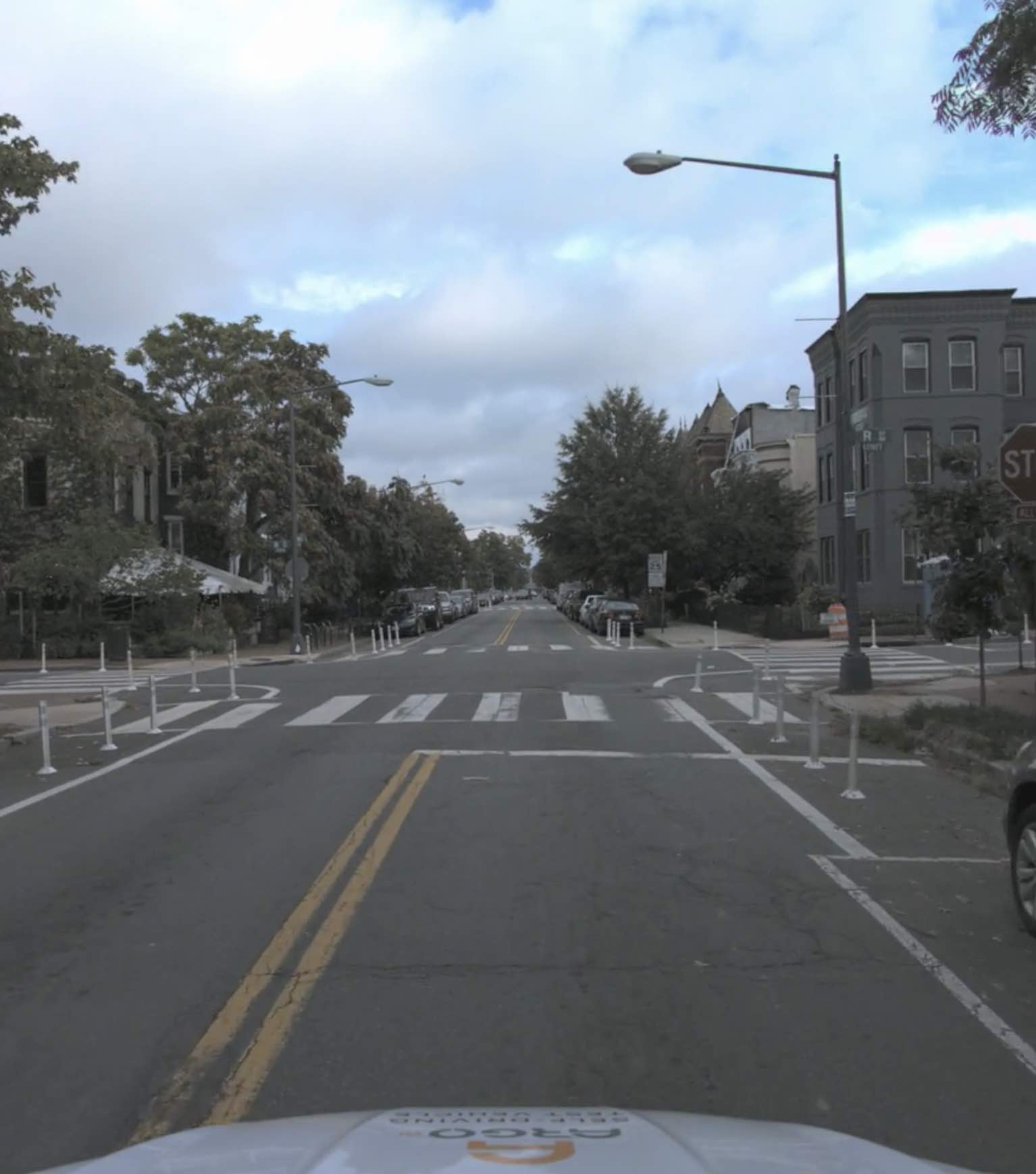}
}
\subfloat[After (WDC)]{
    \includegraphics[width=0.28\columnwidth]{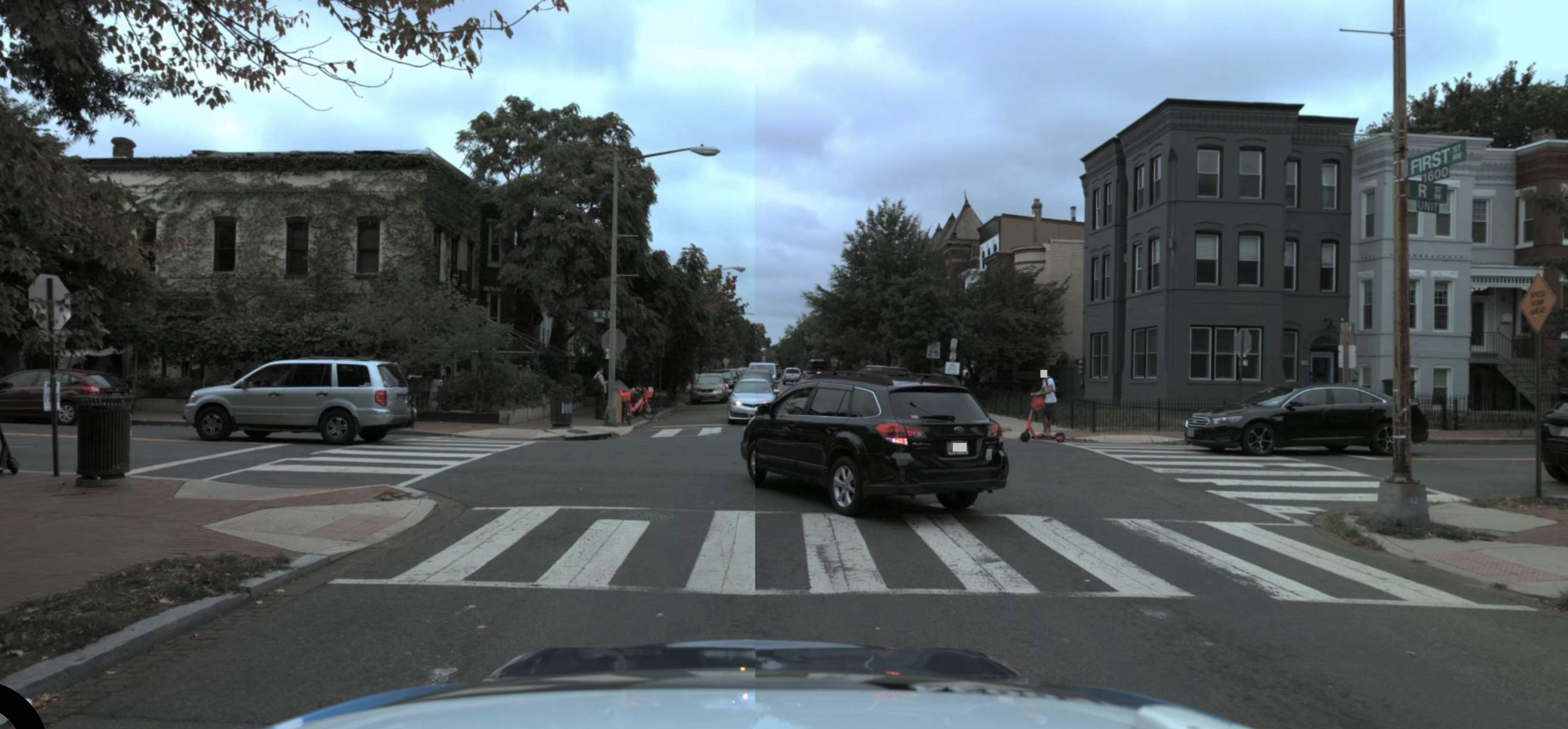}
}
\hspace{10mm}
\subfloat[Before (MIA)]{
    \includegraphics[trim=0 0 0 300,clip,width=0.2\columnwidth]{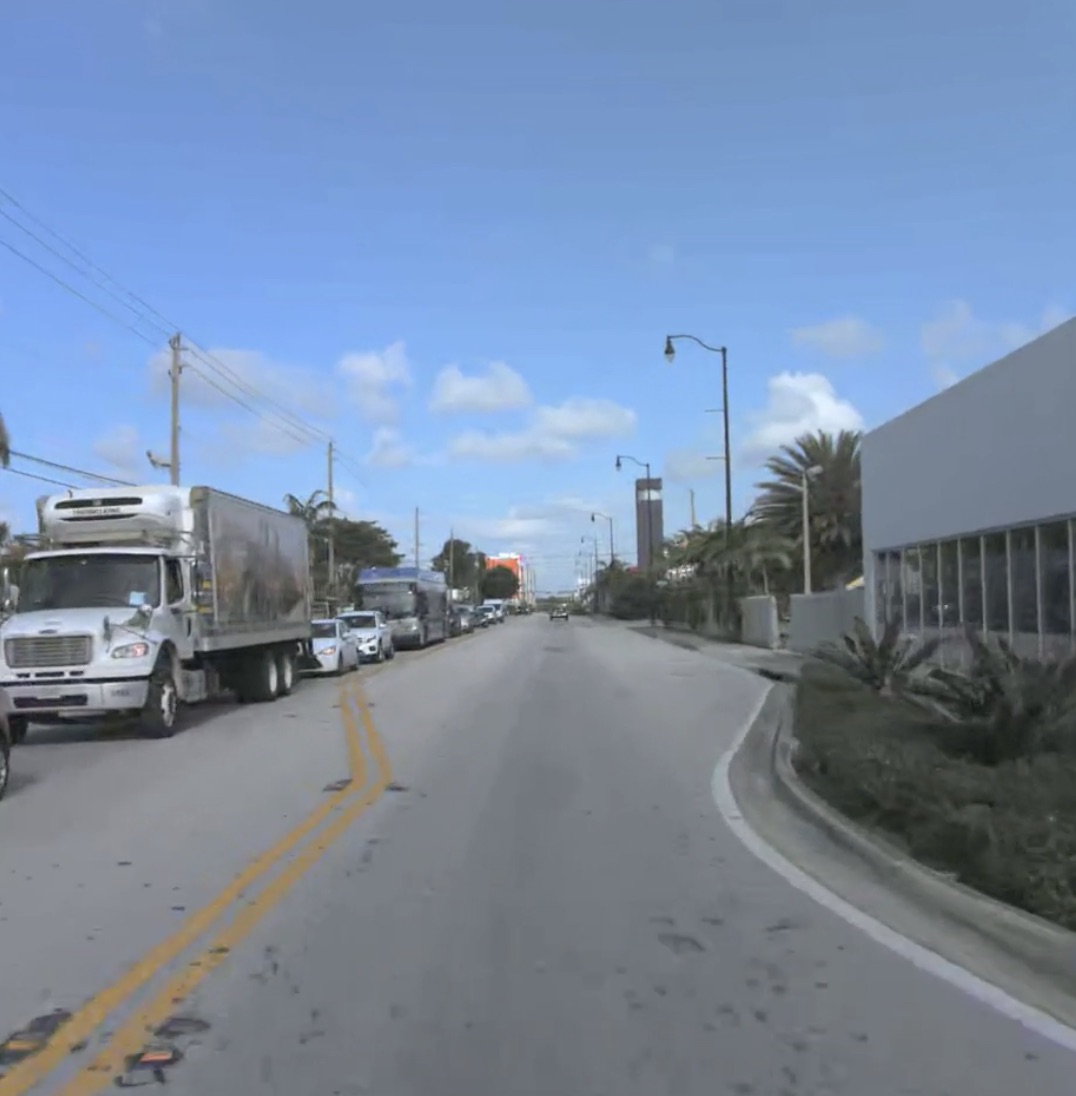}
}
\subfloat[After (MIA)]{
    \includegraphics[width=0.22\columnwidth]{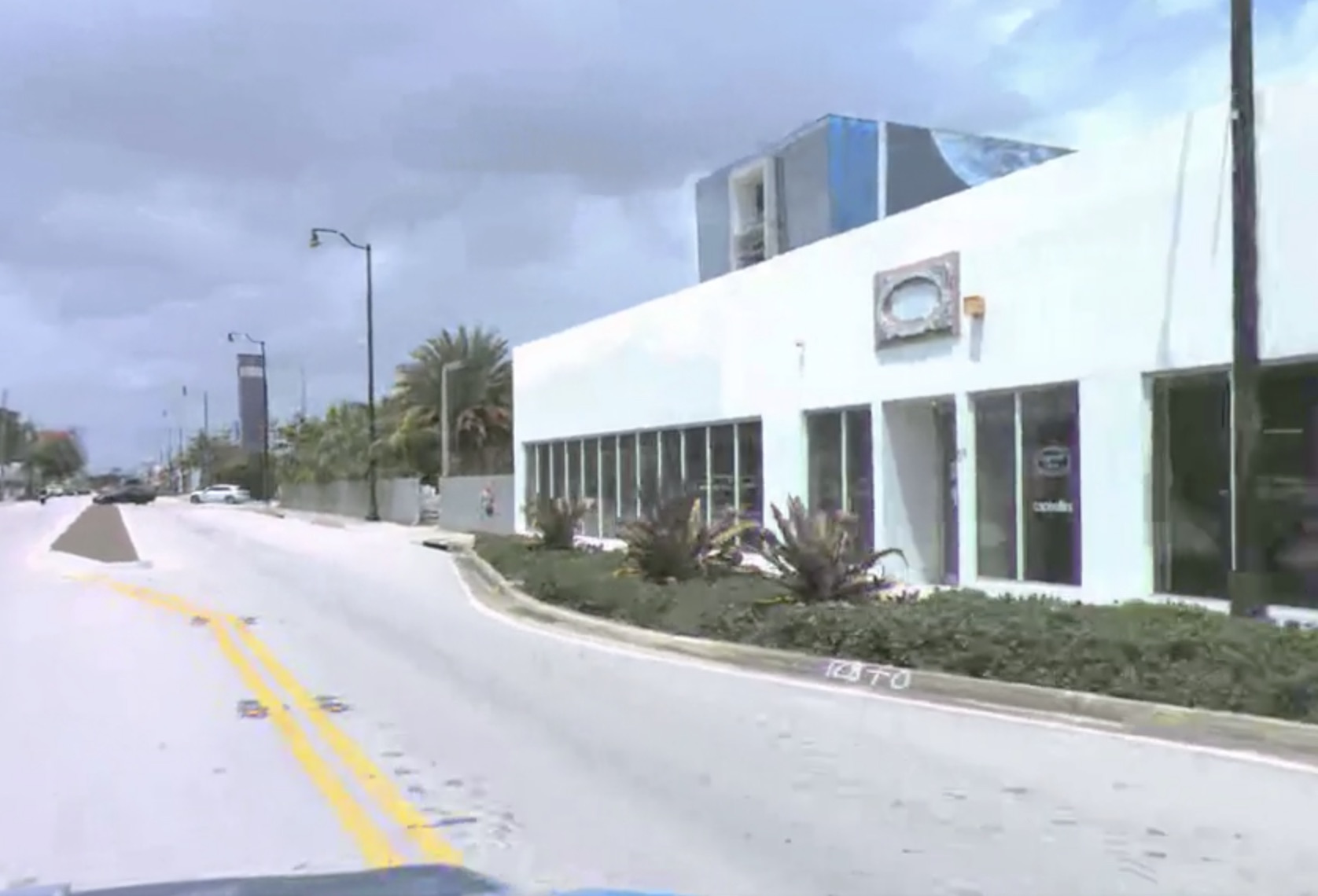}
}
%\vspace{-1em}
\caption{For a number of `negative' logs, our TbV dataset includes corresponding logs captured before the map change was implemented, such that we obtain ``before and after'' imagery. We use the following abbreviations for city names: Washington, DC (WDC), Miami, FL (MIA), Pittsburgh, PA	(PIT), Palo Alto, CA (PAO), Austin, TX (ATX), Detroit, MI (DTW).}
\label{fig:before-and-after}
\end{figure*}

For a subset of the `negative' logs in our TbV dataset, we provide a corresponding `positive' log captured before the change occurred. Example images from pair positive-negative logs are provided in Figure \ref{fig:before-and-after}. This allows for non-learning based approaches (e.g. based upon comparison of 3d reconstructed world models) for a limited amount of the test set.

%%%% before and after figs

\section*{Appendix D: Evaluation}
As our primary accuracy metric, we use a mean of class accuracies over two classes. This accounts for both precision and recall. If a confusion matrix is computed with predicted entries on the rows and actual classes as the columns, and normalized by dividing by the sum of each column, 2-class accuracy can be simply calculated as the mean of the diagonal of the confusion matrix.

More formally, let $n_{cl}=2$ be the number of classes, $\hat{y}_i$ be the prediction for the $i$'th test example, and $y_i$ be the ground truth label for the $i$'th test example. We define per-class accuracy ($\text{Acc}_c$) and mean accuracy (mAcc) as:

\footnotesize
\begin{equation}
\text{mAcc} = \nicefrac{1}{n_{cl}} \sum\limits_{c=0}^{n_{cl}}	 Acc_{c}, \hspace{3mm} \text{Acc}_c = \frac{\sum\limits_{i=0}^N \mathbbm{1}\{ \hat{y}_i = y_i\} \cdot \mathbbm{1}\{ y_i = c \}  }{\sum\limits_{i=0}^N \mathbbm{1}\{y_i = c \} } 
\end{equation}
\normalsize

\section*{Appendix E: Additional Experimental Analysis}

\paragraph{Advantages of BEV.} In principle, the bird's eye view (BEV) representation (orthoimagery) offers two main advantages: a single, dense, accumulated metrically-accurate representation for a single pass through a network, rather than passing in 7 images through 7 separate networks, trained on each frustum, in order to detect changes to the sides and rear of the vehicle. This approach can be costly at inference time given the number of camera frustums required to achieve a panoramic view with traditional cameras. Second, the BEV is generally free of distortion, compared to the ego-view. The ego-view can be seen as ``spoiling'' the map data's metric nature.

\paragraph{Advantages of Ego-view.} However, an ego-view perspective also presents clear advantages over the BEV. Rendering data in the BEV can be seen as ``spoiling'' the sensor data's texture.  Importantly, there is less distraction and less overall content to reason about in the egoview. Therefore, the ego-view task is arguably easier than the BEV task, needing only to detect changes in a $85^{\circ}$ f.o.v. instead of $360^{\circ}$ f.o.v.

\paragraph{Analysis of Map-Only Baseline.} The map-only baseline performs quite poorly when predicting real-world lane geometry changes, slightly over random chance (2\% or 3\% over random chance in the ego-view and 7\% over random change in the BEV). While the map-only stream may seem doomed to fail without access to real-world sensor information, we observe that a certain number of map changes exist to bring the real world into compliance with certain priors, which are already encapsulated in the map. For example, we find that upgrading a 4-way intersection from a single crosswalk to 4 crosswalks, or from a single crosswalk to 0 crosswalks (after repaving) is a common map change, which would agree with priors. Indeed, our experimental results suggest that the map-only baseline, which is completely blind to the real-world, can occasionally succeed at predicting real-world crosswalk changes by learning powerful priors. Inspection via Guided GradCAM demonstrates that the map-only models attends to asymmetric paint patterns along the left and right boundaries of a road, or asymmetric lane subdivisions along two sides of a road; modifications to such map asymmetry which are common real-world map updates.

% However, it is interesting to note that it performs , which may show lane geometry is occasionally updated to align with priors for more safe or canonical road geometry. 

\paragraph{Analysis of Sensor-Only Baseline.} The sensor-only model (see Table \ref{tab:results-modalities} of the main paper) sees randomly perturbed labels, with only ``positive'' training data, and therefore is not a meaningful baseline.
% {tab:results-modalities}

\section*{Appendix F: Orthoimagery Generation Implementation Details}

In this section, we provide additional details about the orthoimagery generation process described in Sections \ref{sec:tbv-sensor-data-repr} and  \ref{sec:impl-details} of the main text. In order to create a metrically-accurate sensor data representation that is free of perspective distortion, we generate orthoimagery using ray-casting. Orthoimagery from LiDAR suffers from extreme sparsity, leading to an impoverished representation. To generate dense panoramic orthoimagery, we use a set of high-definition camera sensors with a panoramic field of view, mounted onboard an autonomous vehicle. We generate the BEV representation (i.e. orthoimagery) by ray-casting image pixels to a ground surface triangle mesh. Our ground height maps exploit LiDAR offline, and in this way our ego-view method incorporates the strengths of LiDAR.

\paragraph{CUDA Ray-Casting Routine.}  We tesselate quads from a ground surface mesh with 1 meter resolution to triangles; rays are cast to triangles up to 25 m away from the egovehicle. For acceleration, we cull triangles outside of the left and right cutting planes of each camera's view frustum. We implement the Moller-Trombore ray-triangle intersection routine \cite{Moller97jgt_FastRayTriangle} in CUDA. 

\paragraph{Density.} Ray-casting yields a vastly more dense set of image rays than LiDAR, on the order of 2 orders of magnitude greater density; for a $1550 \times 2048$ image, one can obtain $\sim 3.17$ million rays per image, and across 7 camera frustums, this translates to over 22.19 million rays with available RGB values per second. With 20 fps imagery per camera frustum, this amounts to 440 million rays per second. Most conventional 10 Hz LiDAR sensors can provide little more than 100k returns per sweep, and thus at most 1 million rays per second. %Mosaicing together pixels from various camera frustums can lead to significant artifacts due to differing luminance. We perform an ablation study of the benefit of histogram matching between per-frustum pixel distributions. 

\paragraph{Aggregation.} In order to prevent holes in the orthoimagery in the area underneath the egovehicle, we aggregate pixels in ring buffer of length 10 sweeps, and wait 10 sweeps before starting rendering. Future sensor data is not used to render the sensor data representation.  We use linear interpolation to account for sparsity at range.

%We compare results with or without dynamic object removal. In order to remove dynamic objects, we use semantic segmentation label masks generated by the MSeg model \cite{Lambert20cvpr_MSeg}. 

\paragraph{Comparison with IPM.} While Inverse Perspective Mapping (IPM) is the dominant approach in the literature, it is inaccurate as it cannot account for ground surface variation. Geiger \cite{Geiger09ivs_MonocularRoadMosaicing} model the image-to-ground plane mapping as a homography (IPM) and mosaics together monocular images, but requires scenes with an approximately-planar ground surface. Zhang \emph{et al.}\cite{Zhang14icrb_LaneLevelOrthophoto} generate orthophoto ground imagery using fisheye cameras and IPM. Rapo \cite{Rapo18thesis_orthoimagery} explored the use of dashboard-mounted cell phones without access to LiDAR or known calibration, instead relying upon SfM, optical flow, and vanishing point estimation for online calibration and also use IPM for pixel-to-world correspondence.

\section*{Appendix G: Additional Examples from Test Set}

\begin{figure}
 \vspace{-2mm}
 \centering
\subfloat[]{
    \includegraphics[width=0.47\columnwidth]{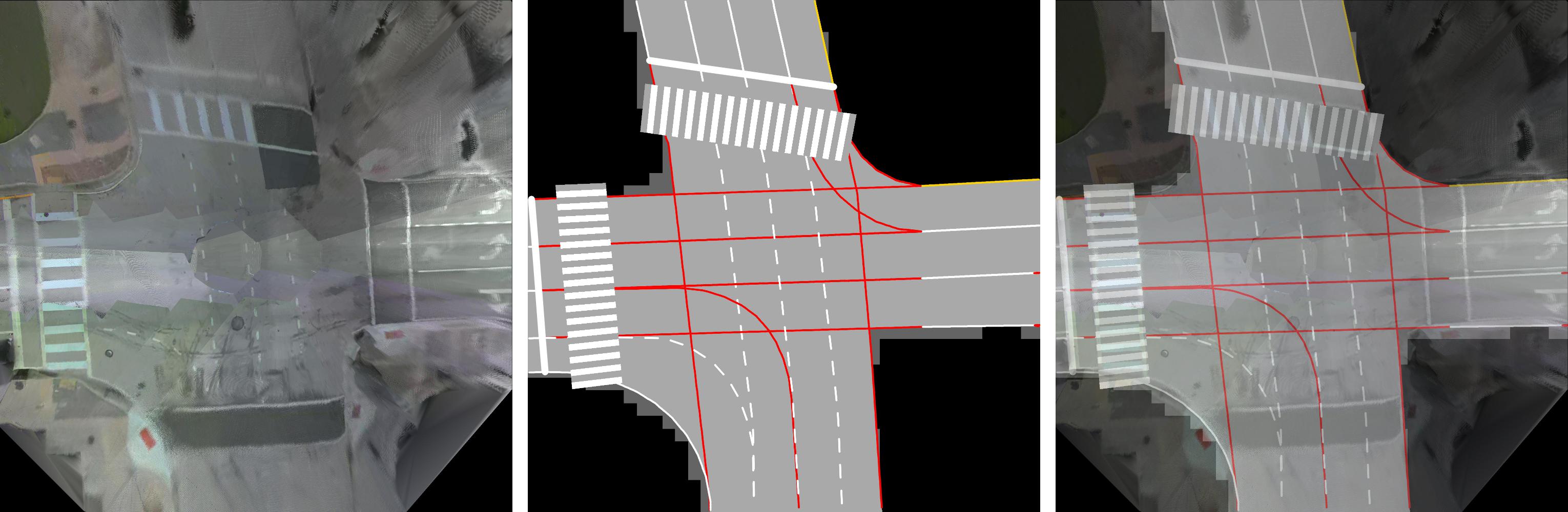}
}
\hspace{2mm}
\subfloat[]{
    \includegraphics[width=0.47\columnwidth]{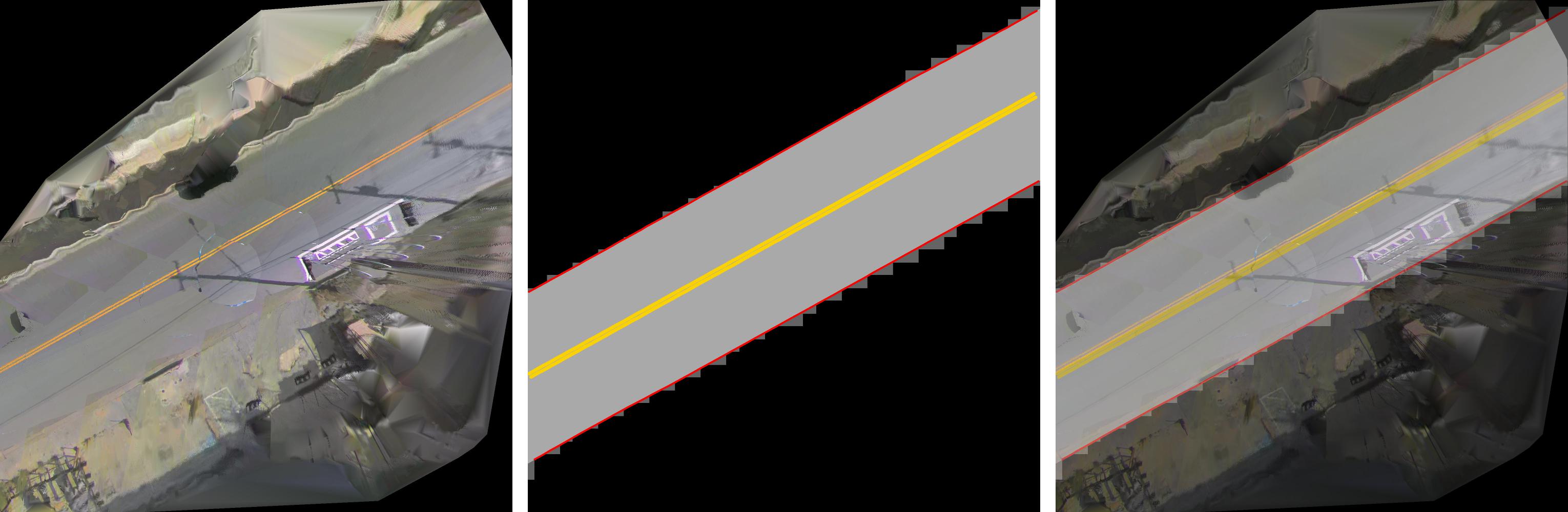}
}
\hspace{0mm}
%  \begin{tabular}{c}

\subfloat[]{
    \includegraphics[width=0.47\columnwidth]{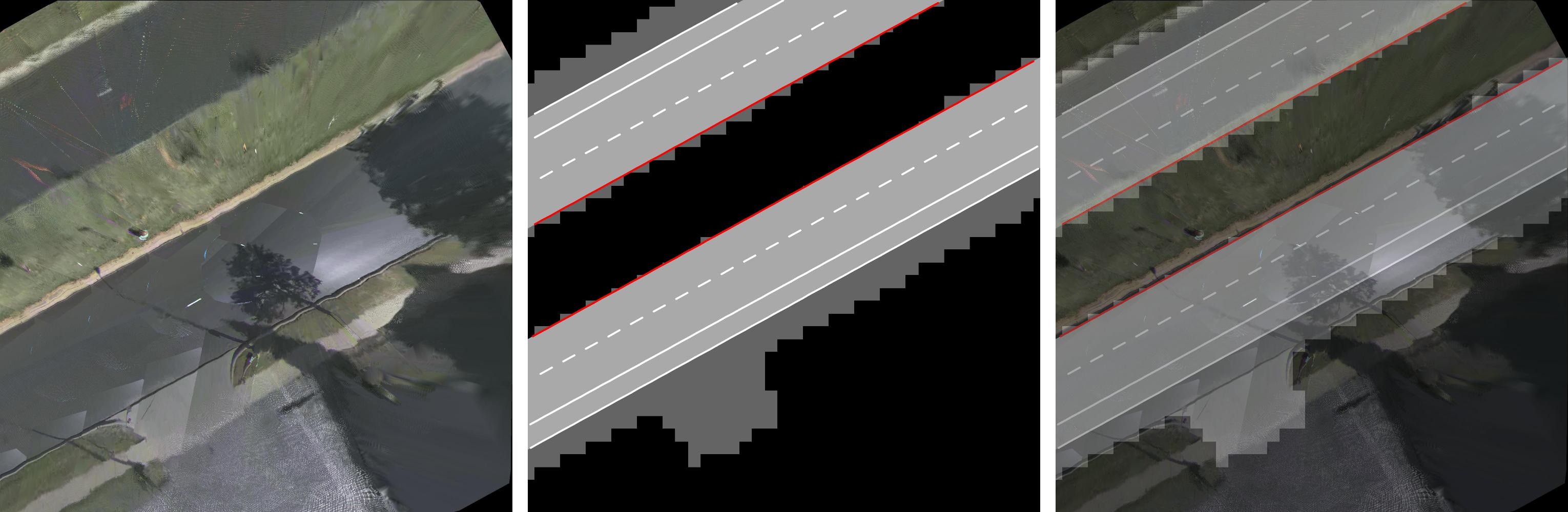}
}
\hspace{2mm}
\subfloat[]{
    \includegraphics[width=0.47\columnwidth]{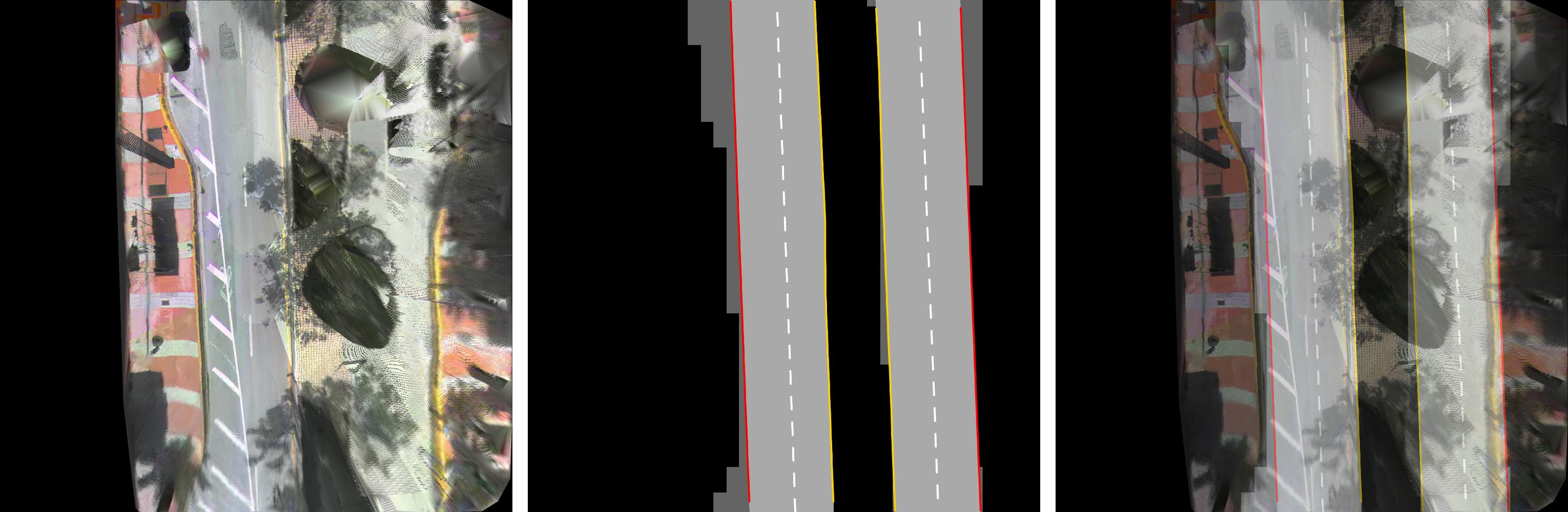}
}
\hspace{0mm}
\subfloat[]{
    \includegraphics[width=0.47\columnwidth]{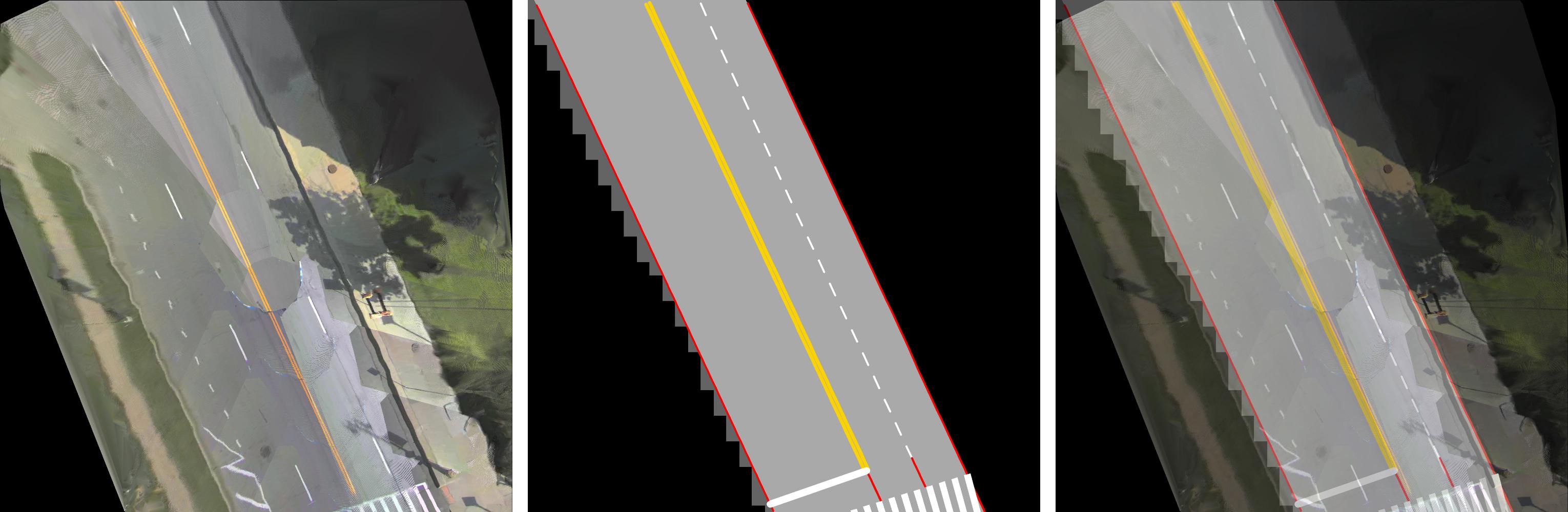}
}
\hspace{2mm}
\subfloat[]{
    \includegraphics[width=0.47\columnwidth]{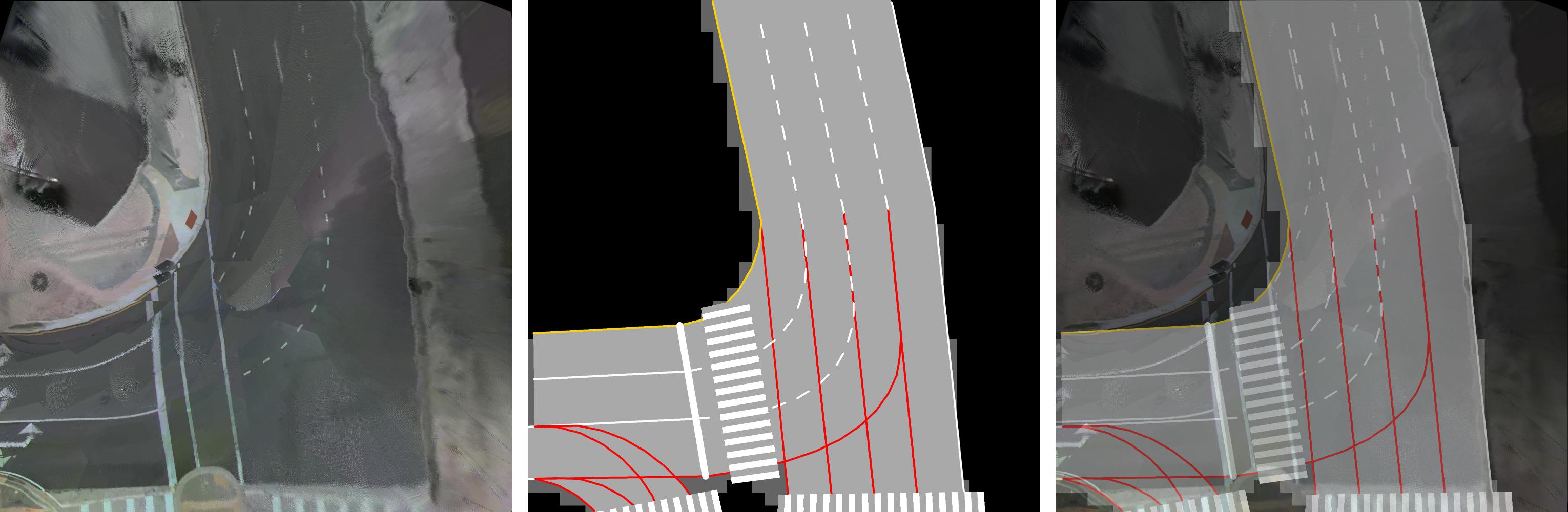}
}
\hspace{0mm}
\subfloat[]{
    \includegraphics[width=0.47\columnwidth]{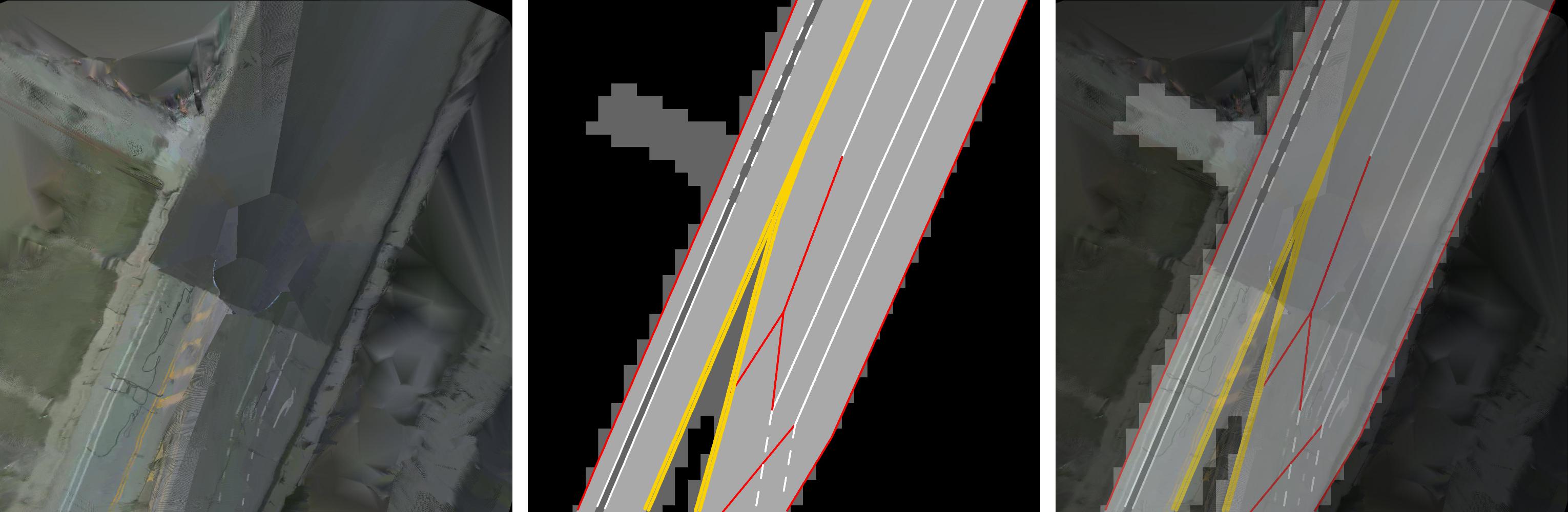}
}
\hspace{2mm}
\subfloat[]{
    \includegraphics[width=0.47\columnwidth]{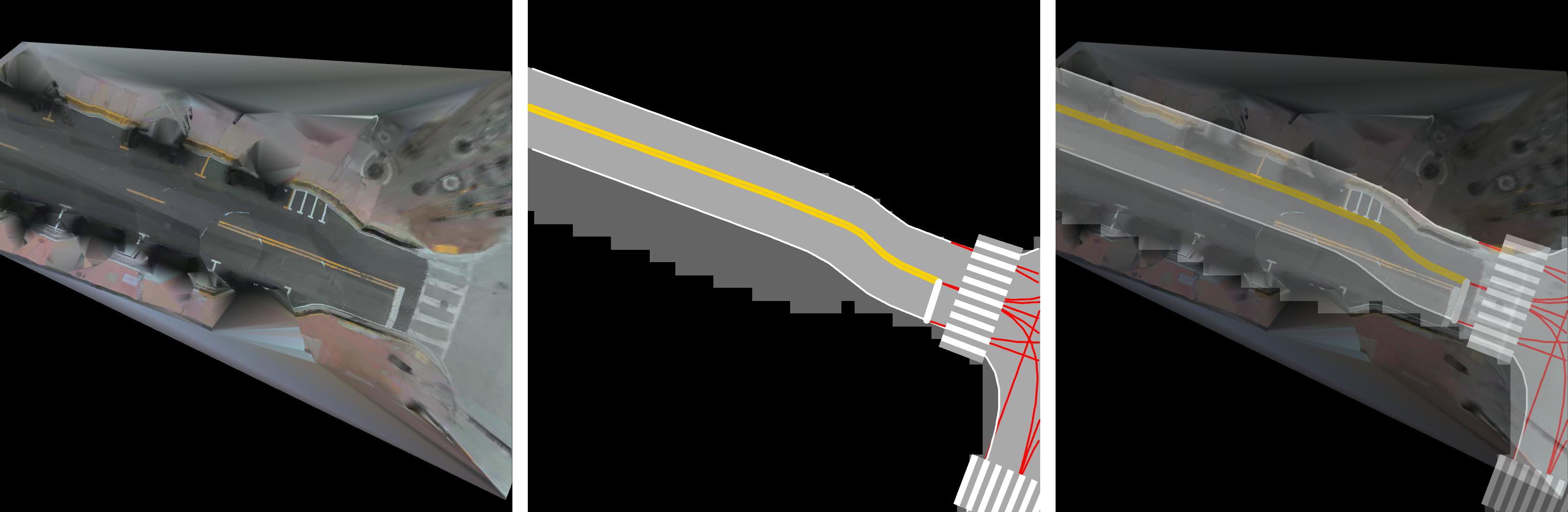}
}
% \end{tabular}
\caption{Examples from the test split of our TbV dataset. Left to right: BEV sensor representation, onboard map representation, blended map and sensor representations. Inset (A) depicts inserted crosswalks, while insets (B-H) represent painted lane geometry changes. }
\label{fig:additionaltestsetexamples}
\end{figure}

In Figure \ref{fig:additionaltestsetexamples}, we show additional examples from our test set, as seen from a bird's eye view.

\section*{Appendix H: Map Changes from Construction}

\begin{figure*}
\centering
\subfloat[Traffic Cones]{
\includegraphics[height=2.5cm]{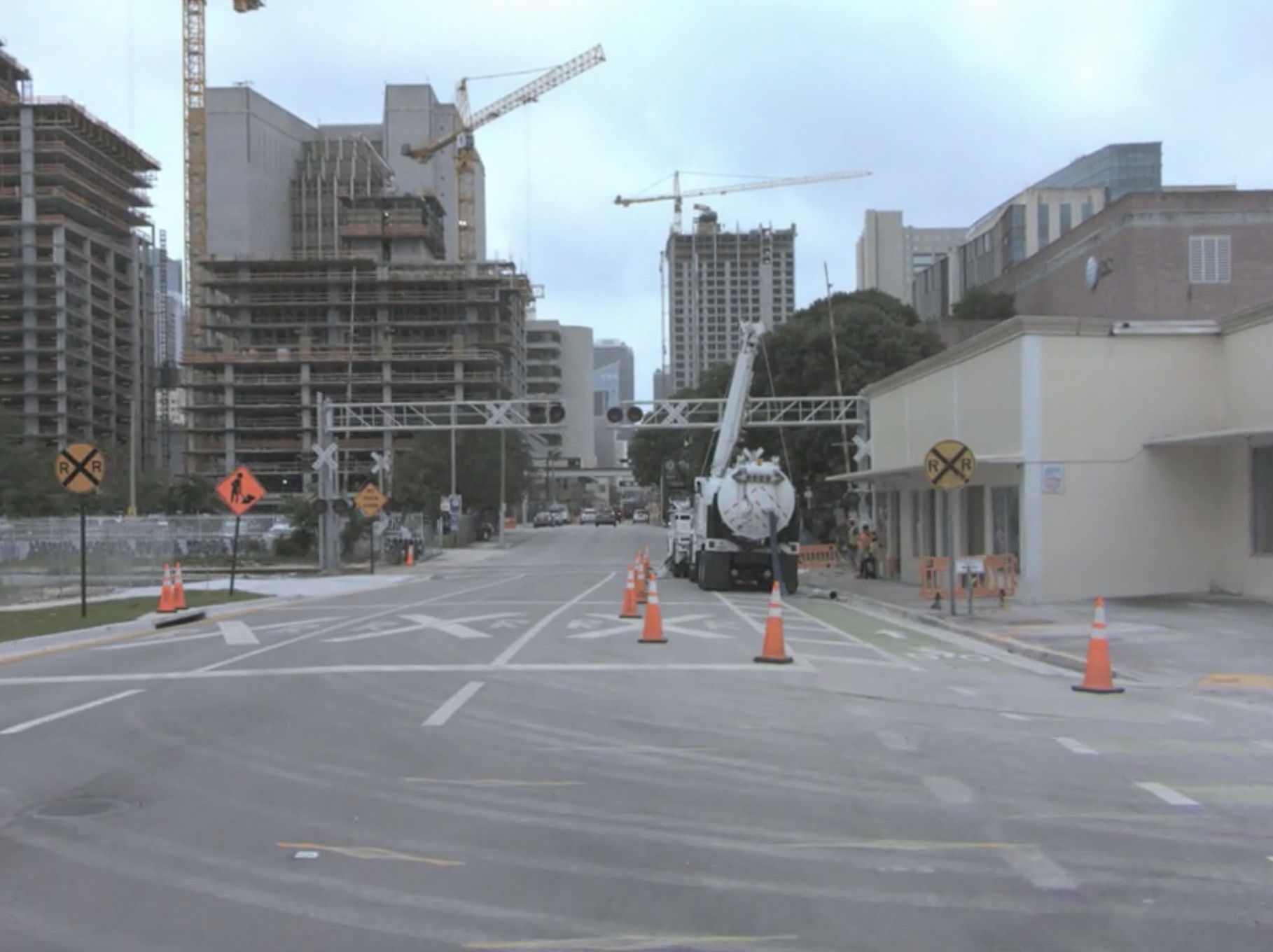}
}
\subfloat[Jersey Barriers]{
\includegraphics[height=2.5cm]{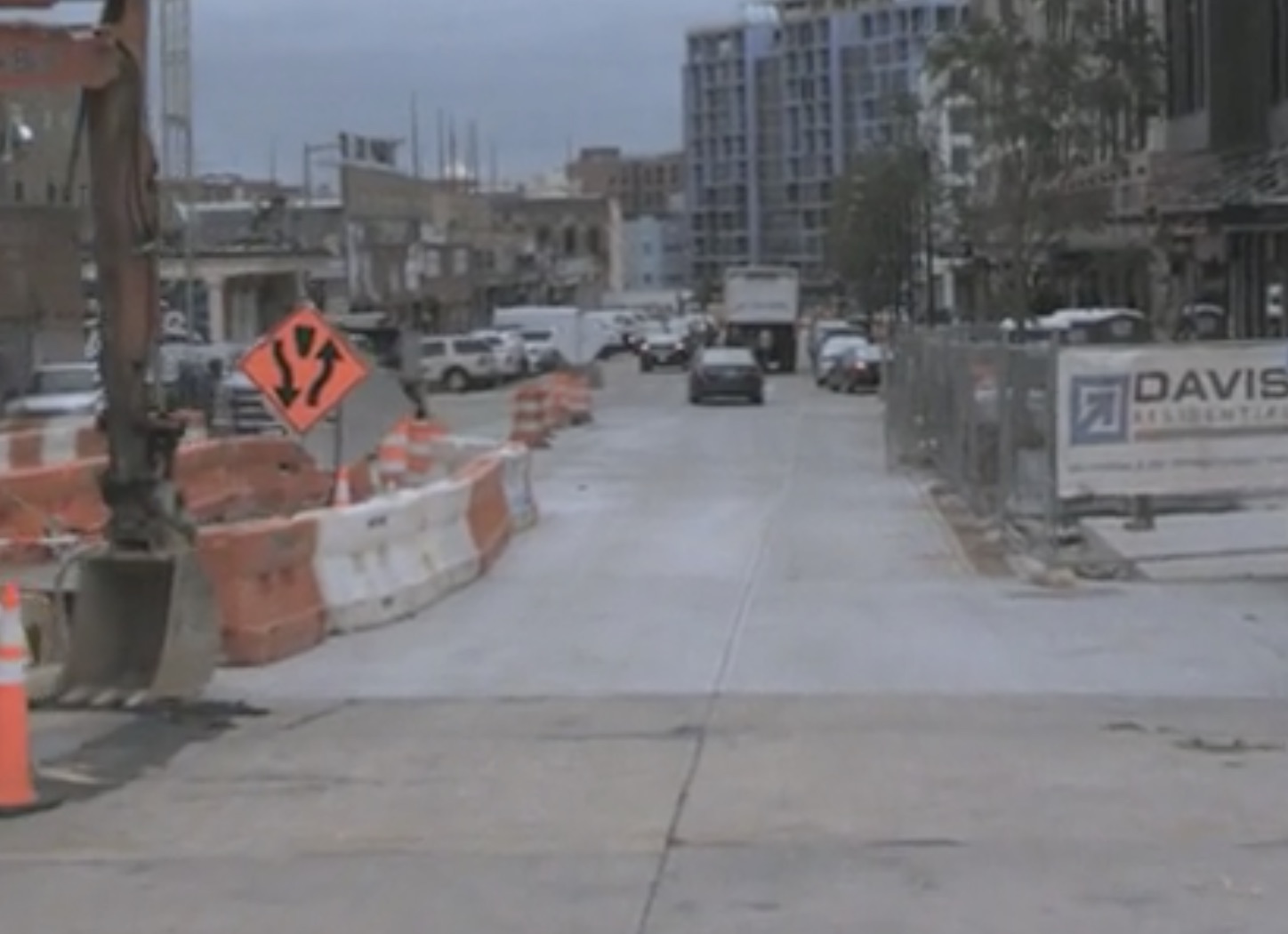}
}
\subfloat[Type III Traffic Barricades]{
\includegraphics[height=2.5cm]{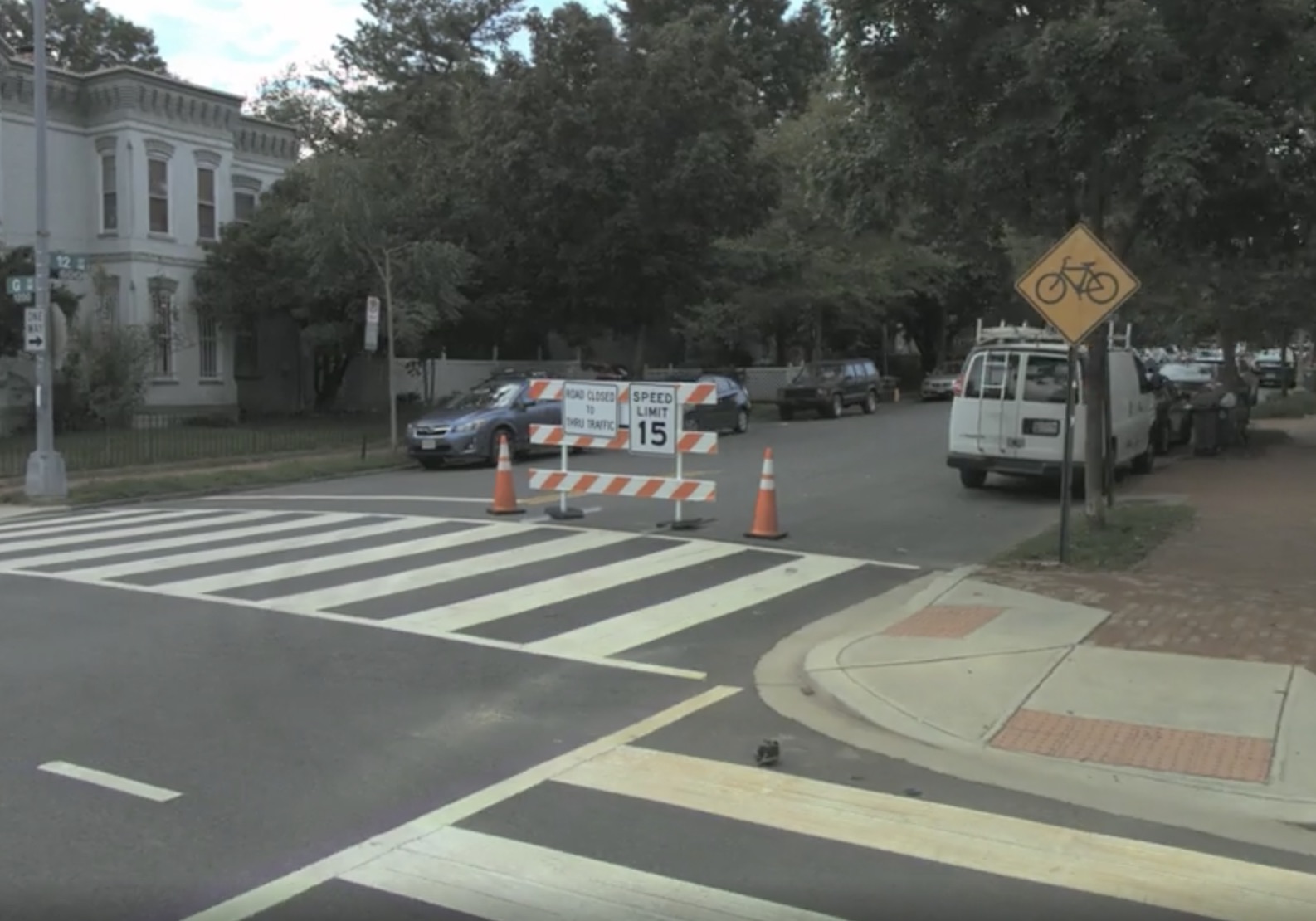}
}
\subfloat[Fallen Trees]{
\includegraphics[height=2.5cm]{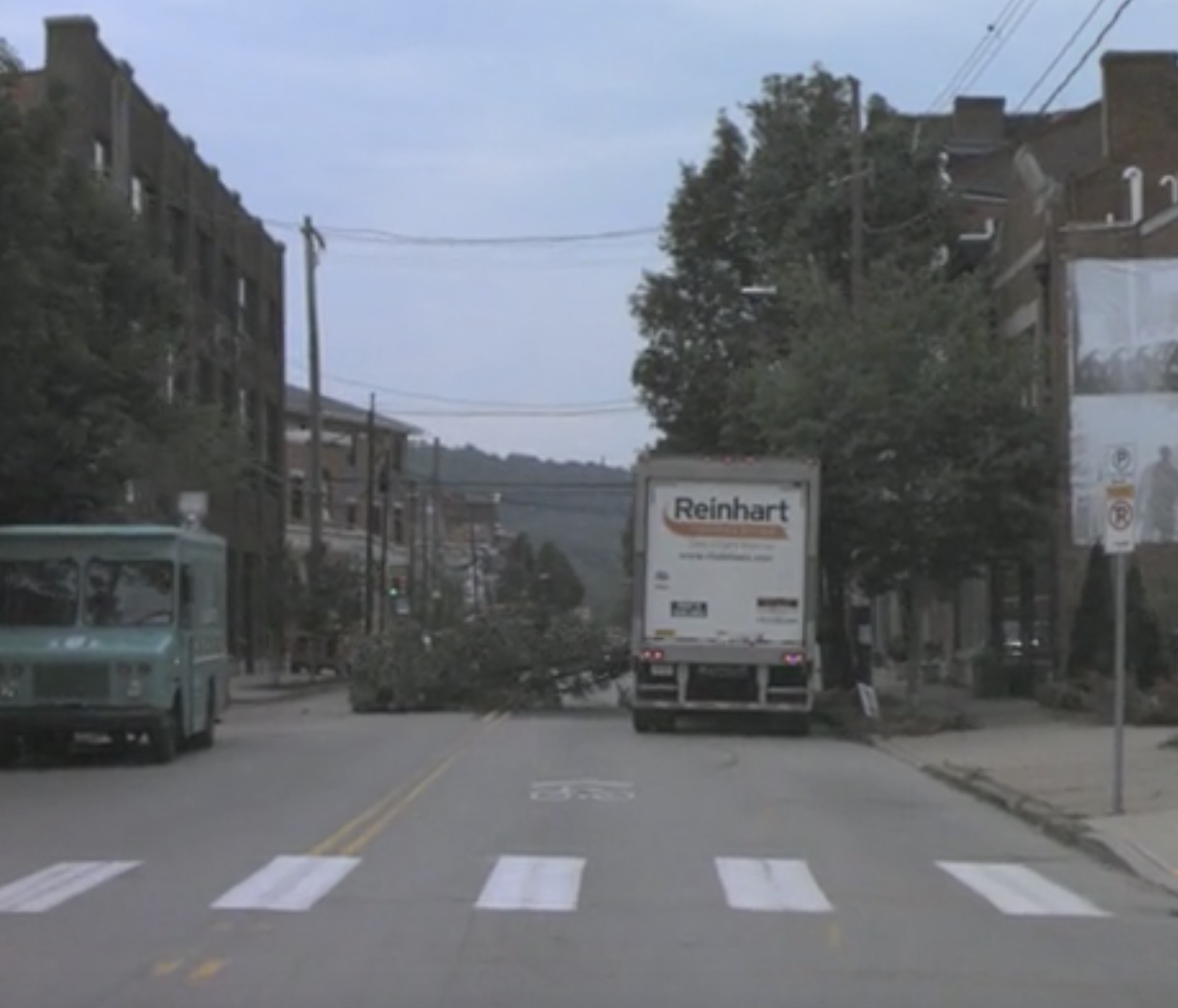}
}
\hspace{0mm}
\subfloat[Construction Signs]{
\includegraphics[height=2.5cm]{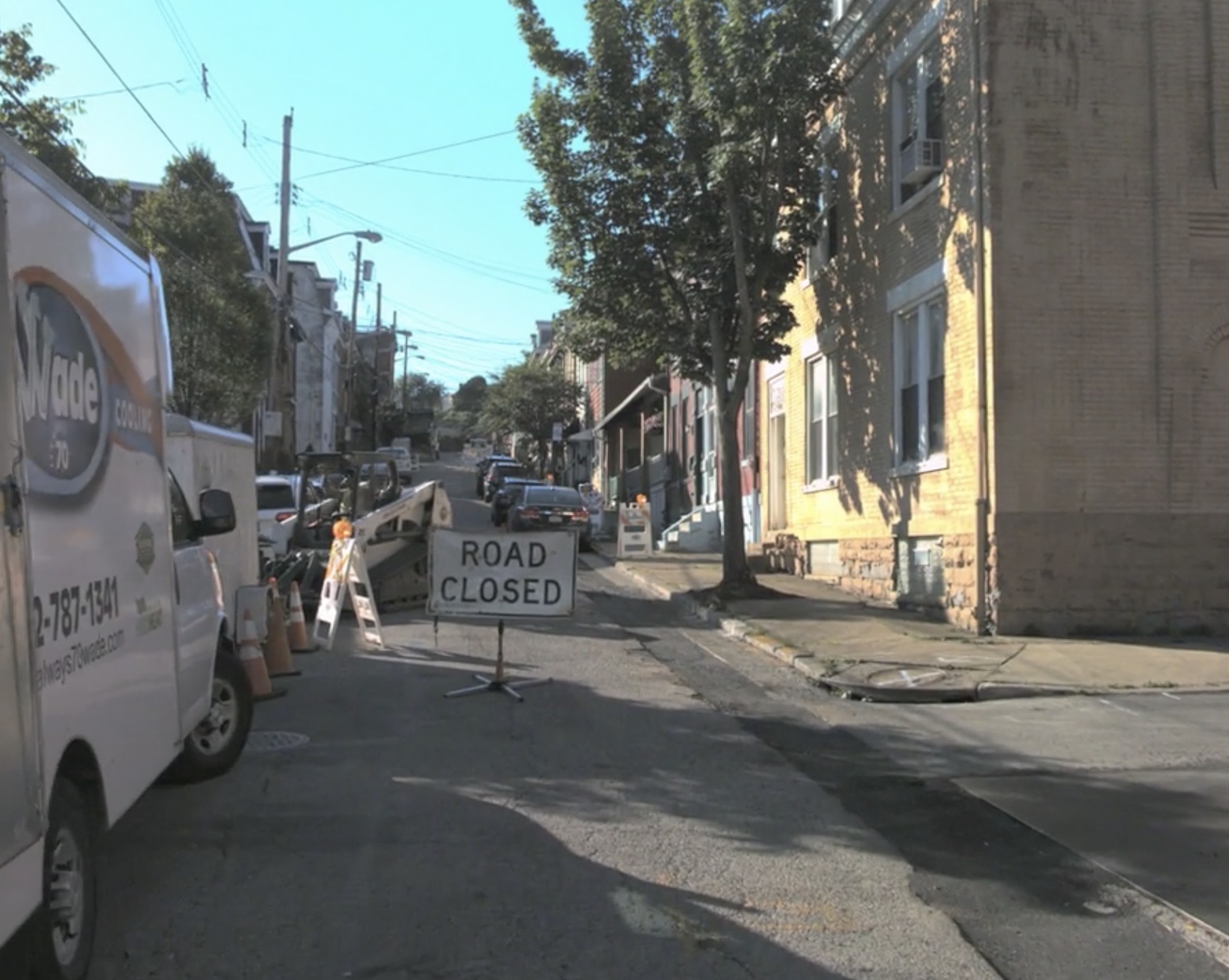}
}
\subfloat[Traffic Barrels / Drums]{
\includegraphics[height=2.5cm]{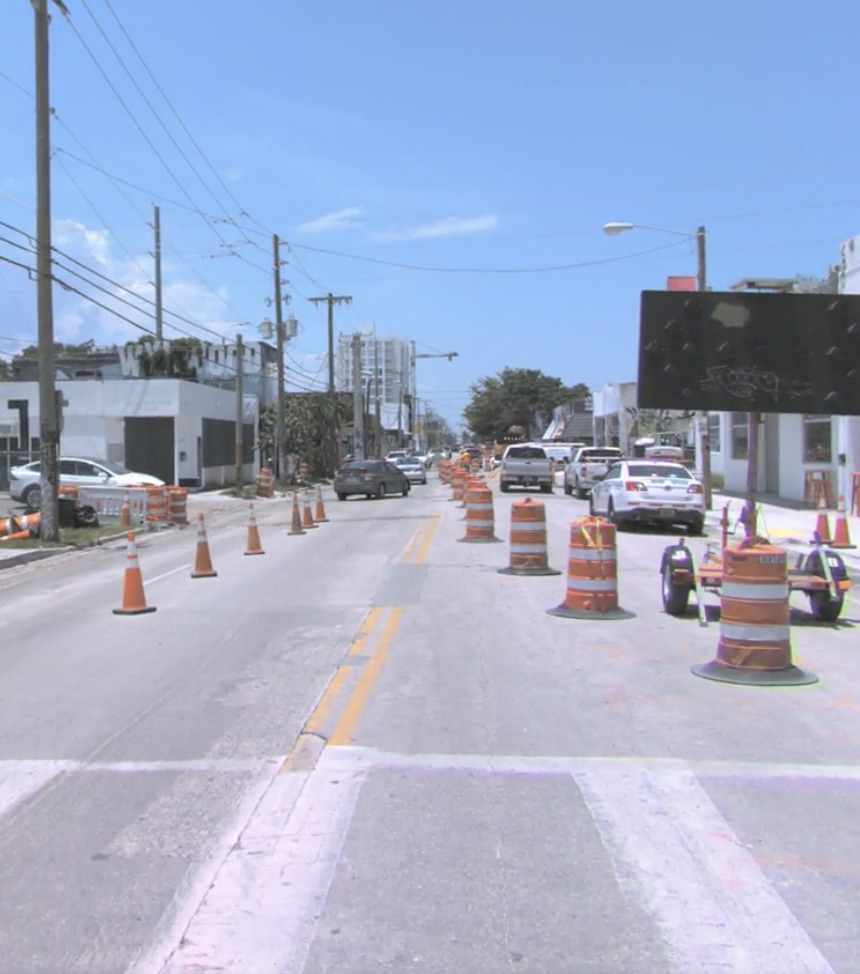}
}
\subfloat[Arrowboard Trailers]{
\includegraphics[height=2.5cm]{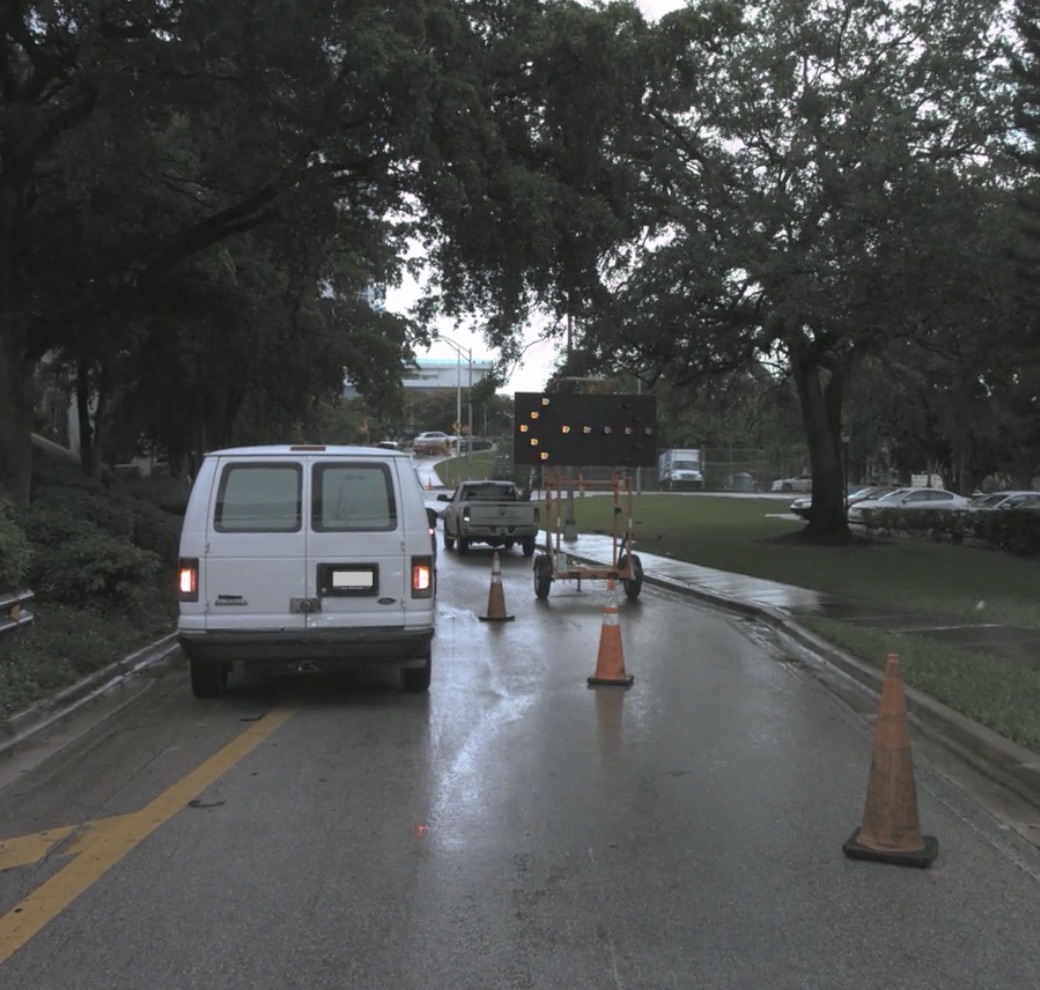}
}
%
%\subfloat[img 1]{
%\includegraphics[height=3cm]{figs/dynamic_obj_examples/dynamic_obj_example_road_fence_npCIXr80OzGjjtu77PhL9x8j8xpx2eZQ__2020-09-02-Z1F0066.png}
%}
\caption{Scenes with temporary object-related map changes collected in our fleet data. Such scenes are not the focus of our work; rather, we believe such changes should be addressed by onboard object recognition systems.}
\label{fig:dynamicobjs}
\end{figure*}

In Figure \ref{fig:dynamicobjs}, we show examples of object-centric map changes inside our TbV dataset, which we do not annotate and are not the focus of our work.

\section*{Appendix I: Additional Analysis of Map Change Frequency}

\begin{table}[]
\centering
\vspace{-1em}
\caption{Entities included in our HD map representation.}
%\vspace{-1em}
\begin{adjustbox}{max width=0.7\columnwidth}
\begingroup
\renewcommand{\arraystretch}{1.2} % General space between rows (1 standard)
    %\begin{tabular}{@{}p{1mm}@{\hspace{4mm}}l@{\hspace{4mm}}l@{\hspace{4mm}}r@{}}
    \begin{tabular}{cl}
    \toprule 
    \textsc{\textbf{HD Map Entity}} & \textsc{\textbf{Corresponding attributes}} \\
    \midrule
    \textsc{Pedestrian Crossings} & \textsc{2 edges oriented along its principal axis} \\ %line segments, corresponding to the edges oriented along its principal axis} \\
    \hline
    \multirow{5}{*}{\textsc{Lanes}} & \textsc{Boundaries: 3d left and right polylines} \\
    & \textsc{Color: yellow, white, or implicit} \\
    & \textsc{Boundary marking type} \\ %:  solid, double-solid, dashed, double-dashed, dashed-solid, or solid-dashed} \\
    & \textsc{Connectivity} \\ %: left neighbor, right neighbor, predecessors, and successors}\\
    & \textsc{Lane type: bike or vehicle lane} \\
    & \textsc{In intersection: true or false} \\
    \hline
    \textsc{Driveable area} & \textsc{polygons} \\
    \hline
    \textsc{Ground surface height} & \textsc{floating point height values at 30 centimeter resolution} \\
    \bottomrule
    \end{tabular}
\endgroup
\end{adjustbox}
%\vspace{-1em}
\label{tab:mapattributes}
\end{table}

In Section \ref{sec:annotation-and-change-freq} and Table \ref{tab:city-tile-spatialchange-prob} of the main paper, we present an analysis of map change frequency. In this section, we provide additional analysis, an extended table, and derivations of our estimates. Map changes occur at random as part of a stochastic process. While some changes are coordinated at a city-administration level, it is still difficult to predict to a specific date or time when construction crews will complete changes. As discussed in the main text, we reason about square spatial areas of size $30 \mbox{ m} \times 30 \mbox{ m} $, which we refer to as tiles, which cover $900 \mbox{ m}^2$ each.

\paragraph{Derivation: Probability of an Encounter} We consider the probability of entering a spatial area that has undergone a crosswalk or lane geometry within it. In other words, it is the probability of encountering a changed area, and thus we name it $p_{eca}$. In order to estimate the probability of encountering a changed area, rather than computing the ratio $\Big(\frac{\text{num. change-discovery miles}}{\text{num. fleet miles}}\Big)$, we compute the ratio $\Big(\frac{\text{num. tiles where change is observed}}{\text{num. tiles entered by fleet}}\Big)$. We do not require that the autonomous vehicle directly drove over the changed tile, as an observed change can very well still affect driving behavior.   We model the probability as a Bernoulli($p$) r.v., with  $p \approx 5.517 \times 10^{-5} $ across the more than 5 North American cities we analyze. A visit would occur once per every 18,124 times a vehicle enters such areas. % If a vehicle drives for \textcolor{red}{XXX} miles per day (say 4,000-5,000 times visits to such area per day), it might encounter such a change once every 4 days.

%\textbf{Derivation of U.S. Estimate}
%Consider 54 spatial units crossed to drive one mile, as
While the change percentage may seem inconsequential, one must consider that drivers in the United States are estimated to drive 3.225 trillion miles per year, according to the U.S. Department of Transportation \cite{UsDot19pr_DrivingMoreThanEverBefore}. If one were to consider our rate of change equal to the rate of change of any stretch of road within the United States, this would amount to an \emph{upper bound} of 9B encounters of spatial areas with changed lane geometry or crosswalks, per year:
\begin{equation}
  \hspace{-3mm}  \frac{  3.225 \cdot 10^{12} \text{miles} } {\text{1 year}} \cdot  \frac{\text{1609 m }}{\text{1 mile}} \cdot \frac{\text{1 tile}}{\text{30 m}} \cdot \frac{ 5.517 \cdot 10^{-5} \text{ changes} }{\text{ 1 tile}} \approx 9.5 B
\end{equation}
This derivation assumes that all roads (including highways) are changed as often as urban roads (a generous estimate).

\begin{table}[!h]
    \centering
    \caption{Across six particular cities, we analyze the probability of change for a $30 m \times 30 m$ spatial area. Since we can likely only catch changes for spatial areas that are somewhat frequently visited, we require that an area is visited by fleet at least $n=5$ times. We provide $n=1$ as well as a lower bound.  }
    \begin{adjustbox}{max width=\columnwidth}
	\begingroup
    \begin{tabular}{l|cc|cc}
    \toprule
                         & \multicolumn{2}{c}{$\geq$\textsc{ 5 visits by fleet}} & \multicolumn{2}{|c}{$\geq$\textsc{ 1 visit by fleet }}  \\
        \textsc{City Name} & \textsc{Probability }          & \textsc{Up to T tiles}    & \textsc{Probability}          & \textsc{Up to T tiles}    \\
                         &  \textsc{of Change}            & \textsc{in a thousand}    &   \textsc{of Change}             &  \textsc{in a thousand}   \\
                         &  \textsc{per tile}            & \textsc{will change }    &     \textsc{per tile}           &  \textsc{will change }   \\
                         &                                & \textsc{ in 5 months}    &                               &   \textsc{in 5 months}    \\
        \midrule
        \textsc{Pittsburgh} & 0.0068 & 7 & 0.0052 & 5\\
        \textsc{Detroit} & 0.0056 & 6 & 0.0049 & 5\\
        \textsc{Washington, D.C.} & 0.0046 & 5 & 0.0037 & 4\\
        \textsc{Miami} & 0.0038 & 4 & 0.0027 & 3\\
        \textsc{Austin} & 0.0009 & 0.9 & 0.0006 & 0.6\\
        \textsc{Palo Alto} & 0.0007 & 0.7 & 0.0006 & 0.6\\ 
    \bottomrule
    \end{tabular}
    \endgroup
    \end{adjustbox}
    \label{tab:full-map-change-stats-table}
\end{table}

\noindent \textbf{Derivation: Probability per Spatial Area} We next estimate the probability of each unique tile in a city seeing a crosswalk or lane geometry change, which we also model as a Bernoulli($p$) random variable, with $p$ estimated as:

\begin{equation}
    p = \frac{\text{\# unique changed tiles in city}}{\text{\# unique tiles in city visited at least n times by fleet}}
\end{equation}
where the numerator and denominator are both measured over $k$ months.

In Table \ref{tab:full-map-change-stats-table}, we analyze the probability of change for a $30 m \times 30 m$ spatial area across six particular cities. Since we can likely only catch changes for spatial areas that are somewhat frequently visited, we require that an area is visited by fleet at least $n=5$ times over $k=5$ months.

\section*{Appendix J: Synthetic Map Perturbation Technique}

\begin{table}[h]
\centering
%\vspace{-1em}
\caption{Training dataset statistics and types of synthetic changes generated from a small subset of logs from \emph{TbV-Beta}. The map rendering counts enumerated below correspond to 25,393 BEV sensor data renderings. Note that a single BEV sensor data rendering and its associated vector map data can often support multiple types of synthetic changes. However, not all scenes can support all synthetic change types. For example, in order to delete a crosswalk from a local map, a crosswalk must be present in the local vicinity of the egovehicle. Significantly more training data can be generated from both \textit{TbV-Beta} and \textit{TbV-1.0} compared to what is shown here (See Table \ref{tab:testsetstatistics}). }
%\vspace{-1em}
\begin{adjustbox}{max width=0.8\columnwidth}
\begingroup
\renewcommand{\arraystretch}{1.2} % General space between rows (1 standard)
    %\begin{tabular}{@{}p{1mm}@{\hspace{4mm}}l@{\hspace{4mm}}l@{\hspace{4mm}}r@{}}
    \begin{tabular}{clc}
    \toprule 
        \textsc{\textbf{Change Category}} & \textsc{\textbf{Description of Change}} & \textsc{\textbf{\# Training }} \\
        & & \textsc{\textbf{Examples Used}} \\
        \midrule
        %\textsc{BEV Sensor Images} & \textsc{N/A} & \textsc{25,393} \\
        %\midrule
        \textsc{No change} & \textsc{None} & \textsc{25,263} \\
        \midrule
        \multirow{4}{*}{\textsc{Lane Geometry Changes}} & \textsc{Delete Lane Marking} & 19,870 \\
                   & \textsc{Change Lane Marking Color} & 25,098 \\
                     & \textsc{Change Lane Boundary Dash-Solid} & 19,875 \\
                       & \textsc{Add Bike Lane} & 21,529 \\
        \midrule
         \multirow{2}{*}{\textsc{Crosswalk Changes}} & \textsc{Delete Crosswalk} & 9,627 \\
                 & \textsc{Insert Crosswalk} & 23,166 \\
        \bottomrule
    \end{tabular}
\endgroup
\end{adjustbox}
%\vspace{-1.5em}
\label{tab:perturbationstatistics}
\end{table}

In Section \ref{sec:tbv-synthesis-mismatched-data}, Table \ref{tab:perturbationstatistics}, and Figure \ref{fig:6-types-synthetic-changes} of the main text, we enumerate a number of hand-designed priors we use to generate realistic-appearing synthetic maps. In this section, we provide detailed descriptions of the generation process.

\subsection*{J.1. Priors on the Crosswalk Perturbation Procedure}
Our main observations from studying mapped data are that crosswalks are generally located near intersections, are orthogonal to lane segment tangents, and have little to no area overlap with other crosswalks. Accordingly, we first sample a random lane segment which will be spanned by the generated, synthetic crosswalk. We perform this random sampling from a biased but normalized probability distribution; lane segments within intersections achieve 4.5x the weight of non-intersection lane segments. In order to determine the orientation of the synthesized crosswalk's principal axis, we compute the normal to the centerline of the sampled lane segment at a randomly sampled waypoint. This waypoint is sampled from 50 waypoints that we interpolate along the centerline. We ensure that the sampled waypoint is not within the outermost 1/8 of pixels along any border of the rendered map image (i.e. within 15 m according to  $\ell_\infty$ norm from the egovehicle). This measure is to allow some perturbation of the random crop for data augmentation, without losing visibility of the changed entity.

Next, in order to determine how many total lane segments the crosswalk must cross in order to span the entire road, we must determine the road extent. We approximate it as the union of all nearby lane segment polygons. The line representing the principal axis of the crosswalk may intersect with this road polygon in more than two locations, since it is often non-convex. We choose the shortest possible length segment that spans the road polygon to be valid, and thus find the closest two intersections to the sampled centerline waypoint. We randomly sample a crosswalk width $w$ in meters from a normal distribution $w \sim \mathcal{N}(\mu=3.5,\sigma=1)$, but clip to the range $w \in [2,4]$ meters afterwards, in accordance to our empirical observations of the real-world distribution.

If the rendered synthetic crosswalk has overlap with any other real crosswalk above a threshold of $\text{IoU}=0.05$, we continue to sample until we succeed. The crosswalk is rendered as a rectangle, bounded between two long edges both extending along the principal axis of the crosswalk. We use alternating parallel strips of white and gray to color the object. Crosswalks are deleted by simply not rendering them in the rasterized image.

\subsection*{J.2. Lane Geometry Perturbation Procedure} Our main observations from studying real-world map changes are that lane changes generally occur over a chain of lane segments, with combined length often over tens or hundreds of meters, although at times the combined length is far shorter. Accordingly, we use the directed lane graph to sample random connected sequences of lane segments, respecting valid successors. We then manipulate either the left or the right boundary only (not both) of this lane sequence.

Our general procedure is to start this sequence at a random lane well-within the field of view of the BEV image. As before, we ensure that the sampled marking is not entirely contained within the outermost 1/8 of pixels along any border of the rendered map image (i.e. within 15 m according to  $\ell_\infty$ norm from the egovehicle).

When deleting lane boundaries, we sample only painted yellow or white lane boundary markings.  When changing the color or structure of lane boundaries, we sample lane boundary markings of any color (including those that are implicit). When adding a bike lane, we sample a sequence of 5 lane segments. For marking deletion and changes to lane marking color and structure, we sample a sequence of length 3.

We render these boundaries as colored polylines; we use red for implicit boundaries, and yellow and white for lane markings of their respective color. Lane boundary markings are deleted by simply not rendering them in the rasterized image. 

Bike lanes generally represent the rightmost lane in the United States. Accordingly, we synthesize a valid location for a new bike lane by iterating through the lane graph until there is no right neighbor; by dividing this rightmost lane into half, we can create two half-width lanes in place of one. We use solid white lines to represent their boundaries.

\section*{Appendix K: Datasheet for TbV}
\label{appendix:datasheet}
% \documentclass{article}

% % if you need to pass options to natbib, use, e.g.:
% %     \PassOptionsToPackage{numbers, compress}{natbib}
% % before loading neurips_data_2021

% % ready for submission
% \usepackage{neurips_data_2021}

% % to compile a preprint version, add the [preprint] option, e.g.:
% %     \usepackage[preprint]{neurips_data_2021}
% % This will indicate that the work is currently under review.

% % to compile a camera-ready version, add the [final] option, e.g.:
% %     \usepackage[final]{neurips_data_2021}

% % to avoid loading the natbib package, add option nonatbib:
% %    \usepackage[nonatbib]{neurips_data_2021}

% % Submissions to the datasets and benchmarks are typically non anonymous. If you feel strongly that you must submit anonymously, you can compile an anonymous version by adding the [anonymous] option, e.g.:
% %     \usepackage[anonymous]{neurips_data_2021}
% % This will hide all author names.

% \usepackage[utf8]{inputenc} % allow utf-8 input
% \usepackage[T1]{fontenc}    % use 8-bit T1 fonts
% \usepackage{hyperref}       % hyperlinks
% \usepackage{url}            % simple URL typesetting
% \usepackage{booktabs}       % professional-quality tables
% \usepackage{amsfonts}       % blackboard math symbols
% \usepackage{nicefrac}       % compact symbols for 1/2, etc.
% \usepackage{microtype}      % microtypography
% \usepackage{xcolor}         % colors

% \usepackage{authblk}
% \makeatletter
% \renewcommand\AB@affilsepx{ \hspace{1em}   \protect\Affilfont}
% \makeatother

\newcommand{\dssectionheader}[1]{%
   \noindent\framebox[\columnwidth]{%
      { \textbf{\textcolor{violet}{#1}}}
   }
}

\newcommand{\dsquestion}[1]{%
    {\noindent  \textcolor{violet}{\textbf{#1}}}
}

\newcommand{\dsquestionex}[2]{%
    {\noindent  \textcolor{violet}{\textbf{#1} #2}}
}

\newcommand{\dsanswer}[1]{%
   {\noindent #1 \medskip}
}

% \title{Data Sheeet for TbV}

% The \author macro works with any number of authors. There are two commands
% used to separate the names and addresses of multiple authors: \And and \AND.
%
% Using \And between authors leaves it to LaTeX to determine where to break the
% lines. Using \AND forces a line break at that point. So, if LaTeX puts 3 of 4
% authors names on the first line, and the last on the second line, try using
% \AND instead of \And before the third author name.

% \author[1,2]{\textbf{John Lambert}}
% \author[1,2]{\textbf{James Hays}}
% \affil[1]{Argo AI} \affil[2]{Georgia Institute of Technology}

% \begin{document}

% \maketitle

In this appendix, we answer the questions laid out in \emph{Datasheets for Datasets} by Gebru \emph{et al.} \cite{Gebru21_datasheets}.

% Color picked from the Datasets for Datasheets paper
\definecolor{darkblue}{RGB}{46,25, 110}

% \begin{singlespace}
% \begin{multicols}{2}

%%%%%%%%%%%%%%%%%%%%%%%%%%%%%%%%%%%%%%%%%%
\dssectionheader{Motivation}

\dsquestionex{For what purpose was the dataset created?}{Was there a specific task in mind? Was there a specific gap that needed to be filled?} % Please provide a description.}

\dsanswer{
TbV was created to allow the community to improve the state of the art in machine learning tasks related to mapping, that are vital for self-driving.

To our knowledge, no prior datasets has ever been publicly released for HD map change detection. It is also one of the largest sensor datasets ever released, paired with HD maps, allowing for new exploration of the synergies between the sensor data and map data. 
}

\dsquestion{Who created this dataset (e.g., which team, research group) and on behalf of which entity (e.g., company, institution, organization)?}

\dsanswer{
The TbV dataset was created by researchers who were employed at Argo AI.
}

\dsquestionex{Who funded the creation of the dataset?}{}%{If there is an associated grant, please provide the name of the grantor and the grant name and number.}

\dsanswer{
The creation of this dataset was funded by Argo AI.
}

%\dsquestion{Any other comments?} N/A

%%%%%%%%%%%%%%%%%%%%%%%%%%%%%%%%%%%%%%%%%%
\bigskip
\dssectionheader{Composition}

\dsquestionex{What do the instances that comprise the dataset represent (e.g., documents, photos, people, countries)?}{ Are there multiple types of instances (e.g., movies, users, and ratings; people and interactions between them; nodes and edges)?} %Please provide a description.}

\dsanswer{
The core instances for TbV are brief ``scenarios'' or ``logs'' of that represent a continuous observation of a scene around a self-driving vehicle. On average, each scenario is 54 seconds in duration, although some capture as little as 4 seconds, while others last for up to 117 seconds.

Each scenario has an HD map representing lane boundaries, crosswalks, drivable area, etc. They also contain a raster map of ground height at 0.3 meter resolution.
}

\dsquestion{How many instances are there in total (of each type, if appropriate)?}

\dsanswer{
The TbV Dataset has 1043 scenarios. 
}

\dsquestionex{Does the dataset contain all possible instances or is it a sample (not necessarily random) of instances from a larger set?}{ If the dataset is a sample, then what is the larger set? Is the sample representative of the larger set (e.g., geographic coverage)? If so, please describe how this representativeness was validated/verified. If it is not representative of the larger set, please describe why not (e.g., to cover a more diverse range of instances, because instances were withheld or unavailable).}

\dsanswer{
The scenarios in the dataset are a sample of the set of observations made by a fleet of self-driving vehicles. The data is not uniformly sampled.\\

The ``negative'' instances in the dataset were chosen to include specific examples where an HD map has become out-of-date, due to real-world changes. \\

The ``positive'' instances in the dataset were chosen to include interesting behavior (e.g. cars making unexpected maneuvers), to contain interesting weather (e.g. rain and snow), and to be geographically diverse (spanning 6 cities -- Pittsburgh, Detroit, Austin, Palo Alto, Miami, and Washington D.C.).
}

\dsquestionex{What data does each instance consist of? “Raw” data (e.g., unprocessed text or images) or features?}{In either case, please provide a description.}

\dsanswer{
Each scenario has 20 fps video from 7 ring cameras, 20 fps video from two forward-facing stereo cameras, and 10 Hz LiDAR returns from two out-of-phase 32-beam LiDARs. The ring cameras are synchronized to fire when either LiDAR sweeps through their field of view. Each scenario contains vehicle pose over time and calibration data to relate the various sensors.

The HD map associated with each scenario contains polylines describing lanes, crosswalks, and drivable area. Lanes form a graph with predecessors and successors, e.g. a lane that splits can have two successors. Lanes have precisely localized lane boundaries that include paint type (e.g. double solid yellow). Drivable area, also described by a polygon, is the area where it is possible (but not necessarily legal) to drive without damaging the vehicle. It includes areas such as road shoulders.
}

\dsquestionex{Is there a label or target associated with each instance?}{}%{If so, please provide a description.}

\dsanswer{
Yes. For the logs found in the train and synthetic validation splits, an up-to-date HD map serves as a label, as these are ``positive'' logs, where the map and sensor data are in agreement.

For the logs found in the ``real'' validation and test splits, 3d coordinates of polygons or polylines are manually annotated for areas where the map has changed, for lane paint and crosswalks, specifically.

In addition, the LiDAR depth estimates can act as ground truth for monocular depth estimation. The vehicle pose data could be considered ground truth labels for visual odometry. The evolving point cloud itself can be considered ground truth for point cloud forecasting.

}

\dsquestionex{Is any information missing from individual instances?}{If so, please provide a description, explaining why this information is missing (e.g., because it was unavailable). This does not include intentionally removed information, but might include, e.g., redacted text.}

\dsanswer{
No. To our knowledge, all instances should be complete.
}

\dsquestionex{Are relationships between individual instances made explicit (e.g., users’ movie ratings, social network links)?}{If so, please describe how these relationships are made explicit.}

\dsanswer{
Each instance of the dataset (a vehicle ``log'') is disjoint. Each carries their own HD map for the region around their scenario. These HD maps may overlap spatially, though. For example, they may be captured at the same intersection, but separated in time by several months. If a user of the dataset wanted to recover the spatial relationship between scenarios, they could do so through our development kit.
}

\dsquestionex{Are there recommended data splits (e.g., training, development/validation, testing)?}{If so, please provide a description of these splits, explaining the rationale behind them.}

\dsanswer{
We define splits of the TbV dataset. The train, validation, and test set include 799 / 111 / 133 logs each.
}

\dsquestionex{Are there any errors, sources of noise, or redundancies in the dataset?}{}%{If so, please provide a description.}

\dsanswer{
    Every sensor used in the dataset -- ring cameras and lidar -- has noise associated with it. Pixel intensities, lidar intensities, and lidar point 3D locations all have noise. Lidar points are also quantized to float16 which leads to roughly a centimeter of quantization error. 6 degree of freedom vehicle pose also has noise. The calibration specifying the relationship between sensors can be imperfect.
    
    The HD map for each scenario can contain noise, both in terms of lane boundary locations and precise ground height.
}

\dsquestionex{Is the dataset self-contained, or does it link to or otherwise rely on external resources (e.g., websites, tweets, other datasets)?}{If it links to or relies on external resources, a) are there guarantees that they will exist, and remain constant, over time; b) are there official archival versions of the complete dataset (i.e., including the external resources as they existed at the time the dataset was created); c) are there any restrictions (e.g., licenses, fees) associated with any of the external resources that might apply to a future user? Please provide descriptions of all external resources and any restrictions associated with them, as well as links or other access points, as appropriate.}

\dsanswer{
The data itself is self-hosted, %like Argoverse 1.0 [see \url{https://www.argoverse.org/}], 
and we will maintain public links to all previous versions of the dataset in case of updates. %The data is independent of any previous datasets, including Argoverse 1.0.
}

\dsquestionex{Does the dataset contain data that might be considered confidential (e.g., data that is protected by legal privilege or by doctor-patient confidentiality, data that includes the content of individuals non-public communications)?}{}%{If so, please provide a description.}

\dsanswer{
No.
}

\dsquestionex{Does the dataset contain data that, if viewed directly, might be offensive, insulting, threatening, or might otherwise cause anxiety?}{}%{If so, please describe why.}

\dsanswer{
No.
}

\dsquestionex{Does the dataset relate to people?}{}%{If not, you may skip the remaining questions in this section.}

\dsanswer{
Yes, the dataset contains images and behaviors of thousands of people on public streets.
}

\dsquestionex{Does the dataset identify any subpopulations (e.g., by age, gender)?}{}%{If so, please describe how these subpopulations are identified and provide a description of their respective distributions within the dataset.}

\dsanswer{
No.
}

\dsquestionex{Is it possible to identify individuals (i.e., one or more natural persons), either directly or indirectly (i.e., in combination with other data) from the dataset?}{If so, please describe how.}

\dsanswer{
We do not believe so. Image data has been anonymized via blurring.  Faces and license plates are obfuscated by replacing their corresponding bounding box with a $5\times 5$ grid, where each grid cell is the average color of the original pixels in that grid cell. The anonymization is done manually. For example, a person sitting on their front porch 10 meters from the road would have their face obscured.
}

\dsquestionex{Does the dataset contain data that might be considered sensitive in any way (e.g., data that reveals racial or ethnic origins, sexual orientations, religious beliefs, political opinions or union memberships, or locations; financial or health data; biometric or genetic data; forms of government identification, such as social security numbers; criminal history)?}{}%{If so, please provide a description.}

\dsanswer{
N/A.
}

%\dsquestion{Any other comments?}

%%%%%%%%%%%%%%%%%%%%%%%%%%%%%%%%%%%%%%%%%%
\bigskip
\dssectionheader{Collection Process}

\dsquestionex{How was the data associated with each instance acquired?}{Was the data directly observable (e.g., raw text, movie ratings), reported by subjects (e.g., survey responses), or indirectly inferred/derived from other data (e.g., part-of-speech tags, model-based guesses for age or language)? If data was reported by subjects or indirectly inferred/derived from other data, was the data validated/verified? If so, please describe how.}

\dsanswer{
The sensor data was directly acquired by a fleet of autonomous vehicles.
}

\dsquestionex{Over what timeframe was the data collected? Does this timeframe match the creation timeframe of the data associated with the instances (e.g., recent crawl of old news articles)?}{If not, please describe the timeframe in which the data associated with the instances was created.}

\dsanswer{
The data was collected from May 2020 to March 2021.
}

\dsquestionex{What mechanisms or procedures were used to collect the data (e.g., hardware apparatus or sensor, manual human curation, software program, software API)?}{How were these mechanisms or procedures validated?}

\dsanswer{
The Trust but Verify (TbV) data comes from Argo `Z1' fleet vehicles. These vehicles use Velodyne lidars and traditional RGB cameras. All sensors are calibrated by Argo. HD maps are created and validated through a combination of computational tools and human annotations. Map change labels are created through human annotation.
}

\dsquestion{If the dataset is a sample from a larger set, what was the sampling strategy (e.g., deterministic, probabilistic with specific sampling probabilities)?}

\dsanswer{
The dataset scenarios were chosen from a larger set through manual review. The test set scenarios were selected to illustrate unambiguous map changes.
}

\dsquestion{Who was involved in the data collection process (e.g., students, crowdworkers, contractors) and how were they compensated (e.g., how much were crowdworkers paid)?}

\dsanswer{
Argo employees and Argo interns curated the data. Data collection and data annotation was done by Argo employees. Crowdworkers were not used.
}

\dsquestionex{Were any ethical review processes conducted (e.g., by an institutional review board)?}{}%{If so, please provide a description of these review processes, including the outcomes, as well as a link or other access point to any supporting documentation.}

\dsanswer{
No.
}

\dsquestionex{Does the dataset relate to people?}{}%{If not, you may skip the remaining questions in this section.}

\dsanswer{
Yes.
}

\dsquestion{Did you collect the data from the individuals in question directly, or obtain it via third parties or other sources (e.g., websites)?}

\dsanswer{
The data is collected from vehicles on public roads, not from a third party.
}

\dsquestionex{Were the individuals in question notified about the data collection?}{If so, please describe (or show with screenshots or other information) how notice was provided, and provide a link or other access point to, or otherwise reproduce, the exact language of the notification itself.}

\dsanswer{
No, but the data collection was not hidden. The Argo fleet vehicles are well-marked and have obvious cameras and LiDAR sensors. The vehicles only capture data from public roads.
}

\dsquestionex{Did the individuals in question consent to the collection and use of their data?}{If so, please describe (or show with screenshots or other information) how consent was requested and provided, and provide a link or other access point to, or otherwise reproduce, the exact language to which the individuals consented.}

\dsanswer{
No. People in the dataset were in public settings and their appearance has been anonymized. Drivers, pedestrians, and vulnerable road users are an intrinsic part of driving on public roads, so it is important that datasets contain people so that the community can develop more accurate perception systems.
}

\dsquestionex{If consent was obtained, were the consenting individuals provided with a mechanism to revoke their consent in the future or for certain uses?}{If so, please provide a description, as well as a link or other access point to the mechanism (if appropriate).}

\dsanswer{
N/A.
}

\dsquestionex{Has an analysis of the potential impact of the dataset and its use on data subjects (e.g., a data protection impact analysis) been conducted?}{If so, please provide a description of this analysis, including the outcomes, as well as a link or other access point to any supporting documentation.}

\dsanswer{
No.
}

% \dsquestion{Any other comments?}

% \dsanswer{
% N/A.
% }

%%%%%%%%%%%%%%%%%%%%%%%%%%%%%%%%%%%%%%%%%%
\bigskip
\dssectionheader{Preprocessing/cleaning/labeling}

\dsquestionex{Was any preprocessing/cleaning/labeling of the data done (e.g., discretization or bucketing, tokenization, part-of-speech tagging, SIFT feature extraction, removal of instances, processing of missing values)?}{If so, please provide a description. If not, you may skip the remainder of the questions in this section.}

\dsanswer{
Yes. Images are reduced from their full resolution, and are JPEG compressed. 3D point locations are quantized to float16. Ground height maps are quantized to 0.3 meter resolution from their full resolution. HD map polygon vertex locations are quantized to 0.01 meter resolution. 
}

\dsquestionex{Was the “raw” data saved in addition to the preprocessed/cleaned/labeled data (e.g., to support unanticipated future uses)?}{If so, please provide a link or other access point to the “raw” data.}

\dsanswer{
Yes, but such data is not public.
}

\dsquestionex{Is the software used to preprocess/clean/label the instances available?}{}%{If so, please provide a link or other access point.}

\dsanswer{
No.
}

% \dsquestion{Any other comments?}

% \dsanswer{
% N/A.
% }

%%%%%%%%%%%%%%%%%%%%%%%%%%%%%%%%%%%%%%%%%%
\bigskip
\dssectionheader{Uses}

\dsquestionex{Has the dataset been used for any tasks already?}{}%{If so, please provide a description.}

\dsanswer{
Yes, this manuscript benchmarks a novel HD map change detection method on the TbV dataset.
}

\dsquestionex{Is there a repository that links to any or all papers or systems that use the dataset?}{}%{If so, please provide a link or other access point.}

\dsanswer{
Yes, at \url{https://github.com/johnwlambert/tbv}. We plan to add a leaderboard for the HD map change detection task using the test split of the TbV dataset. %For the Argoverse 1.0 datasets, we maintain four leaderboards for 3D Tracking [\url{https://eval.ai/web/challenges/challenge-page/453/overview}], 3D Detection [\url{https://eval.ai/web/challenges/challenge-page/725/overview}], Motion Forecasting [\url{https://eval.ai/web/challenges/challenge-page/454/overview}], and Stereo Depth Estimation [\url{https://eval.ai/web/challenges/challenge-page/917/overview}]. Argoverse 1.0 was also used as the basis for a Streaming Perception challenge [\url{https://eval.ai/web/challenges/challenge-page/800/overview}].

}

\dsquestion{What (other) tasks could the dataset be used for?}

\dsanswer{
The TbV dataset could be used for research on visual odometry, lane detection, synthetic HD map generation, map automation, self-supervised learning, scene flow, point cloud forecasting, and more.
}

\dsquestionex{Is there anything about the composition of the dataset or the way it was collected and preprocessed/cleaned/labeled that might impact future uses?}{For example, is there anything that a future user might need to know to avoid uses that could result in unfair treatment of individuals or groups (e.g., stereotyping, quality of service issues) or other undesirable harms (e.g., financial harms, legal risks) If so, please provide a description. Is there anything a future user could do to mitigate these undesirable harms?}

\dsanswer{
No.
}

\dsquestionex{Are there tasks for which the dataset should not be used?}{}%{If so, please provide a description.}

\dsanswer{
The dataset should not be used for tasks which depend on faithful appearance of faces or license plates since that data has been obfuscated. For example, running a face detector to try and estimate how often pedestrians use crosswalks will not result in meaningful data.
}

% \dsquestion{Any other comments?}

% \dsanswer{
% N/A.
% }

%%%%%%%%%%%%%%%%%%%%%%%%%%%%%%%%%%%%%%%%%%
\bigskip
\dssectionheader{Distribution}

\dsquestionex{Will the dataset be distributed to third parties outside of the entity (e.g., company, institution, organization) on behalf of which the dataset was created?}{}%{If so, please provide a description.}

\dsanswer{
Yes, the dataset is hosted on \url{https://www.argoverse.org/}. Our dataset requires no user registration for access. The dataset’s metadata page will include structured metadata.

In addition to long term hosting on Argoverse.org, the Creative Commons license enables rehosting by any repository. The authors will ensure that the dataset is accessible.
% like Argoverse 1.0 and 1.1.
}

\dsquestionex{How will the dataset will be distributed (e.g., tarball on website, API, GitHub)}{Does the dataset have a digital object identifier (DOI)?}

\dsanswer{
The TbV dataset is distributed as a series of tar.gz files. %, as was the case for Argoverse 1.0 and Argoverse 1.1. See \url{https://www.argoverse.org/data.html#download-link}
The files are broken up to make the process more robust to interruption (e.g. a single 1 TB file failing after 3 days would be frustrating) and to allow easier file manipulation (an end user might not have 1 TB free on a single drive, and if they do, they might not be able to decompress the entire file at once).

The dataset can be read with the Argoverse 2.0 API. See \url{https://github.com/argoai/av2-api} for details on usage.
}

\dsquestion{When will the dataset be distributed?}

\dsanswer{
The data is currently available for download, at the time of NeurIPS 2021.
}

\dsquestionex{Will the dataset be distributed under a copyright or other intellectual property (IP) license, and/or under applicable terms of use (ToU)?}{If so, please describe this license and/or ToU, and provide a link or other access point to, or otherwise reproduce, any relevant licensing terms or ToU, as well as any fees associated with these restrictions.}

\dsanswer{
Yes, the dataset is released under the same Creative Commons license as Argoverse 1.0 (CC BY-NC-SA 4.0). The authors are responsible for the contents of the dataset and are responsible for any possible violation of rights.
%Details can be seen at \url{https://www.argoverse.org/about.html#terms-of-use}.
}

\dsquestionex{Have any third parties imposed IP-based or other restrictions on the data associated with the instances?}{}%{If so, please describe these restrictions, and provide a link or other access point to, or otherwise reproduce, any relevant licensing terms, as well as any fees associated with these restrictions.}

\dsanswer{
No.
}

\dsquestionex{Do any export controls or other regulatory restrictions apply to the dataset or to individual instances?}{}%{If so, please describe these restrictions, and provide a link or other access point to, or otherwise reproduce, any supporting documentation.}

\dsanswer{
No.
}

% \dsquestion{Any other comments?}

% \dsanswer{
% N/A.
% }

%%%%%%%%%%%%%%%%%%%%%%%%%%%%%%%%%%%%%%%%%%
\bigskip
\dssectionheader{Maintenance}

\dsquestion{Who will be supporting/hosting/maintaining the dataset?}

\dsanswer{
Argo AI.
}

\dsquestion{How can the owner/curator/manager of the dataset be contacted (e.g., email address)?}

\dsanswer{
The TbV team will respond through the Github page \url{https://github.com/johnwlambert/tbv/issues} (where training code and pre-trained models have been made available). For privacy concerns, contact information may be found here: \url{https://www.argoverse.org/about.html\#privacy}.
}

% and/or through the Argoverse API: \url{https://github.com/argoai/argoverse-api/issues} (where official evaluation code will be made available). %So far, the Argoverse API page contains 2 open issues and 126 closed issues, which we have devoted considerable time to answering and addressing.

\dsquestionex{Is there an erratum?}{} %{If so, please provide a link or other access point.}

\dsanswer{
No.
}

\dsquestionex{Will the dataset be updated (e.g., to correct labeling errors, add new instances, delete instances)?}{If so, please describe how often, by whom, and how updates will be communicated to users (e.g., mailing list, GitHub)?}

\dsanswer{
It is possible that the TbV 1.0 Dataset will be updated to correct errors. %This was the case with Argoverse 1.0 which was incremented to Argoverse 1.1. 
Updates will be communicated on Github and through a mailing list we will create.
}

\dsquestionex{If the dataset relates to people, are there applicable limits on the retention of the data associated with the instances (e.g., were individuals in question told that their data would be retained for a fixed period of time and then deleted)?}{If so, please describe these limits and explain how they will be enforced.}

\dsanswer{
No.
}

\dsquestionex{Will older versions of the dataset continue to be supported/hosted/maintained?}{If so, please describe how. If not, please describe how its obsolescence will be communicated to users.}

\dsanswer{
Yes. %For example, we still host Argoverse 1.0 even though we have declared it ``deprecated''. See \url{https://www.argoverse.org/data.html#download-link}. We will use the same warning if we ever deprecate TbV 1.0.
If we ever deprecate TbV 1.0, we will continue to host it, although we will declare it ``deprecated.''
}

\dsquestionex{If others want to extend/augment/build on/contribute to the dataset, is there a mechanism for them to do so?}{If so, please provide a description. Will these contributions be validated/verified? If so, please describe how. If not, why not? Is there a process for communicating/distributing these contributions to other users? If so, please provide a description.}

\dsanswer{
Yes. %Argoverse 1.0 was another dataset also released by Argo AI, and has been built upon by CMU researchers who added 2d object annotations for the Streaming Perception Challenge. 
The Creative Commons license we use for TbV ensures that the community can do the same thing without needing Argo's permission.

We do not have a mechanism for these contributions/additions to be incorporated back into the `base' TbV Dataset. Our preference would generally be to keep the `base' dataset as is, and to give credit to noteworthy additions by linking to them. % as we have done in the case of the Streaming Perception Challenge (see link near the top of this Argoverse page: \url{https://www.argoverse.org/tasks.html}.
}

% \dsquestion{Any other comments?}

% \dsanswer{
% }

% \end{multicols}
% \end{singlespace}

% \end{document}

% \end{document}

\textbf{Environmental Impact Statement}\\
\emph{Amount of Compute Used}: We estimate 5000 CPU hours and 3000 GPU hours for all of the data extraction, preparation and experiments.

% \section{Appendix}
% \input{latex/tex/supplementary.tex}

%% 2-column table use:
% \begin{verbatim}
%     \let\linenumbers\nolinenumbers\nolinenumbers
% \end{verbatim}

\end{document}